\documentclass[11pt]{article}

\RequirePackage[OT1]{fontenc}
\usepackage{fullpage}
\usepackage{amsthm,amsmath,amssymb,amsfonts,amsbsy}
\usepackage{epsfig}
\usepackage{subfigure}
\usepackage{pifont}
\usepackage[usenames]{color}
\RequirePackage[colorlinks,citecolor=blue,urlcolor=blue]{hyperref}
\usepackage{algorithm,algorithmic}
\usepackage{natbib}
\usepackage{bm,nicefrac}

\newcommand{\ignore}[1]{}{}

\newtheorem{theorem}{Theorem}[section]
\newtheorem{proposition}[theorem]{Proposition}
\newtheorem{lemma}[theorem]{Lemma}

\newtheorem{remark}{Remark}[section]
\newtheorem{example}[theorem]{Example}

\newtheorem{innercustomexp}{Example}
\newenvironment{customexp}[1]
{\renewcommand\theinnercustomexp{#1}\innercustomexp}
{\endinnercustomexp}
\newtheorem{innercustomthm}{Theorem}
\newenvironment{customthm}[1]
{\renewcommand\theinnercustomthm{#1}\innercustomthm}
{\endinnercustomthm}
\newtheorem{innercustomprop}{Proposition}
\newenvironment{customprop}[1]
{\renewcommand\theinnercustomprop{#1}\innercustomprop}
{\endinnercustomprop}


\newcommand{\argmin}{\mathop{\rm arg\min}}

\def\R{\mathbb{R}}
\def\I{\textbf{I}}
\def\ep{\mathbb{E}}
\def\pr{\mathbb{P}}
\def\Cov{\textsf{Cov}}
\def\0{\boldsymbol{0}}
\def\a{\boldsymbol{a}}
\def\u{\boldsymbol{u}}
\def\v{\boldsymbol{v}}
\def\x{\boldsymbol{x}}
\def\w{\boldsymbol{w}}
\def\z{\boldsymbol{z}}
\def\A{\boldsymbol{A}}
\def\X{\boldsymbol{X}}

\def\xii{\boldsymbol{\xi}}

\def\de{\boldsymbol{\delta}}
\def\tee{\boldsymbol{\theta}}

\def\zet{\boldsymbol{\zeta}}

\def\S{\boldsymbol{\Sigma}}
\let\hat\widehat
\let\tilde\widetilde

\newcommand{\teeSGD}{\hat\tee_{\mathrm{SGD}}}
\newcommand{\teeDC}{\hat\tee_{\mathrm{DC}}}
\newcommand{\teeFONE}{\hat\tee_{\mathrm{FONE}}}
\newcommand{\teedis}{\hat\tee_{\mathrm{dis},K}}

\definecolor{DSgray}{cmyk}{0,1,0,0}

\renewcommand{\baselinestretch}{1.2}

\begin{document}

	\title{First-order Newton-type Estimator for Distributed Estimation and Inference}
	
		\author{Xi Chen\footnote{Stern School of Business, New York University, Email: xichen3@stern.nyu.edu.} ~ Weidong Liu\footnote{School of Mathematical Sciences, Shanghai Jiao Tong University, Email: weidongl@sjtu.edu.cn. Corresponding author.} ~ Yichen Zhang\footnote{Krannert School of Management, Purdue University, Email: zhang@purdue.edu.}}
	\date{}
	\maketitle
	\begin{abstract}
This paper studies distributed estimation and inference for a general statistical problem with a convex loss that could be non-differentiable. For the purpose of efficient computation,  we restrict ourselves to stochastic first-order optimization, which enjoys low per-iteration complexity. To motivate the proposed method, we first investigate the theoretical properties of a straightforward Divide-and-Conquer Stochastic Gradient Descent (DC-SGD) approach. Our theory shows that there is a restriction on the number of machines and this restriction becomes more stringent when the dimension $p$ is large. To overcome this limitation, this paper proposes a new  multi-round distributed estimation procedure that approximates the Newton step only using  stochastic subgradient. The key component in our method is the proposal of a computationally efficient estimator of $\S^{-1} \w$, where $\S$ is the population Hessian matrix and $\w$ is any given vector. Instead of estimating $\S$ (or $\S^{-1}$) that usually requires the second-order differentiability of the loss,  the proposed First-Order Newton-type Estimator (FONE) directly estimates the vector of interest $\S^{-1} \w$ as a whole and is applicable to non-differentiable losses.  Our estimator also facilitates the inference for the empirical risk minimizer. It turns out that the key term in the limiting covariance has the form of $\S^{-1} \w$, which can be estimated by FONE.
	\end{abstract}

\section{Introduction}
\label{sec:intro}

The development of modern technology has enabled data collection of unprecedented size, which poses new challenges to many statistical estimation and inference problems. 
First,  given $N$ samples with a very large $N$, a standard machine might not have enough memory to load the entire dataset all at once. Second, a deterministic optimization approach is computationally expensive. To address the storage and computation issues,  distributed computing methods, originated from computer science literature, has been recently introduced into statistics. A general distributed computing scheme  partitions the entire dataset into $L$ parts, 
and then loads each part into the memory  to compute a local estimator. The final estimator will be obtained via some communication and aggregation among local estimators. 

To  further accelerate the computation, we consider stochastic first-order methods (e.g., stochastic gradient/subgradient descent (SGD)), which have been widely adopted in practice. There are a few significant advantages of SGD. First, as a first-order method, it only requires the subgradient information. As compared to \emph{second-order Newton-type} approaches, it is not only computationally efficient and more scalable but also has a wider range of applications to problems where the empirical Hessian matrix does not exist (e.g., when the loss is non-smooth such as quantile regression). Second, a stochastic approach is usually more efficient than its deterministic counterpart. 
Although SGD has been widely studied in machine learning and optimization, using SGD for the purpose of statistical inference has not been sufficiently explored.

This paper studies a general statistical estimation and inference problem under the distributed computing setup. As we mentioned, to achieve an efficient computation, we restrict ourselves to the use of only stochastic subgradient information. In particular, consider a general statistical estimation problem in the following risk minimization form,
\begin{eqnarray}\label{eq:opt}
\tee^*=\argmin_{\tee \in \R^p} F(\tee):=\ep_{\xii \sim \Pi} f(\tee,\xii),
\end{eqnarray}
where $f(\cdot, \xii): \R^p \rightarrow \R$ is a convex loss function that can be non-differentiable (e.g., in quantile regression), and $\xii$ denotes the random sample from a probability distribution $\Pi$ (e.g., $\xii=(Y, \X)$ in a regression setup).  Our goal is to estimate $\tee^* \in \mathbb{R}^p$ under the \emph{diverging dimension case}, where the dimensionality $p$ is allowed to go to infinity as the sample size grows (but $p$ grows at a slower rate than the sample size). This regime is more challenging than the fixed $p$ case. On the other hand, since this work does not make any sparsity assumption,  the high dimensional setting where $p$ could be potentially larger than the sample size is beyond our scope. For the ease of illustration, we will use two motivating examples throughout the paper:  (1) logistic regression with a differentiable loss, and (2) quantile regression with a non-differentiable loss. 

Given $n$ \emph{i.i.d.} samples\footnote{With a slight abuse of notation, we use $n$ to denote either the sample size in  non-distributed settings or the local sample size of a single machine in distributed settings. 
} $\{\xii_i\}_{i=1}^n$, a traditional non-distributed approach for estimating $\tee^*$ is to minimize the empirical risk via a deterministic optimization:
\begin{equation}\label{eq:erm}
\widehat\tee=\argmin_{\tee \in \R^p} \frac{1}{n} \sum_{i=1}^n f(\tee, \xii_i).
\end{equation}
Moreover, let $g(\tee, \xii)$ be the gradient (when $f(\tee, \xii)$ is differentiable) or a subgradient (when $f(\tee, \xii)$ is non-differentiable) of $f(\tee, \xii)$ at $\tee$. In terms of \emph{statistical inference}, for many popular statistical models, the empirical risk minimizer (ERM) $\widehat\tee$ has an asymptotic normal distribution. That is, under some regularity conditions, for a fixed unit length vector $\w \in \R^p$, as $n,p \rightarrow \infty$,
\begin{equation}\label{eq:asy}
\frac{\sqrt{n}\w'(\hat{\tee}-\tee^*)}{\sqrt{\w'\S^{-1}\A\S^{-1}\w}}\rightarrow \mathcal{N}(0,1),
\end{equation}
where
\begin{equation}\label{eq:sigma}
\S:= \nabla_{\tee} \ep g(\tee, \xii) |_{\tee=\tee^*} \quad \A=\Cov(g(\tee^*,\xii))=\ep \left[g(\tee^*,\xii)g(\tee^*,\xii)'\right].
\end{equation}

Under this framework, the main goal of our paper is twofold:
\begin{enumerate}
	\item \textbf{Distributed estimation}: Develop a distributed stochastic first-order method for estimating $\tee^*$ in the case of diverging $p$, with the aim to achieve the best possible convergence rate (i.e., the rate of the pooled ERM estimator $\widehat\tee$). The method should be applicable to non-differentiable loss $f(\tee, \xii)$ and only requires the local strong convexity of $F(\tee)$ at $\tee=\tee^*$ (instead of the strong convexity of $F(\tee)$ for any $\tee$).
	\item  \textbf{Distributed inference}: Based on \eqref{eq:asy}, develop a consistent estimator of the limiting variance $\w'\S^{-1}\A\S^{-1}\w$ to facilitate the inference. 
\end{enumerate}

Let us first focus on the distributed estimation problem. We will first investigate the theoretical proprieties of a straightforward method that combines the stochastic subgradient descent (SGD)  and divide-and-conquer (DC) scheme and discuss the theoretical limitation of this method. To overcome the theoretical limitation, we propose a new method called the distributed First-Order Newton-type Estimator (FONE), where the key idea is to approximate the Newton step only using stochastic subgradient information in a distributed setting.

In a distributed setting, the divide-and-conquer (DC) strategy has been recently adopted in many statistical estimation problems (see, e.g., \cite{li2013statistical,chen2014split,HuangHuo:15, zhang2015divide,battey2015distributed,zhao2016partially,shi2016massive,banerjee2019divide,volgushev2017distributed,Fan17distributed}.
A standard DC approach estimates a local estimator for each local machine.
and then aggregates the local estimators to obtain the final estimator. Combining the idea of DC with the mini-batch SGD naturally leads to a divide-and-conquer SGD (DC-SGD) approach, where we run SGD on each local machine and then aggregate the obtained solutions by an averaging operation. 
In fact, DC-SGD is not the main focus/contribution of this paper. It has been an existing popular distributed algorithm in practice for a long time. Nevertheless, the theoretical property of DC-SGD with mini-batch in the diverging dimension case has not been fully understood yet. We first establish the theoretical properties of DC-SGD and explain its limitations, which better motivates our distributed estimator (see below).	
For DC-SGD to achieve the optimal convergence rate, the number of machines $L$ has to be $O(\sqrt{N/p})$ (see Section \ref{sec:theory_DC_SGD}), where $N$ is the total number of samples across $L$ machines. The condition could be restrictive when the number of machines is large but each local machine has a limited storage (e.g., in a large-scale sensor network). 
Moreover, as compared to the standard condition $L=O(\sqrt{N})$ in a fixed $p$ setting, the condition $L=O(\sqrt{N/p})$ becomes more stringent when $p$ diverges. In fact, this constraint is not only for the case of DC-SGD. Since the averaging only reduces the variance but not the bias term, all the results for the standard DC approach in the literature inevitably involve a constraint on the number of machines, which aims to make the variance the dominating term.

To relax this condition on $L$ and further improve the performance of DC-SGD,  this paper  proposes a new approach called distributed first-order Newton-type estimator, which successively refines the estimator by multi-round aggregations. The starting point of our approach is the Newton-type method based on a consistent initial estimator $\hat\tee_0$:

\begin{equation}\label{eq:one-step}
\widetilde{\tee}=\hat\tee_0- \S^{-1} \left(\frac{1}{n} \sum_{i=1}^n g(\hat\tee_0, \xii_i)\right),
\end{equation}
where $\S$ is the population Hessian matrix and $\left(\frac{1}{n} \sum_{i=1}^n g(\hat\tee_0, \xii_i)\right)$ is the subgradient vector. However, the estimation of $\S$ is not easy when $f$ is non-differentiable and the empirical Hessian matrix does not exist.

To address this issue, our key idea is that instead of estimating $\S$ and computing its inverse, we propose an  estimator of  $\S^{-1} \w \in \R^p$ for any given vector $\w \in \R^p$, which solves \eqref{eq:one-step} as a special case (with $\w=\frac{1}{n} \sum_{i=1}^n g(\hat\tee_0, \xii_i)$). In fact, the estimator of $\S^{-1} \w$ kills two birds with one stone: it not only constructs a Newton-type estimator of $\tee^*$ but also provides an estimator for the asymptotic variance in \eqref{eq:asy}, which facilitates the inference. In particular, the proposed FONE estimator of $\S^{-1} \w$ is an iterative procedure that only utilizes the mini-batches of subgradient to approximate the Newton step.  

\label{page:jordan}
It is also worthwhile noting that our method extends the recent work by \cite{jordan2016communication} and \cite{wang2017improved}, which approximates the Newton step by using local Hessian matrix computed on a single machine. However, to compute the local Hessian matrix, their method requires the second-order differentiability on the loss function and thus is not applicable to problems such as quantile regression. In contrast, our approach approximates the Newton step via stochastic subgradient and thus can handle the non-differentiability in the loss function. We also note that the idea of approximating Newton step has been applied to specific statistical learning problems, such as SVM \citep{wang2019distributed}, quantile regression \citep{chen2019quantile,chen2020distributed},  and PCA \citep{chen2020PCA}. This paper provides a general framework for both smooth and non-smooth loss functions. 
Moreover, our method subsumes a recently developed stochastic first-order  approach---stochastic variance reduced gradient (SVRG, see e.g., \cite{johnson2013accelerating,lee2017distributed,wang2017improved,li2018approximate} and references therein) as a special case. 
While SVRG requires $\w$ to be the averaged gradient and its theory only applies to strongly convex smooth loss functions, we allow a general $\w$ vector and non-smooth losses. 

Based on FONE, we further develop a multi-round distributed version of FONE which successively refines the estimator  and does not impose any strict condition on the number of machines $L$.  Theoretically, we show that for a smooth loss, when the number of rounds $K$ exceeds a constant threshold $K_0$, the obtained distributed FONE $\teedis$ achieves the optimal convergence rate.  For a non-smooth loss, such as quantile regression, our convergence rate only depends on the sample size of one local machine with the largest sub-sample size. This condition is weaker than the case of DC-SGD since the bottleneck in the convergence of DC-SGD is the local machine with the smallest sub-sample size.

We further apply the developed FONE to obtain a consistent estimator of the asymptotic variance in \eqref{eq:asy} for the purpose of inference.
Note that the term $\A$ can be easily estimated via replacing the expectation by its sample version. Instead of estimating $\S^{-1}$ in \eqref{eq:asy}, our method directly estimates $\S^{-1}\w$ for any fixed unit length vector $\w$ (please see Section~\ref{sec:unitw} for more details).

The remainder of the paper is organized as follows: 
Section \ref{sec:method_DC_SGD} describes the mini-batch SGD algorithm with diverging dimension and the DC-SGD estimator. We further propose FONE and distributed FONE in Section \ref{sec:fone}. Section \ref{sec:theory} presents the theoretical results. Section \ref{sec:unitw} discusses the application of FONE to the inference problem. In Section \ref{sec:exp}, we demonstrate the performance of the proposed estimators by simulation experiments and real data analysis, followed by conclusions in Section \ref{sec:conc}. Some additional theoretical results, technical proofs and additional experimental results are provided in Appendix.

In this paper, we denote the Euclidean norm for a vector $\x\in\R^p$ by $\|\x\|_2$, and we denote the spectral norm for a matrix $\X$ by $\|\X\|$. 
In addition, since the distributed estimation and inference usually involve quite a few notations, we briefly summarize them here. We use $N$, $L$, $n=N/L$, and $m$ to denote the total number of samples, the number of machines (or the number of data partitions), the sample size on each local machine (when evenly distributed), and the batch size for mini-batch SGD, respectively. When we discuss a problem in the classical single machine setting, we will also use $n$ to denote the sample size. We will use $\tee^*$, $\hat\tee$, and $\hat\tee_0$ to denote the minimizer of the popular risk, the ERM, and the initial estimator, respectively. The random sample will be denoted by $\xii$ and in a regression setting $\xii=(Y, \X)$.

\section{Methodology}
\label{sec:method}

In this section, we first introduce the standard DC-SGD algorithm. The main purpose of introducing this DC-SGD algorithm is to better motivate the proposed FONE and its distributed version.

\subsection{Divide-and-conquer SGD (DC-SGD) algorithm}
\label{sec:method_DC_SGD}

Before we introduce our DC-SGD algorithm, we first present the mini-batch SGD algorithm for solving the stochastic optimization in \eqref{eq:opt} on a single machine with total $n$ samples. In particular, we consider the setting when the dimension $p\rightarrow \infty$ but at a slower rate than $n$, i.e., $p \leq n^{\kappa}$ for some $\kappa \in (0,1)$. Given $n$ \emph{i.i.d.} samples $\{\xii_1, \ldots, \xii_n\}$, we partition the index set $\{1,\ldots, n\}$ into $s$ disjoint mini-batches $H_{1},...,H_{s}$, where each mini-batch has the size $|H_{i}|=m$ (for $i=1,2,\dots, s$), and $s=n/m$ is the number of mini-batches. The mini-batch SGD algorithm starts from a consistent initial estimator $\hat\tee_0$ of $\tee^*$. Let $\z_0=\hat\tee_0$. The mini-batch SGD iteratively updates $\z_i$ from $\z_{i-1}$ as follows and outputs $\teeSGD=\z_s$ as its final estimator,
\begin{eqnarray}\label{eq:SGD_mini}
\z_{i}=\z_{i-1}-\frac{r_{i}}{m}\sum_{j\in H_{i}}g(\z_{i-1},\xii_j),  \quad \text{for} \;\;\; i=1,2,\ldots, s,
\end{eqnarray}
where we set the step-size $r_{i}=c_{0}/\max(i^{\alpha},p)$ for some $0<\alpha\leq 1$ and $c_{0}$ is a positive constant. It is worthwhile that a typical choice of $r_i$  in the literature is $r_i=c_0 \cdot {i}^{-\alpha}$  \citep{polyak1992acceleration,chen2020statistical}.  Since we are considering a diverging $p$ case, our step-size incorporates the dimension $p$.  As one can see, this mini-batch SGD algorithm only uses one pass of the data and enjoys a low per-iteration complexity.

We  provide two examples on logistic regression and quantile regression to illustrate the subgradient function $g(\tee, \xii)$ in our mini-batch SGD and will refer to these  examples throughout the paper.

\begin{example}[Logistic regression]\label{ex:log}
	Consider a logistic regression model with the response $Y \in \{-1,1\}$, where
	\begin{eqnarray*}
		\pr(Y=1|\X)=1-\pr(Y=-1|\X)=\frac{1}{1+\mathrm{exp}(-\X'\tee^*)},
	\end{eqnarray*}
	and $\tee^*\in \R^p$ is the true model parameter. Define  $\xii=(Y,\X)$. We have the smooth loss function $f(\tee,\xii)=\log(1+\exp(-Y\X'\tee))$ and its gradient $g(\tee,\xii)=-Y\X\big(1+\exp(Y\X'\tee)\big)^{-1}$.
\end{example}

\begin{example}[Quantile regression]\label{ex:qr}
	Consider a quantile regression model
	$
	Y=\X'\tee^*+\epsilon,
	$
	where we assume that $\X=(1,X_{1},...,X_{p-1})'$  and $\pr(\epsilon\leq 0|\X)=\tau$ is the so-called quantile level.  Define  $\xii=(Y,\X)$. We have the non-smooth quantile loss function $f(\tee,\xii)=\ell_{\tau}(Y-\X'\tee)$ and $\ell_{\tau}(x)=x(\tau-I\{x\leq 0\})$. A subgradient of the quantile loss is given by $g(\tee,\xii)=\X(I\{Y\leq \X'\tee\}-\tau)$.
\end{example}

The bias and $L_2$-estimation error of the mini-batch SGD will be provided in Theorem B.1 (see Appendix B). In particular, in the diverging dimension setting, it is necessary to have a consistent initial estimator to guarantee the consistency of obtained solution from the mini-batch SGD (see Proposition B.2 in Appendix).

Given the mini-batch SGD, we are ready to introduce the divide-and-conquer SGD (DC-SGD). For the ease of illustration, suppose that the entire sample with the size $N$  is evenly distributed on $L$ machines (or split into $L$ parts) with the sub-sample size $n=N/L$ on each local machine. For the ease of presentation, we assume that $N/L$ is a positive integer. On each machine $k=1,2,\dots,L$, we run the mini-batch SGD with the batch size $m$  in \eqref{eq:SGD_mini}. Let $\mathcal{H}_k$ be the indices of the data points on the $k$-th machine, which is further split into $s$ mini-batches $\{H_{k,i}, i=1,2,\dots, s\}$ with $|H_{k,i}|=m$ and $s=n/m$. On the $k$-th machine, we run our mini-batch SGD in \eqref{eq:SGD_mini}  and obtain the local estimator $\teeSGD^{(k)}$. The final estimator is aggregated by averaging the local estimators from $L$ machines, i.e.,
\begin{eqnarray}\label{eq:dcsgd}
\teeDC=\dfrac{1}{L}\sum\limits_{k=1}^L \teeSGD^{(k)}.
\end{eqnarray}
Note that the DC-SGD algorithm only involves one round of aggregation. The details of the  DC-SGD are presented in Algorithm \ref{algo:dc-sgd}.

\renewcommand{\baselinestretch}{1.1}
\begin{algorithm}[!t]
	\caption{ DC-SGD algorithm}
	\label{algo:dc-sgd}
	{\textbf{Input:}  The initial estimator $\hat\tee_0\in\mathbb{R}^p$, the step-size sequence $r_i=c_{0}/\max(i^{\alpha},p)$ for some $0<\alpha\leq 1$, the mini-batch size $m$.}
		\begin{algorithmic}[1]
		\STATE Distribute the initial estimator $\hat\tee_0$ to each local machine $k=1,2,\dots, L$.
		\FOR{each local machine $k=1,2, \dots, L$}
		\STATE Set the starting point $\z_0^{(k)}=\hat\tee_0$.
		\FOR{each iteration $i=1,\dots, s$}
		\STATE   Update \begin{equation*}
			\z_{i}^{(k)}=\z_{i-1}^{(k)}-\frac{r_{i}}{m}\sum_{j\in H_{k,i}}g(\z_{i-1}^{(k)},\xii_j),
		\end{equation*}
		\ENDFOR 
		\STATE Set $\teeSGD^{(k)}=\z_s^{(k)}$ as the local SGD estimator on the machine $k$.
		\ENDFOR
		\STATE Aggregate the local estimators $\teeSGD^{(k)}$ by averaging and compute the final estimator: \begin{equation*}
		\teeDC=\dfrac{1}{L}\sum\limits_{k=1}^L \teeSGD^{(k)}.
		\end{equation*}
		\STATE \textbf{Output:} $\teeDC$.
	\end{algorithmic}
	\vspace{-.1cm}
\end{algorithm}

\renewcommand{\baselinestretch}{1.43}
In Theorem \ref{th_dc}, we establish the convergence rate of the DC-SGD in terms of the dimension $p$, the number of machines $L$, the total sample size $N$ and the mini-batch size $m$. Moreover, we show that for the DC-SGD to achieve the same rate as the mini-batch SGD running on the entire dataset, it requires a condition on the number of machines $L$. 
This condition is essential because the averaging scheme in a divide-and-conquer approach only reduces the variance but not the bias term.

\subsection{First-Order Newton-type Estimator (FONE)}
\label{sec:fone}

\renewcommand{\baselinestretch}{1.1}
\begin{algorithm}[!t]
	\caption{First-Order Newton-type Estimator (FONE) of  $\S^{-1}\a$}
	\label{algo:fone}
	{\textbf{Input:} Dataset $\{\xii_1,\xii_2,\dots, \xii_n\}$, the initial estimator $\hat\tee_0$, step-size $\eta$, the batch-size $m$, and a given vector $\a\in\mathbb{R}^p$.\\} 
	\begin{algorithmic}[1]
		\STATE Set $\z_{0}=\hat\tee_{0}$.
		\FOR{each $t=1,2,\dots, T$}
		\STATE  Choose $B_t$ to be $m$ distinct elements uniformly from $\{1,2,...,n\}$.
		\STATE   Calculate
		\[
		g_{B_{t}}(\z_{t-1})=\frac{1}{m}\sum_{i\in B_{t}}g(\z_{t-1},\xii_i), \qquad g_{B_{t}}(\z_0)=\frac{1}{m}\sum_{i\in B_{t}}g(\z_0,\xii_i).
		\]
		\STATE Update
		\begin{equation*}
			\z_{t}=\z_{t-1}-\eta\{g_{B_{t}}(\z_{t-1})-g_{B_{t}}(\z_0)+\a\}.
		\end{equation*}
		\ENDFOR
		\STATE \textbf{Output:} 
		\begin{equation}\label{eq:fone_est}
		\teeFONE=\hat\tee_0-\z_T.
		\end{equation}
	\end{algorithmic}
	\vspace{-.45em}
\end{algorithm}

\renewcommand{\baselinestretch}{1.43}
To relax the condition on the number of machines $L$, one idea is to perform a Newton-type step in \eqref{eq:one-step}. However, as we have pointed out, the estimation of $\S$ requires the second-order differentiability of the loss function. Moreover, a typical Newton method successively refines the estimator of $\S$ based on the current estimate of $\tee^*$ and thus requires the computation of  matrix inversion in \eqref{eq:one-step} for multiple iterations, which could be computationally expensive when $p$ is large.

In this section, we propose a new First-Order Newton-type Estimator that directly estimates $\S^{-1} \a$ (for any given vector $\a$) only using the stochastic first-order information. Then for a given initial estimator $\hat\tee_0$, we can perform the Newton-type step in \eqref{eq:one-step} as
\begin{equation}\label{eq:one-step-1}
\tilde\tee=\hat\tee_0- \widehat{\S^{-1} \a }, \qquad  \a=\left(\frac{1}{n} \sum_{i=1}^n g(\hat\tee_0, \xii_i)\right),
\end{equation}
where $\widehat{\S^{-1} \a }$ is our estimator of  $\S^{-1}\a$.

To estimate $\S^{-1}\a$, we note that $\S^{-1}\a=\sum\limits_{i=0}^\infty (	I-\eta\S)^i\eta\a$ for small enough $\eta$ such that $\|\eta\S\|<1$. Then we can use the following iterative procedure $\{\tilde\z_t\}$ to approximate $\S^{-1}\a$:
\begin{eqnarray}\label{eq:wt}
\tilde\z_t=\tilde\z_{t-1}-\eta(\S\tilde\z_{t-1}-\a),\ 1\leq t\leq T,
\end{eqnarray}
where  $\eta$ here can be viewed as a constant step-size. To see that \eqref{eq:wt} leads to an approximation of $\S^{-1}\a$, when $T$ is large enough, we have
\begin{eqnarray*}
	\tilde\z_T&=&\tilde\z_{T-1}-\eta(\S\tilde\z_{T-1}-\a)=(I-\eta\S)\tilde\z_{T-1}+\eta\a\cr
	&=&(I-\eta\S)^2\tilde\z_{T-2}+(I-\eta\S)\eta\a+\eta\a\cr
	&=&(I-\eta\S)^{T-1}\tilde\z_1+\sum_{i=0}^{T-2}(I-\eta\S)^{i}\eta\a\approx \S^{-1}\a.
\end{eqnarray*}
As the iterate $\tilde\z_t$ approximates  $\S^{-1}\a$, let us define $\z_t=\hat\tee_0-\tilde\z_t$, which is the quantity of interest (see the left-hand side of
the Newton-type step in \eqref{eq:one-step-1}). To avoid estimating $\S$ in the recursive update in \eqref{eq:wt}, we adopt the following first-order approximation:
\begin{equation}\label{eq:first-approx}
-\S\tilde\z_{t-1}=\S(\z_{t-1}-\hat\tee_0)\approx g_{B_t}(\z_{t-1})-g_{B_t}(\hat\tee_0),
\end{equation}
where $g_{B_t}(\tee)=\frac{1}{m}\sum_{i\in B_t}g(\tee,\xii_i)$ is the averaged stochastic subgradient over a subset of the data indexed by $B_t \subseteq \{1,2,\ldots,n\}$. Here $B_t$ is randomly chosen from $\{1,\ldots, n\}$ with replacement for every iteration.

Given \eqref{eq:first-approx}, we construct our FONE of $\hat\tee_0- \S^{-1} \a$ by the following recursive update from $t=1,2,\dots, T$:
\begin{eqnarray}\label{eq:fone}
\z_{t}=\z_{t-1}-\eta\{g_{B_{t}}(\z_{t-1})-g_{B_{t}}(\hat\tee_0)+\a\}, \quad \z_{0}=\hat\tee_{0}.
\end{eqnarray}
The obtained $\z_T$, as an estimator of $\hat\tee_0- \S^{-1} \a$ can be directly used in the Newton-type step in \eqref{eq:one-step-1}.  The choices of the input parameters and the convergence rate of our FONE will be proved in Propositions \ref{prop:fone} and \ref{prop:fone_non}. Also note that for constructing the estimator of $\S^{-1} \a$, we can simply use $\hat\tee_0-\z_T$ and the procedure is summarized in Algorithm \ref{algo:fone}.

\renewcommand{\baselinestretch}{1.1}
\begin{algorithm}[!t]
	\caption{Distributed FONE for Estimating $\tee^*$ in \eqref{eq:opt}}
	\label{algo:distributed_FONE}
	{\textbf{Input:}  The total sample size $N$, the entire data $\{\xii_1,\xii_2,\dots, \xii_N\}$ is distributed into $L$ machines/parts $\{\mathcal{H}_k\}$ for $k=1,2\ldots, L$ with $|\mathcal{H}_k|=n_k$. Initial estimator $\hat\tee_0\in\mathbb{R}^p$, the batch size $m$, step-size $\eta$. Number of rounds $K$.} \vspace{0cm}
	\begin{algorithmic}[1]
		\FOR{each round $j=1,2,\dots, K$}
		\FOR{each local machine $k=1,2,\dots, L$}
		\STATE Calculate $\sum\limits_{i\in\mathcal{H}_k}g(\hat{\tee}_{j-1},\xii_{i}).$
		\ENDFOR
		\STATE Collect  $\sum\limits_{i\in\mathcal{H}_k}g(\hat{\tee}_{j-1},\xii_{i})$ from each local machine to compute their average:
		\begin{equation}\label{eq:a_dist}
		\a=\frac1N\sum\limits_{k=1}^L\sum\limits_{i\in\mathcal{H}_k}g(\hat{\tee}_{j-1},\xii_{i})= \frac1N\sum\limits_{i=1}^Ng(\hat{\tee}_{j-1},\xii_{i}).
		\end{equation}
		\STATE Send $\a$ to the first machine (the local machine with the largest sub-sample size).
		\STATE Set $\z_0=\hat\tee_{j-1}$
		\FOR{each $t=1,2,\dots, T$}
		\STATE  Choose $B_t$ to be $m$ distinct elements uniformly drawn from the data on the first machine $\mathcal{H}_1$.
		\STATE Calculate
		\[
		g_{B_{t}}(\z_{t-1})=\frac{1}{m}\sum_{i\in B_{t}}g(\z_{t-1},\xii_i), \qquad g_{B_{t}}(\z_0)=\frac{1}{m}\sum_{i\in B_{t}}g(\z_0,\xii_i).
		\]
		\STATE Update
		\begin{equation}\label{eq:dis_fone}
		\z_{t}=\z_{t-1}-\eta\{g_{B_{t}}(\z_{t-1})-g_{B_{t}}(\z_0)+\a\}.
		\end{equation}
		\ENDFOR
		\STATE Set $\widehat{\tee}_{j}=\z_T$.
		\ENDFOR
		\STATE \textbf{Output:} $\teedis=\hat\tee_K$.
	\end{algorithmic}
\vspace{-.25em}
\end{algorithm}

\renewcommand{\baselinestretch}{1.43}

\subsection{Distributed FONE for estimating $\tee^*$}
\label{sec:dist_FONE}
Based on the FONE for $\S^{-1} \a$, we present a distributed FONE for estimating $\tee^*$. 
Suppose the entire dataset with $N$ samples is distributed on $L$ local machines $\{\mathcal{H}_1,\mathcal{H}_2,\dots,\mathcal{H}_L\}$ (not necessarily evenly distributed). 
Our distributed FONE is a multi-round approach with $K$ rounds, where $K$ is a pre-specified constant. For each round $j=1,2,\dots, K$, with the initialization $\hat\tee_{j-1}$, we first calculate $\a=\frac1N\sum_{i=1}^N g(\hat\tee_{j-1},\xii_i)$ by averaging the subgradients from each local machine. Then we apply FONE (Algorithm \ref{algo:fone}) with $\a$ on the local machine with the largest sub-sample size. Since FONE is performed on one local machine, this iterative procedure does not incur any extra communication cost.
The detailed algorithm is given in Algorithm \ref{algo:distributed_FONE}.    In fact, the presented Algorithm  \ref{algo:distributed_FONE} is essentially estimating $\hat\tee_0-\S^{-1} \a$ with $\a=\frac{1}{N} \sum_{i=1}^N g(\hat\tee_0, \xii_i)$ and $\hat\tee_0$ is a pre-given initial estimator.

It is worthwhile noting that in contrast to DC-SGD where each local machine plays the same role, distributed FONE performs the update in \eqref{eq:dis_fone} only on one local machine. The convergence rate of distributed FONE will depend on the sub-sample size of this machine (see Theorems \ref{thm:d-fone} and \ref{thm:d-fone-non}). Therefore, to achieve the best convergence rate, we perform the update in \eqref{eq:dis_fone} on the machine with the largest sub-sample size and \emph{index it by the first machine} without loss of generality.  We also note that since the first machine collects the gradient and performs FONE, the distributed FONE can be easily implemented in a de-centralized setting.

Instead of using the first machine to compute FONE, one can further leverage the idea of the divide-and-conquer (DC) in  the distributed FONE. In particular, one may let each local machine run FONE based on the aggregated gradient $\a$ simultaneously, and then take the average of all the local estimators on $L$ machines as $\hat{\tee}_{j}$ for the $j$-th round.

Finally, we make a brief comment on the communication cost of distributed FONE. First, both the distributed FONE and DC-SGD are transmitting $p$-dimensional vectors from the local machines, which are usually considered as \emph{communication-efficient} distributed protocols in the literature.  More precisely, DC-SGD only has one round communication and thus the communication cost on each local machine is $O(p)$. For distributed FONE, which requires $K$ rounds of communication, the total communication cost on each local machine is $O(Kp)$. However, we note that from Theorems \ref{thm:d-fone} and \ref{thm:d-fone-non} in our theoretical results below, the number of rounds $K$ only needs be a constant (instead of diverging to infinity). Therefore, the communication of the distributed FONE is on the same (asymptotic) order as DC-SGD.

\section{Theoretical Results}\label{sec:theory}

In this section, we provide theoretical results for mini-batch SGD in the diverging $p$ case, DC-SGD, the newly proposed FONE and its distributed version.  We first note that in most cases,  
the minimizer $\tee^*$ in \eqref{eq:opt} is also a solution of the following estimating equation:
\begin{equation}\label{eq:SA}
\ep g(\tee^*, \xii)=0,
\end{equation}
where $g(\tee, \xii)$ is the gradient or a subgradient of $f(\tee, \xii)$ at $\tee$. We will assume that \eqref{eq:SA} holds throughout our paper. In fact, we can introduce \eqref{eq:SA} as our basic model (instead of \eqref{eq:opt}) as in the literature from stochastic approximation (see, e.g., \cite{lai2003stochastic}). However, we choose to present the minimization form in  \eqref{eq:opt} as it is more commonly used in statistical learning literature.

Now let us first establish the theory for the DC-SGD approach in the diverging $p$ case.

\subsection{Theory for DC-SGD}\label{sec:th-sgd}

To establish the theory for DC-SGD, we first state our assumptions.  The first assumption is on the relationship among the dimension $p$, the sample size $n$, and the mini-batch size $m$.
Recall that $\alpha$ is the decaying rate in the step-size of SGD (see the input of Algorithm \ref{algo:dc-sgd}).\vspace{2mm}

\noindent{\bf \tt (A1).} Suppose that  $p\rightarrow \infty$ and $p=O(n^{\kappa_1})$ for some $0<\kappa_1<1$. The mini-batch size $m$ satisfies $p\log n=o(m)$ and $n^{\tau_{1}}\leq m\leq n/p^{1/\alpha+\tau_{2}}$ for some $0<\tau_{1}, \tau_{2}<1$. 

\vspace{2mm}

Throughout this paper, we define a loss function $f$ to be smooth when $f$ is continuously differentiable (e.g., logistic regression), and non-smooth when $f$ is non-differentiable $f(\tee, \xii)$ (e.g., quantile regression). 

We give two separate conditions for smooth and non-smooth loss functions, respectively. To simplify the illustration, we only present Conditions {\tt(A2-Log)} and {\tt(A2-QR)} for the two representative examples of smooth and non-smooth loss functions. In particular, Condition {\tt(A2-Log)} applies to logistic regression in Example \ref{ex:log}, and Condition {\tt(A2-QR)} applies to the quantile regression in Example \ref{ex:qr}. In Appendix A.1, we provide very general conditions for smooth and non-smooth loss functions, which are not necessarily restricted to the regression case. We will also verify that the logistic and quantile regressions satisfy our general conditions.

\vspace{2mm}
\noindent {\bf \tt(A2-Log)}. (For logistic regression in Example \ref{ex:log}.) Assume that 
\[
{c}_{1}\leq \lambda_{\min}(\ep(\X\X^{'}))\leq \lambda_{\max}(\ep(\X\X^{'}))\leq {c}^{-1}_{1},
\]
and $\|\tee^*\|_2\leq C_1$, $\sup_{\|\v\|_{2}=1}\ep \exp(t_{0} (\v^{'}\X)^{2})\leq C_2$  for some $c_1, t_0, {C}_{1}, C_2>0$.

\noindent {\bf\tt(A2-QR)}.  (For quantile regression in Example \ref{ex:qr}.)  Assume that $\epsilon|\X$ has density function $\rho_{\X}$ that is bounded and satisfies $|\rho_{\X}(x_{1})-\rho_{\X}(x_{2})|\leq C_1|x_{1}-x_{2}|$ for some $C_1>0$. Moreover,
	\[
	c_{1}\leq \lambda_{\min}(\ep[\X\X'\rho_{\X}(0)])\leq \lambda_{\max}(\ep[\X\X'\rho_{\X}(0)])\leq c^{-1}_{1}
	\]
	and $\sup_{\|\v\|_{2}=1}\ep \exp(t_{0} |\v^{'}\X|)\leq C_2$  for some $c_1, t_0, {C}_{2}>0$. 

Conditions {\tt(A2-Log)} and {\tt(A2-QR)} are standard regularity conditions (i.e., covariance and moment conditions) on the covariates of a regression model.  The minimum eigenvalue conditions ensure that the population risk $F(\tee)$ is locally strongly convex at $\tee=\tee^*$.
We note that the general version of these two conditions (A2-Log) and (A2-QR) for arbitrary loss functions  $f(\tee, \xii)$  will be provided in Appendix A.1.

\medskip

\label{sec:theory_DC_SGD}
Due to the space constraint, we introduce the theory of mini-batch SGD in Appendix B. Despite the simplicity and wide applicability of the mini-batch SGD, the theoretical investigation of the asymptotic properties of this approach, especially in the diverging $p$ case, is still quite limited. In fact, our theoretical analysis reveals several interesting phenomena of the mini-batch SGD when $p$ is diverging, which also leads to useful practical guidelines when implementing mini-batch SGD. A natural  starting point in a standard mini-batch SGD is  random initialization. However, we show that when $p$ diverges to infinity, a random initialized SGD will no longer converge to $\tee^*$,  with the $L_2$-estimation error being a polynomial of $p$ (see Proposition B.2 in Appendix). To address the challenge arising from $p \rightarrow \infty$, a consistent initial estimator $\hat\tee_0$ is both sufficient and necessary to ensure the convergence of SGD (see Theorem B.1 and Proposition B.2 in Appendix).  Since DC-SGD is built on the mini-batch SGD, a consistent initialization is also required in DC-SGD, which can be easily constructed by running a deterministic optimization on a small batch of data. Given that, we provide the convergence result of the DC-SGD estimator $\teeDC$ in \eqref{eq:dcsgd} (see Algorithm \ref{algo:dc-sgd}). For the ease of presentation, we assume that the data are evenly distributed, where each local machine has $n=N/L$ samples.

\begin{theorem} \label{th_dc} Assume Conditions {\tt(A1)} and {\tt(A2-Log)} or {\tt(A2-QR)} hold, suppose the initial estimator $\hat\tee_0$ is independent to $\{\xii_i, i=1,2,\dots, N\}$.
	On the event $\{\|\hat\tee_{0}-\tee^*\|_2\leq d_n\}$ with $d_n\rightarrow 0$, the DC-SGD estimator achieves the following convergence rate:
	\begin{equation}\label{eq:DC_SGD_MSE}
	\ep_{0}\|\teeDC-\tee^*\|_2^{2}=O\left(\frac{p}{L^{1-\alpha}m^{1-\alpha}N^{\alpha}}  +\frac{p^{2}L^{2\alpha}}{m^{2-2\alpha}N^{2\alpha}}\right).
	\end{equation}
\end{theorem}

Again, we note that throughout the theoretical results,  Condition {\tt(A2-Log)} can be generalized to Conditions {\tt(C2)} and {\tt(C3)} in Appendix (and correspondingly {\tt(A2-QR)} to {\tt(C2)} and {\tt(C$3^*$)}).

The convergence rate in \eqref{eq:DC_SGD_MSE} contains two terms. The first term comes from the variance of the DC-SGD estimator, while the second one comes from the squared bias term.  Note that $n=N/L$, the squared bias term in \eqref{eq:DC_SGD_MSE} can be written as $\left(\frac{p}{m^{1-\alpha}n^\alpha}\right)^2$, which is the same as the square of the bias from the mini-batch SGD on one machine (see Theorem B.1 in the supplementary material). This is because the averaging of the local estimators from $L$ machines cannot reduce the bias term. On the other hand, the variance term is reduced by a factor of $1/L$ by averaging over $L$ machines. Therefore, when $L$ is not too large, the variance will become the dominating term and gives the optimal convergence rate. An upper bound on $L$ is a universal condition in the divide-and-conquer (DC) scheme to achieve the optimal rate in a statistical estimation problem (see, e.g., \cite{li2013statistical, chen2014split, zhang2015divide,HuangHuo:15, battey2015distributed,  zhao2016partially, lee2017communication, volgushev2017distributed}).  In particular, let us consider the optimal step-size $r_{i}$ where $\alpha=1$. When the number of machines $L=O(\sqrt{N/p})$, the rate in (\ref{eq:DC_SGD_MSE}) becomes $O(p/N)$, which is a classical optimal rate when using all the $N$ samples.

We next show on the two motivating examples that the constraint on the number of machines $L=O(\sqrt{N/p})$ is necessary to achieve the optimal rate by DC-SGD.  To this end, we provide the lower bounds on our two examples for the bias of the SGD estimator on each local machine.

 \begin{customexp}{\ref{ex:log} (Continued)}  For a logistic regression model with $\xii=(Y, \X)$, let $\X=(1,X_{1},...,X_{p-1})'$ with $\ep X_{i}=0$ for all $1\leq i\leq p-1$	and  $\tee^*=(1,0,...,0)$. Suppose that $\ep\|\X\|^{2}_{2}\geq cp$ for some $c>0$ and 
	$\sup_{\|\v\|_{2}=1}\ep \exp(t_{0}|\v'\X|)\leq C$. Suppose the initial estimator $\hat\tee_0$ is independent to $\{\X_i, i=1,2,\dots, n\}$.
	On the event $\{\|\hat\tee_{0}-\tee^*\|_2\leq d_n\}$ with $d_n\rightarrow 0$, we have $\|\ep_{0}(\teeSGD)-\tee^*\|_2\geq \frac{cp}{m^{1-\alpha}n^{\alpha}}$.
\end{customexp}
\begin{customexp}{\ref{ex:qr} (Continued)}  For a quantile regression model, assume that $\epsilon$ is independent with $\X$ and $\ep X_{i}=0$ for all $1\leq i\leq p-1$. Let $F(x)$ be the cumulative distribution function of $\epsilon$. Suppose that $\ep\|\X\|^{2}_{2}\geq cp$ for some $c>0$ and 
	$\sup_{\|\v\|_{2}=1}\ep \exp(t_{0}|\v'\X|)\leq C$. 
	Suppose the initial estimator $\hat\tee_0$ is independent to $\{\X_i, i=1,2,\dots, n\}$, and assume that $F(\cdot)$ has bounded third-order derivatives and $F'(0)$, $F''(0)$ are positive. On the event $\{\|\hat\tee_{0}-\tee^*\|_2\leq d_n\}$ with $d_n\rightarrow 0$, we have $\|\ep_{0}(\teeSGD)-\tee^*\|_2\geq \frac{cp}{m^{1-\alpha}n^{\alpha}}$.
\end{customexp}

For the DC-SGD estimator $\hat\tee_{\text{DC}}$, it is easy to see that the mean squared error
$
\ep_{0}\|\teeDC-\tee^*\|_2^{2}\geq \|\ep_{0}(\hat{\tee}_{\text{DC}})-\tee^{*}\|^{2}_{2}$ (the squared bias of $\hat{\tee}_{\text{DC}}$). Recall that the bias of $\hat{\tee}_{\text{DC}}$ is the average over local machines, and each local machine induces the same bias $\|\ep_{0}(\hat{\tee}_{\text{SGD}})-\tee^{*}\|_{2}$ (see the bias in the above two examples). Therefore, for logistic regression and quantile regression, when $\alpha=1$ and $\sqrt{N/p}=o(L)$ , we have
\begin{eqnarray*}
	\frac{\ep_{0}\|\teeDC-\tee^*\|_2^{2}}{{p/N}}\geq\frac{\|\ep_{0}(\hat{\tee}_{\text{SGD}})-\tee^{*}\|^{2}_{2}}{p/N}\geq \dfrac{c^2p^2/n^2}{p/N}=c^2\frac{L^2}{N/p}\rightarrow \infty.
\end{eqnarray*}
This shows that when the number of machines $L$ is much larger than $\sqrt{N/p}$, the convergence rate of DC-SGD will no longer be  optimal.

\begin{remark}\label{rem:asgd} 
It is worthwhile noting that convergence rate of stochastic gradient estimators can be improved by the use of averaging \citep{polyak1992acceleration}. In particular, given the SGD iterates $\{\z_i\}_{i=1}^s$ in \eqref{eq:SGD_mini}, an average stochastic gradient (ASGD) algorithm outputs $\hat\tee_{\mathrm{ASGD}}=\frac1s\sum_{i=1}^s\z_i$ instead of $\teeSGD=\z_s$. ASGD is known to achieve a faster convergence rate than SGD when $\alpha<1$. Similar to $\teeDC$ in \eqref{eq:dcsgd}, we may implement the divide-and-conquer scheme on ASGD estimator, denoted by $\frac{1}{L}\sum\limits_{k=1}^L \hat\tee_{\mathrm{ASGD}}^{(k)}$. Assuming the mini-batch size $m=1$, we have
	\[
	\ep_{0}\|\hat\tee_{\mathrm{DC-ASGD}}-\tee^*\|_2^{2}=O\left(\frac{p}{N}+\frac{p^{2}L^{2}}{N^{2}}\right).
\]
  As compared to the convergence rate of DC-SGD in Theorem \ref{th_dc}, when the exponent in the stepsize $\alpha<1$, the convergence rate of DC-ASGD is faster than the that of DC-SGD. In other words, the averaging idea indeed accelerates the DC-SGD approach. Nevertheless, the DC-ASGD estimator requires the same condition as $\teeDC$ on the number of machines (i.e., $L=O(\sqrt{N/p})$) to achieve optimal rate. 
\end{remark}

\subsection{Theory for First-order Newton-type Estimator (FONE)}\label{sec:FONE}

We provide our main theoretical results on FONE for estimating $\S^{-1}\a$ and the distributed FONE for estimating $\tee^*$. The smooth loss and non-smooth loss functions are discussed separately in Section \ref{sec:smooth} and Section \ref{sec:non-smooth}. 

Recall that $n$ denotes the sample size used in FONE in the single machine setting (see Algorithm \ref{algo:fone}). In our theoretical results, we denote the step-size in FONE by $\eta_n$ (instead of $\eta$ in Algorithms \ref{algo:fone} and \ref{algo:distributed_FONE}) to highlight the dependence of the step-size on $n$.  For the FONE method, Condition {\tt(A1)} can be further weakened to the following condition:

\vspace{2mm}

\noindent{\bf\tt (A1$^{*}$).} Suppose that  $p\rightarrow \infty$ and $p=O(n^{\kappa_1})$ for some $0<\kappa_1<1$. The mini-batch size $m$ satisfies $p\log n=o(m)$ with $m=O(n^{\kappa_{2}})$ for some $0<\kappa_{2}<1$. 

\subsubsection{Smooth loss function $f$}
\label{sec:smooth}

To establish the convergence rate of our distributed FONE, we first provide a consistency result for $\teeFONE$ in \eqref{eq:fone_est}.
\begin{proposition}[On $\teeFONE$ for $\S^{-1} \a$ for smooth loss function $f$]\label{prop:fone}
	Assume  Conditions {\tt(A1$^{*}$)} and {\tt(A2-Log)} (or  Conditions {\tt(C2)} and {\tt(C3)} in the supplement)  hold. Suppose that the initial estimator satisfies $\|\hat{\tee}_{0}-\tee^{*}\|_{2}=O_{\pr}(d_n)$, and $\|\a\|_2=O(\tau_{n})$ (or $O_{\pr}(\tau_{n})$ for the random case). The iteration number $T$ and step-size $\eta_{n}$ satisfy $\log n=o(\eta_{n}T)$ and $T=O(n^{A})$ for some $A>0$. We have
	\begin{equation}\label{es1}
	\|\teeFONE-\S^{-1}\a\|_2=O_{\pr}\big{(}\tau_{n}d_n+\tau^{2}_{n}+\sqrt{\frac{p\log n}{n}}\tau_{n}+\sqrt{\eta_{n}}\tau_{n}+n^{-\gamma}\big{)}
	\end{equation}
	for any large $\gamma>0$. 
\end{proposition}

The relationship between $\eta_n$ and $T$ (i.e., $\log n=o(\eta_{n}T)$) is intuitive since when the step-size $\eta_n$ is small, Algorithm \ref{algo:fone} requires more iterations $T$ to converge.  
The consistency of the estimator requires that the length of the vector $\a$ goes to zero, i.e., $\tau_n=o(1)$, since $\tau_n^2$ appears in the convergence rate in \eqref{es1}. In Section \ref{sec:unitw}, we  further discuss a slightly modified FONE that deals with any vector $\a$, which applies to the estimation of the limiting variance of $\hat\tee$ in \eqref{eq:asy}.  When  $\tau_n=o(1)$, $d_n=o(1)$, and $\eta_n=o(1)$, each term in $O_{\pr}$ in \eqref{es1} goes to zero and thus the proposition guarantees that $\teeFONE$ is a consistent estimator of $\S^{-1}\a$. Moreover, since Proposition \ref{prop:fone} will be used as an intermediate step for establishing the convergence rate of the distributed FONE,  to facilitate the ease of use of Proposition \ref{prop:fone}, we leave $d_n$, $\tau_n$, and $\eta_n$ unspecified here and discuss their magnitudes in Theorem \ref{thm:d-fone}. A practical choice of $\eta_n$ is further discussed in the experimental section.

Given Proposition \ref{prop:fone}, we now provide the convergence result for the multi-round distributed FONE for estimating $\tee^*$ and approximating $\hat{\tee}$  in Algorithm \ref{algo:distributed_FONE}. To this end, let us first provide some intuitions on the improvement for one-round distributed FONE from the initial estimator $\hat \tee_0$  to $\hat \tee_{\mathrm{dis},1}$. For the first round in  Algorithm \ref{algo:distributed_FONE}, the algorithm essentially estimates $\S^{-1} \a$ with $\a= \frac1N\sum\limits_{i=1}^N g(\hat\tee_0,\xii_i)$. When $f(\tee,\xii)$ is  differentiable and noting that $\frac1N\sum\limits_{i=1}^N g(\hat\tee,\xii_i)=0$ (where $\hat\tee$ is the ERM in \eqref{eq:erm}), we can prove that (see more details in the proof of Theorem \ref{thm:d-fone}), 
\begin{eqnarray}\label{sdsa}
\a&=&\frac1N\sum\limits_{i=1}^N \Big(g(\hat\tee_0,\xii_i)-g(\hat\tee,\xii_i)\Big) \nonumber \\
&=&G(\hat\tee_0)-G(\hat\tee)+\frac1N\sum\limits_{i=1}^N \Big\{[g(\hat\tee_0,\xii_i)-g(\hat\tee,\xii_i)]-[G(\hat\tee_0)-G(\hat\tee)]\Big\} \cr
&=& \S(\hat\tee_0-\hat\tee)+O_{\pr}(1)\left(\|\hat\tee_0-\hat\tee\|_2\|\hat\tee_0-\tee^{*}\|_2+\|\hat\tee_0-\hat\tee\|^{2}_2\right)\cr
& &+O_{\pr}(1)\Big(\sqrt{\frac{p\log N}{N}}\|\hat\tee_0-\hat\tee\|_{2}+N^{-\gamma}\Big),
\end{eqnarray}
for  any $\gamma>0$. Note that in Algorithm \ref{algo:distributed_FONE}, the FONE procedure is executed on the first machine. For the ease of plugging the result in Proposition \ref{prop:fone} on FONE, we let $n:=n_1$ to denote the sub-sample size on the first machine. 

Assuming that the initial estimator $\hat{\tee}_{0}$  and $\hat{\tee}$ satisfy $\|\hat\tee_0-\tee^{*}\|_2+\|\hat\tee-\tee^{*}\|_2=O_{\pr}(n^{-\delta_1})$ for some $\delta_1>0$, then by \eqref{sdsa}, we have $\|\a\|_2=O_{\pr}(n^{-\delta_1})$ (i.e., the length $\tau_n=O(n^{-\delta_1})$ in Proposition \ref{prop:fone}). Moreover, we can further choose $d_n$ in Proposition \ref{prop:fone} to be $d_n=O(n^{-\delta_1})$. Let the step-size  $\eta_n=n^{-\delta_2}$ for some $\delta_2>0$. After one round of distributed FONE in Algorithm \ref{algo:distributed_FONE}, by Proposition \ref{prop:fone} and the proof of Theorem \ref{thm:d-fone}  in the supplementary material, we can obtain that $\|\hat\tee_{\mathrm{dis},K=1}-\hat\tee\|_2=O_{\pr}(n^{-\delta_1-\delta_0})$ with $\delta_0=\min(\delta_1,\delta_2/2,(1-\kappa_{1})/2)$, where $\kappa_{1}$ is the parameter in our assumption $p=O(n^{\kappa_{1}})$
(see Condition {\tt(A1$^*$)}).  Now we show the convergence rate of the $K$-th round distributed FONE by induction. In the $K$-th round, the output of the $(K-1)$-th round $\hat\tee_{\mathrm{dis},K-1}$ is used as the initial estimator $\hat\tee_0$ where $\|\hat\tee_0-\hat\tee\|_2=n^{-\delta_1-(K-1)\delta_0}$. Therefore, we can choose $d_n$ and $\tau_n$ in Proposition \ref{prop:fone} by $d_n=O(n^{-\delta_1})$ and $\tau_n=n^{-\delta_1-(K-1)\delta_0}$ from \eqref{sdsa}. As a result of Proposition \ref{prop:fone}, we obtain that $\|\hat\tee_{\mathrm{dis},K}-\hat\tee\|_2=O_{\pr}(n^{-\delta_1-K\delta_0})$. This convergence result of distributed FONE is formally stated in the next theorem. 

\begin{theorem}[distributed FONE for smooth loss function $f$]\label{thm:d-fone} 
Assume  Conditions {\tt(A1$^{*}$)} and {\tt(A2-Log)} (or  Conditions {\tt(C2)} and {\tt(C3)} in the supplement) hold, $N=O(n^{A})$ for some $A>0$. Suppose that $\|\hat{\tee}-\tee^{*}\|_{2}+\|\hat\tee_0-\tee^*\|_2=O_{\pr}(n^{-\delta_1})$ for some $\delta_1>0$. Let $\eta_n=n^{-\delta_{2}}$ for some $\delta_{2}>0$,  $\log n=o(\eta_{n}T)$, $T=O(n^{A})$ for some $A>0$, and $p\log n=o(m)$. For any $\gamma>0$, there exists $K_0>0$ such that, for any $K\geq K_0$, we have $\|\teedis-\hat{\tee}\|_2=O_{\pr}(n^{-\gamma})$. 
\end{theorem}

As we illustrate before Theorem \ref{thm:d-fone}, since $\|\hat\tee_{\mathrm{dis},K}-\hat\tee\|_2=O_{\pr}(n^{-\delta_1-K\delta_0})$ with $\delta_0=\min(\delta_1,\delta_2/2,(1-\kappa_1)/2)$. We have $K_{0}=(\gamma-\delta_1)/\delta_0$ in Theorem \ref{thm:d-fone}. We recall that $n=n_{1}$ denotes the number of samples on the first machine. Note that $\gamma$ in Theorem \ref{thm:d-fone} can be arbitrarily large. Under some regular conditions, it is typical that $\|\hat{\tee}-\tee^{*}\|_{2}=O_{\pr}(\sqrt{p/N})$. Therefore, for a smooth loss function $f$, our distributed FONE  achieves the optimal rate $O_{\pr}(\sqrt{p/N})$. Note that it does not need any condition on the number of machines $L$.
Given the step-size $\eta_n=n^{-\delta_2}$, by the condition $\log n=o(\eta_{n}T)$, we can choose the number of iterations $T=n^{\delta_2}(\log n)^2$ in the distributed FONE. Therefore, the computation complexity of distributed FONE is $O(np+n^{\delta_2}(\log n)^2mp)$ for each round, on the first machine. We also note that $n$ is the sub-sample size on the first machine, which is much smaller than the total sample size $N$. 	 In terms of the communication cost, each machine only requires to transmit an $O(p)$ vector for each round.

We note that although DC-SGD assumes the independence between the initial estimator and the sample, such a condition is no longer required in our distributed FONE for both smooth loss and non-smooth loss. Therefore, one can use the sub-sample on one local machine to construct the initial estimator. 
We also note that, in contrast to DC-SGD, it is unknown how the averaging scheme would benefit the Newton approach. Therefore, as compared to DC-ASGD in Remark \ref{rem:asgd}, it is unclear whether the use of averaging in Dis-FONE could improve the convergence rate. We leave it to future investigation.

\subsubsection{Non-smooth loss function $f$}
\label{sec:non-smooth}
For a  non-smooth loss, we provide the following convergence rate of the FONE of $\S^{-1}\a$ under Conditions {\tt(A1$^{*}$)} and {\tt(A2-QR)}. 
\begin{proposition}[On $\teeFONE$ for $\S^{-1} \a$ for non-smooth loss function $f$]\label{prop:fone_non}
	Assume  Conditions {\tt(A1$^{*}$)} and {\tt(A2-QR)} (or  Conditions {\tt(C2)} and {\tt(C3$^*$)} in the supplement)  hold. Suppose that the initial estimator satisfies $\|\hat{\tee}_{0}-\tee^{*}\|_{2}=O_{\pr}(d_n)$, and $\|\a\|_2=O(\tau_{n})$ (or $O_{\pr}(\tau_{n})$ for the random case). The iteration number $T$ and step-size $\eta_{n}$ satisfy $\log n=o(\eta_{n}T)$ and $T=O(n^{A})$ for some $A>0$. We have
	{\small\begin{align}
		\|\teeFONE&-\S^{-1}\a\|_2\nonumber \\
		& = O_{\pr}\Big{(}\tau_{n}d_n+\tau^{2}_{n}+\sqrt{\frac{p\log n}{n}}\sqrt{\tau_{n}}+\frac{p\log n}{m}\sqrt{\eta_{n}}+\sqrt{\eta_{n}}\tau_{n}+\frac{p\log n}{n}\Big{)}. \label{es2}
		\end{align}}
\end{proposition}

Compared to Proposition \ref{prop:fone}, the mini-batch size $m$ appears in the error bound of Proposition \ref{prop:fone_non} due to the discontinuity of the gradient in the nonsmooth setting. Consequently, the average gradient $\frac1m\sum_{i=1}^mg_i(\tee,\xii_i)$ has a non-negligible bias that enters into the error bound.

With Proposition \ref{prop:fone_non} in hand, we now provide the convergence rate of the distributed FONE in Algorithm \ref{algo:distributed_FONE} under Condition {\tt(A2-QR)}. It is worthwhile noting that when $f(\tee,\xii)$ is non-differentiable, then $\frac{1}{N}\sum_{i=1}^{N}g(\hat{\tee},\xii_i)$ can be nonzero  due to the discontinuity in the function $g(\tee,\xii)$, where $\widehat{\tee}$ is the ERM in \eqref{eq:erm}. Therefore we need to assume that
\begin{eqnarray}\label{c4}
\sum_{i=1}^{N}g(\hat{\tee},\xii_i)=O_{\pr}(q_{N})
\end{eqnarray}
with $q_{N}=O(N^{\kappa_{3}})$ for some $0<\kappa_{3}<1$. For example, for a quantile regression, $q_{N}=O(p^{3/2}\log N)$ \citep{heshao2000}, which satisfies this condition when $p=o(N^{\kappa_4})$ with $0<\kappa_4 < 2/3$.

Given {\tt(A1$^{*}$)}, {\tt(A2-QR)} and \eqref{c4}, we have the following convergence rate of $\teedis$:

\begin{theorem}[distributed FONE for non-smooth loss function $f$]\label{thm:d-fone-non} 
Suppose  Conditions  {\tt(A1$^{*}$)} and {\tt(A2-QR)} (or  Conditions {\tt(C2)} and {\tt(C3$^*$)} in the supplement) and \eqref{c4} hold, $N=O(n^{A})$ and $T=O(n^{A})$ for some $A>0$.  Suppose that $\|\hat{\tee}-\tee^{*}\|_{2}+\|\hat\tee_0-\tee^*\|_2=O_{\pr}(n^{-\delta_1})$ for some $\delta_1>0$. Let $\eta_n=n^{-\delta_{2}}$ for some $\delta_{2}>0$,  $\log n=o(\eta_{n}T)$, and $p\log n=o(m)$. For any $\frac12<\gamma<1$, there exists $K_0>0$ such that, for any $K\geq K_0$, we have
	\begin{eqnarray}\label{eq:conv_non-smooth}
	\|\teedis-\hat{\tee}\|_2=O_{\pr}\Big{(}\frac{q_{N}}{N}+\sqrt{\eta_{n}}\frac{p\log n}{m}+\big{(}\frac{p\log n}{n}\big{)}^{\gamma}\Big{)}.
	\end{eqnarray}
\end{theorem}

As one can see from \eqref{eq:conv_non-smooth},  the distributed FONE has a faster convergence rate when the sub-sample size on the first machine $n_1$ is large (recall that $n:=n_1$). In practice, it is usually affordable to increase the memory and computational resources for only one local machine. This is different from the case of DC-SGD, where the convergence rate actually depends on the smallest sub-sample size among local machines.\footnote{Noting that although we present the evenly distributed setting for DC-SGD for the ease of illustration, one can easily see the convergence rate is actually determined by the smallest sub-sample size from the proof.}

The parameter $\gamma$ in the exponent of the last term serves as a target rate of convergence. 
More specifically, the convergence rate after $K$ rounds is (See Section D.4 for more details),
\[\small
\|\teedis-\hat{\tee}\|_2=O_{\pr}\Big{(}n^{-\delta_{1}-K\delta_0}+\frac{q_{N}}{N}+\sqrt{\eta_{n}}\frac{p\log n}{m}+\big{(}\frac{p\log n}{n}\big{)}^{\gamma}\Big{)}\text{, where }\delta_0=\min\left\{\frac{\delta_{1}(1-\gamma)}{2\gamma-1}, \frac{\delta_1}{2},\frac{\delta_2}{4}\right\}.
\]
Therefore, when $K>K_{0}:=(\kappa_1\gamma-\delta_1)/\delta_0$,  the term $n^{-\delta_{1}-K\delta_0}$ is bounded by the term of in the convergence rate $\big{(}\frac{p\log n}{n}\big{)}^{\gamma}$, where $\kappa_1$ is the parameter in the assumption of $p=O(n^{\kappa_1})$ in Condition {\tt(A$1^*$)}. 
For the second last term  in the right-hand side of \eqref{eq:conv_non-smooth}, we can choose the step-size $\eta_n$ and the batch size $m$ such that $\sqrt{\eta_n}/m\leq (p\log n)^{\gamma-1}/n^{\gamma}$, and the convergence rate of $\|\teedis-\hat{\tee}\|_2$ is thus dominated by $q_{N}/N+\big((p\log n)/n\big)^\gamma$. Due to $q_{N}$ in \eqref{c4}, the relationship between these two terms depends on the specific model. Usually, under some conditions on the dimension $p$, $\|\teedis-\hat{\tee}\|_2$ achieves a faster rate than  $\|\hat{\tee}-\tee^*\|_2$, which makes $\teedis$ attain the optimal rate for estimating $\tee^*$.  Let us take the quantile regression as an example, where the ERM $\hat\tee$ has an error rate of $\|\hat\tee-\tee^*\|_2=O_{\pr}(\sqrt{p/N})$ and  $q_{N}=O(p^{3/2}\log N)$ \citep{heshao2000}. Assuming that $p=O(\sqrt{N}/\log N)$  and $n\geq cN^{\frac{1}{2\gamma}}p^{1-\frac{1}{2\gamma}}\log n$, we have $\|\teedis-\tee^*\|_2=O_{\pr}(\sqrt{p/N})$. 


Similar to the smooth case, the computation complexity is $O(np+n^{\delta_2}(\log n)^2mp)$ for each round, on the first machine. Assuming the second term of \eqref{eq:conv_non-smooth} is dominated by the third term,  we may specify $m=\sqrt{\eta_n}n\log n$ and the corresponding computation complexity becomes $O(n^{1+\delta_2/2}(\log n)^3p)$. Again, each machine only requires to transmit an $O(p)$ vector for each round.

\section{Inference: application of FONE to the estimation of $\S^{-1}\w$ with $\|\w\|_2=1$}\label{sec:unitw}

An important application of the proposed FONE is to conduct the inference of $\tee^{*}$ in the diverging $p$ case. To provide asymptotic valid inference, we only need a consistent estimator of the limiting variance in  \eqref{eq:asy}.

To estimate the limiting variance, we note that $\A$ can be easily estimated by  $\widehat{\A}=\frac1n\sum\limits_{i=1}^n g(\widehat{\tee}_{0},\xii_{i})g(\widehat{\tee}_{0},\xii_{i})'$. Therefore, we only need to estimate $\S^{-1}\w$. The challenge here is the theory of our Propositions \ref{prop:fone} and \ref{prop:fone_non} only applies to the case $\S^{-1}\a$ with $\|\a\|_2=o(1)$ or $o_{\pr}(1)$. However, in the inference application, we have $\|\w\|_2=1$. To address this challenge, given the unit length vector $\w$, we define $\a=\tau_n \w$, where $\|\a\|_2=\tau_n=o(1)$ and its rate will be specified later in our theoretical results in Theorems \ref{thm:sinvw} and \ref{thm:sinvw-non}. We run Algorithm \ref{algo:fone} and its output $\hat\tee_0-\z_T$ is an estimator of $\tau_n \S^{-1} \w$. Then the estimator of $\S^{-1}\w$ can be naturally constructed as,
\begin{eqnarray}\label{eq:Sw_infer}
\widehat{\S^{-1}\w}=\frac{\hat{\tee}_{0}-\z_{T}}{\tau_{n}},\quad\mbox{where  in  Algorithm \ref{algo:fone} \; $\a=\tau_{n}\w$. }
\end{eqnarray}
We note that the initial estimator $\hat{\tee}_{0}$ for estimating $\S^{-1}\w$ needs to be close to the target parameter  $\tee^*$. In a non-distributed setting, we could choose the ERM $\hat\tee$ as $\hat{\tee}_{0}$ for inference, while in the distributed setting, we use $\teedis$ from distributed FONE in Algorithm \ref{algo:distributed_FONE} with a sufficiently large $K$. 

Furthermore, we briefly comment on an efficient implementation for computing the limiting variance $\w'\S^{-1}\A\S^{-1}\w$. Instead of explicitly constructing the estimator of $\A$ by a $p \times p$ matrix $\widehat{\A}=\frac1n\sum\limits_{i=1}^n g(\widehat{\tee}_{0},\xii_{i})g(\widehat{\tee}_{0},\xii_{i})'$, we can directly compute the estimator by 
\begin{equation}\label{eq:imp}
(\widehat{\S^{-1} \w})'\widehat{\A}(\widehat{\S^{-1} \w}) = \frac{1}{n} \sum_{i=1}^n \left(g(\hat{\tee}_0, \xii_{i})' \widehat{\S^{-1} \w}\right)^2,
\end{equation}
where $\widehat{\S^{-1} \w}$ is pre-computed by FONE. The implementation in \eqref{eq:imp} only incurs a computation cost of $O(np)$. 

We next provide the theoretical results of the estimator in \eqref{eq:Sw_infer} for two cases: $f$ is smooth and $f$ is non-smooth. We note that for the purpose of asymptotic valid inference, we only need $\widehat{\S^{-1}\w}$ in \eqref{eq:Sw_infer} to be a consistent estimator of $\S^{-1}\w$. To show the consistency of our estimator, we provide the convergence rates in the following Theorems \ref{thm:sinvw} and \ref{thm:sinvw-non} for smooth and non-smooth loss functions, respectively:

\begin{theorem}[Estimating $\S^{-1}\w$ for a smooth loss function $f$]\label{thm:sinvw}  Under the conditions of Proposition \ref{prop:fone}, let $\tau_{n}=\sqrt{(p\log n)/n}$. Assuming that $\|\hat\tee_0-\tee^*\|_2=O_{\pr}(d_n)$ and $\log n=o(\eta_{n}T)$, we have
	\begin{eqnarray}\label{eq:Sw_smooth}
	\|\widehat{\S^{-1}\w}-\S^{-1}\w\|_2=O_{\pr}\Big{(}\sqrt{\frac{p\log n}{n}}+\sqrt{\eta_n}+d_n\Big{)}.
	\end{eqnarray}
\end{theorem}

From Theorem \ref{thm:sinvw}, the estimator $\widehat{\S^{-1}\w}$ is consistent as long as $d_n=o(1)$ and the step-size $\eta_n=o(1)$. Let us further provide some discussion on the convergence rate in \eqref{eq:Sw_smooth}.   If we choose a good initiation such that $d_n=O(\sqrt{(p\log n)/n})$, the term $d_n$ in \eqref{eq:Sw_smooth} will be a smaller order term. For example, the  initialization rate $d_n=O(\sqrt{(p\log n)/n})$ can be easily satisfied by using either the ERM $\hat\tee$ or $\teedis$ from distributed FONE with a sufficiently large $K$.  Moreover, we can specify $\eta_n$ to be small (e.g., $\eta_n=O((p\log n)/n)$). Then the rate in  \eqref{eq:Sw_smooth} is $\sqrt{(p\log n)/n}$, which almost matches the parametric rate for estimating a $p$ dimensional vector.

For non-smooth loss function $f$, we have the following convergence rate of $\hat{\S^{-1}\w}$:

\begin{theorem}[Estimating $\S^{-1}\w$ for non-smooth loss function $f$]\label{thm:sinvw-non} Under the conditions of Proposition \ref{prop:fone_non}, let $\tau_n=\big((p\log n)/n\big)^{1/3}$. Assuming that $\|\hat\tee_0-\tee^*\|_2=O_{\pr}(d_n)$ and $\log n=o(\eta_{n}T)$, we have 
	\begin{equation}\label{eq:Sw_nonsmooth}
	\|\widehat{\S^{-1}\w}-\S^{-1}\w\|_2=O_{\pr}\Big(\big(\frac{p\log n}{n}\big)^{1/3}+\sqrt{\eta_n}\big(\frac{n^{1/3}(p\log n)^{2/3}}{m}+1\big)+d_n\Big{)}.
	\end{equation}
\end{theorem}

To make $d_n$ a smaller order term in the rate in \eqref{eq:Sw_nonsmooth}, we choose a good initiation such that $d_n=O((p\log n)/n\big)^{1/3})$. As long as the step-size $\eta_n$ is small such that $\eta_{n}=\min\big(\frac{(p\log n)^{2/3}}{n^{2/3}},\frac{m^2}{(p\log n)^{2/3}n^{4/3}}\big)$, the convergence rate in \eqref{eq:Sw_nonsmooth} is 
$O_{\pr}\big(((p\log n)/n)^{1/3}\big)$, which implies that $\widehat{\S^{-1}\w}$ is a consistent estimator of $\S^{-1}\w$.

\section{Experimental Results}
\label{sec:exp}
In this section, we provide simulation studies and real data analysis to illustrate the performance of our methods on two statistical estimation problems in Examples \ref{ex:log}--\ref{ex:qr}, i.e., logistic regression and quantile regression (QR). 

\subsection{Simulation Studies}
For regression problems in the two motivating examples, let $\xii_i=(Y_i,\X_i)$ for $i=1,2,\dots, N$, where $\X_i=(1,X_{i,1},X_{i,2},\dots,X_{i,p-1})'\in \mathbb{R}^p$ is a random covariate vector and $N$ is the total sample size. Here $(X_{i,1},X_{i,2},\dots,X_{i,p-1})$ follows a multivariate normal distribution $\mathcal{N}(\mathbf{0},\mathbf{I}_{p-1})$, where $\mathbf{I}_{p-1}$ is a $p-1$ dimensional identity matrix. We also provide the simulation studies with correlated design $\X$ as well as the cases that the samples on local machines are non-identically distributed, which are relegated to the supplementary material due to space limitations (see Section G.1). The true coefficient $\tee^*$ follows a uniform distribution $\mathrm{Unif}([-0.5,0.5]^p)$. For QR in Example \ref{ex:qr}, we follow the standard approach  (see, e.g., \cite{pang2012variance}) that first generates the data from a linear regression model $Y_i = \X_i' \tee+ \epsilon_i$, where $\tee$ follows a uniform distribution $\mathrm{Unif}([-0.5,0.5]^p)$ and $\epsilon_i \sim N(0,1)$. For each quantile level $\tau$, we need to compute the true QR coefficient $\tee^*$ by shifting $\epsilon_i$ such that $\Pr(\epsilon_i \leq 0)=\tau$. Thus, the true QR coefficient $\tee^*=\tee+(\Phi^{-1}(\tau), 0,0,\ldots,0)'$, where $\Phi$ is the CDF of the standard normal distribution. In our experiment, we set the quantile level $\tau=0.25$.  All of the data points are evenly distributed on $L$ machines with sub-sample size $n=n_i=N/L$ for $i=1,2,\dots, L$. We further discuss the imbalanced situation in Section \ref{sec:n1}.

In the following experiments, we evaluate the proposed distributed FONE (Dis-FONE, see Algorithm \ref{algo:distributed_FONE}) in terms of the $L_2$-estimation errors, and compare its performance with that of the DC-SGD estimator (see Algorithm \ref{algo:dc-sgd}). In particular, we report the $L_2$-distance to the true coefficient $\tee^*$ as well as the $L_2$-distance to the ERM $\hat\tee$ in \eqref{eq:erm}, which is considered as the non-distributed ``oracle'' estimator.  
We also compare the methods with mini-batch SGD in \eqref{eq:SGD_mini} on the entire dataset in a non-distributed setting, which can be considered as a special case of DC-SGD when the number of machines $L=1$. For all these methods, it is required to provide a consistent initial estimator $\hat\tee_0$. In our experiments below, we compute the initial estimator by minimizing the empirical risk function in \eqref{eq:erm} with a small batch of fresh samples (which is also used by ERM for the fair comparison). It is clear that as dimension $p$ grows, it requires more samples to achieve the desired accuracy of the initial estimator. Therefore, we specify the size of the fresh samples as $n_0=10p$. We note that the fresh samples are used only because DC-SGD requires the independence between the initial estimator and the samples. This is not a requirement for our distributed FONE method. Also, although we allow $p$ to diverge, the sample size $10p$ is still considered as a small batch of samples since the condition {\tt(A1)} requires $p=o(n)$, i.e., $p$ grows much slower than $n$.
We also discuss the effect of the accuracy of the initial estimator $\hat\tee_0$ by varying $n_0$ (see Section G.2 in the supplementary material).

For DC-SGD, the step-size is set to $r_i=c_0/\max(i^\alpha,p)$ with $\alpha=1$, and $c_0$ is a positive scaling constant. We use an intuitive data-driven approach to choose $c_0$. We first specify a set $\mathcal{C}$ of candidate choices for $c_0$ ranging from $0.001$ to $1000$. We choose the best $c_0$ that achieves the smallest objective function in \eqref{eq:erm}  with $\tee=\teeSGD^{(1)}$ using data points from the first machine  (see Algorithm \ref{algo:dc-sgd}), i.e.,
$
c_0=\argmin_{c \in \mathcal{C}} \frac{1}{n} \sum_{i=1}^n f(\teeSGD^{(1)}, \xii_i^{(1)}),
$
where $\{\xii_i^{(1)},i=1,2,\dots,n\}$ denotes the samples on the first machine.
For Dis-FONE, the step-size is set to $\eta=c_0'm/n$, where $c_0'$ is also selected from a set $\mathcal{C}$ of candidate constants. Similarly, we choose the best tuning constant that achieves the smallest objective in  \eqref{eq:erm} with $\tee=\hat\tee_{\mathrm{dis},1}$ and samples from the first machine. Here,  $\hat\tee_{\mathrm{dis},1}$ is the output of Dis-FONE after the first round of the algorithm. That is, 
$
c_0'=\argmin_{c \in \mathcal{C}} \frac{1}{n} \sum_{i=1}^n f(\hat\tee_1, \xii_i^{(1)}).
$
More simulation results with different choices of the stepsizes are provided in Section G.3 of the supplementary material. 

Moreover, by Condition {\tt(A1)}, we set the mini-batch size in DC-SGD (or the size of $B_t$ in Dis-FONE, see Algorithm \ref{algo:distributed_FONE}) as $m= 
\lfloor p\log n \rfloor $, where $\lfloor x\rfloor$ denotes the largest integer less than or equal to $x$. For Dis-FONE, we first set the number of the iterations $T$ in each round as $T=20$ and the number of rounds $K=20$ for logistic regression and $K=80$ for quantile regression. Note that due to the non-smoothness in the loss function of quantile regression, it requires more rounds of iterations $K$ to ensure the convergence. In practice, we could also adopt a simple data-driven approach to determine the number of rounds $K$. In particular, we could stop the algorithm when the change (norm of the difference) between the estimators for two consecutive rounds is negligible. We carefully evaluate the effect of $T$ and $K$ (by considering different values of $T$ and $K$) in Section \ref{sec:T_K}. We also compare the methods with $\hat\tee_{\mathrm{FONE}}$ (Algorithm \ref{algo:fone}), which  corresponds to  the Dis-FONE $\teedis$ with only $K=1$ round. All results reported below are based on the average of $100$ independent runs of simulations.

\begin{table}[!t]
	\centering
	\caption{Logistic regression: comparisons of $L_2$-errors when varying the total sample size $N$ and dimension $p$ changes. Here the number of machines $L=20$. Denote by $\hat\tee_{\mathrm{DC}}$ the DC-SGD estimator, $\hat\tee_{\mathrm{SGD}}$ the SGD estimator on the entire dataset in a non-distributed setting,  $\hat\tee_{\mathrm{FONE}}$ the single-round FONE of Algorithm \ref{algo:fone}, and $\teedis$ the Dis-FONE with $K=20$.}
	\vspace{.25cm}
	\begin{tabular}{cc|cccccc|ccc}
		\hline
		& $p$ &	\multicolumn{6}{c|}{$L_2$-distance to the truth $\tee^*$} & 	\multicolumn{3}{c}{$L_2$-distance to ERM $\hat\tee$}\\
		&  & $\hat\tee_0$ & $\hat\tee_{\mathrm{DC}}$ & $\hat\tee_{\mathrm{SGD}}$ & $\hat\tee_{\mathrm{FONE}}$ & $\teedis$  & $\hat\tee$ & $\hat\tee_{\mathrm{DC}}$ & $\hat\tee_{\mathrm{SGD}}$ & $\teedis$ \\
		\hline
		\multicolumn{2}{l|}{$N=10^5$}	&   &      &   &     &      &         &       &           &      \\
		& 100 & 1.251 & 0.447 & 0.148 &0.724& 0.103  & 0.093 & 0.445 & 0.116 & 0.038 \\
		& 200 & 1.899 & 1.096 & 0.523 & 1.046& 0.168  & 0.153 & 1.091 & 0.494 & 0.049 \\
		& 500 & 4.509 & 3.853 & 3.111 &2.154& 0.338  & 0.301 & 3.748 & 3.021 & 0.085 \\
		\hline
		\multicolumn{2}{l|}{$N=2\times10^5$} &     &&      &      &      &         &       &           &      \\
		& 100 & 1.303 & 0.390 & 0.100 &0.594& 0.072  & 0.067 & 0.386 & 0.074 & 0.025 \\
		& 200 & 2.094 & 1.248 & 0.315 &0.821& 0.115  & 0.109 & 1.235 & 0.286 & 0.034 \\
		& 500 & 4.717 & 3.920 & 2.189 &1.725& 0.222  & 0.211 & 3.891 & 2.133 & 0.045 \\
		\hline
		\multicolumn{2}{l|}{$N=5\times10^5$}        &      &      &     &      &         &       &           &      \\
		& 100 & 1.342 & 0.313 & 0.081&0.347 & 0.046  & 0.042 & 0.304 & 0.069 & 0.018 \\
		& 200 & 1.833 & 0.874 & 0.169 &0.749& 0.073  & 0.068 & 0.868 & 0.152 & 0.023 \\
		& 500 & 4.835 & 3.885 & 1.006  &1.413& 0.141 & 0.130 & 3.859 & 0.989 & 0.036 	\\
		\hline
	\end{tabular}
	\label{table:np}
\end{table}

\begin{table}[!t]
	\centering
	\caption{Quantile regression: comparisons of $L_2$-errors when varying the total sample size $N$ and dimension $p$. Here the number of machines $L=20$.  Denote by  $\hat\tee_{\mathrm{DC}}$ the DC-SGD estimator, $\hat\tee_{\mathrm{SGD}}$ the SGD estimator on the entire dataset in a non-distributed setting, $\hat\tee_{\mathrm{FONE}}$ the single-round FONE of Algorithm \ref{algo:fone}}, and $\teedis$ the Dis-FONE with  $K=80$.
	\vspace{.25cm}
	
  \begin{tabular}{cc|cccccc|ccc}
    \hline
    & $p$ & \multicolumn{6}{c|}{$L_2$-distance to the truth $\tee^*$} &   \multicolumn{3}{c}{$L_2$-distance to ERM $\hat\tee$}\\
    &  & $\hat\tee_0$ & $\hat\tee_{\mathrm{DC}}$ & $\hat\tee_{\mathrm{SGD}}$ & $\hat\tee_{\mathrm{FONE}}$ & $\teedis$  & $\hat\tee$ & $\hat\tee_{\mathrm{DC}}$ & $\hat\tee_{\mathrm{SGD}}$ & $\teedis$ \\
    \hline
    \multicolumn{2}{l|}{$N=10^5$} &   &      &   &     &      &         &       &           &      \\
		& 100 & 0.450     & 0.079  & 0.063    &0.191  & 0.047 & 0.043    & 0.073  & 0.050      & 0.020 \\
		& 200 & 0.715     & 0.114  & 0.109    &0.342 & 0.082 & 0.071    & 0.106  & 0.097      & 0.035 \\
		& 500 & 1.278     & 0.198  & 0.176    &0.554  & 0.144 & 0.126    & 0.176  & 0.142      & 0.062 \\
		\hline
		\multicolumn{2}{l|}{$N=2\times10^5$} &     &      &      &      &         &       &           &      \\
		& 100 & 0.450     & 0.070  & 0.037     &0.130 & 0.035 & 0.030    & 0.067  & 0.021      & 0.015 \\
		& 200 & 0.726     & 0.101  & 0.067    &0.246  & 0.059 & 0.054    & 0.098  & 0.037      & 0.027 \\
		& 500 & 1.287    & 0.176  & 0.118    &0.379  & 0.098 & 0.076    & 0.157  & 0.065      & 0.046 \\
		\hline
		\multicolumn{2}{l|}{$N=5\times10^5$}        &      &      &     &      &         &       &           &      \\
		& 100 & 0.451     & 0.043  & 0.030    &0.114  & 0.029 & 0.025    & 0.037  & 0.017      & 0.014 \\
		& 200 & 0.719     & 0.067  & 0.047     &0.157 & 0.041 & 0.037    & 0.064  & 0.026      & 0.020 \\
		& 500 & 1.294     & 0.105  & 0.076    &0.264  & 0.074 & 0.057   & 0.091  & 0.041      & 0.035 \\
		\hline
	\end{tabular}
	\label{table:np_qr}
\end{table}

\begin{figure}[!t]
	\centering
	\subfigure[Logistic regression: $L_2$-distance to $\tee^*$]{\includegraphics[width = 0.47\textwidth]{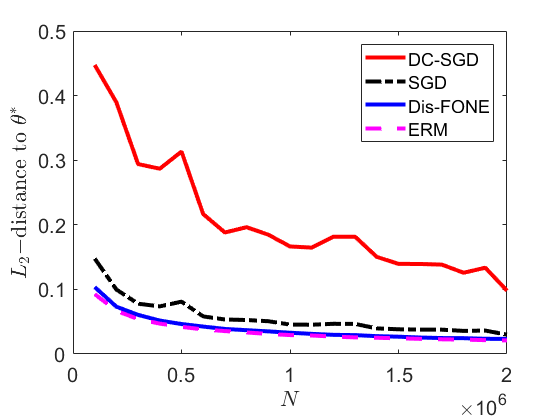}\label{fig:N1}}
	\subfigure[Logistic regression: $L_2$-distance to $\hat\tee$]{\includegraphics[width = 0.47\textwidth]{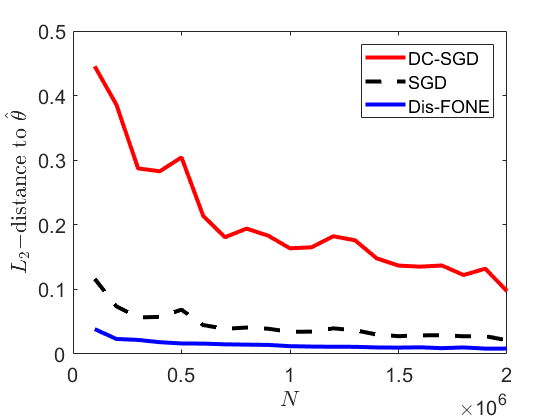}\label{fig:N2}}
	\subfigure[Quantile regression: $L_2$-distance to $\tee^*$]{\includegraphics[width = 0.47\textwidth]{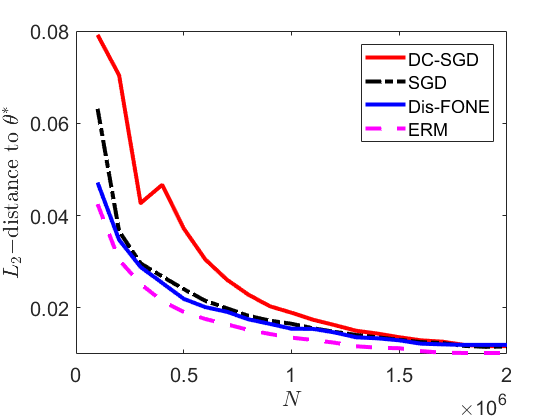}\label{fig:N3}}
	\subfigure[Quantile regression: $L_2$-distance to $\hat\tee$]{\includegraphics[width = 0.47\textwidth]{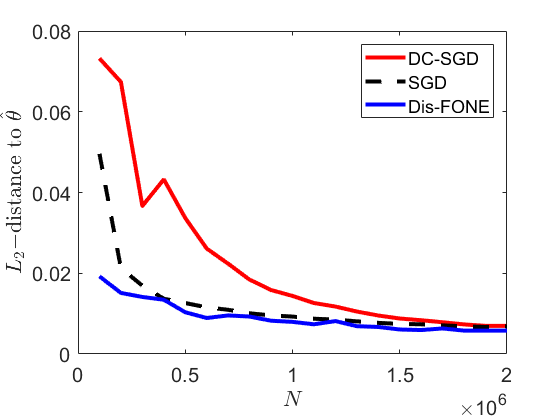}\label{fig:N4}}
	\caption{Comparison of $L_2$-errors when $N$ increases. The left column reports the $L_2$-errors with respect to the truth $\tee^*$ and the right column reports the $L_2$-errors with respect to the ERM $\hat\tee$. Here the dimension $p=100$ and the number of machines $L=20$. In Dis-FONE, we set   $K=20$ in the logistic regression case and  $K=80$ in the quantile regression case.}
	\label{fig:N}
\end{figure}
\subsubsection{Effect of $N$ and $p$}

\label{sec:exp_np}
In Tables \ref{table:np}--\ref{table:np_qr}, we fix the number of machines $L=20$ and vary the total sample size $N$ from $\{10^5,2\times 10^5, 5\times 10^5\}$ and dimension $p\in \{100,200,500\}$. Results for logistic regression are reported in Table \ref{table:np} and results for quantile regression are in Table \ref{table:np_qr}. In both tables, the left columns provide the $L_2$ estimation errors (with respect to the truth $\tee^*$) of the   DC-SGD estimator $\hat\tee_{\mathrm{DC}}$, the SGD estimator $\hat\tee_{\mathrm{SGD}}$, Dis-FONE $\teedis$, and the ERM $\hat \tee$. We note that both SGD and ERM are non-distributed algorithms for pooled data. We will show that in many cases our distributed Dis-FONE estimator even outperforms the non-distributed SGD.  For reference, we also report  $L_2$-errors of the initial estimator  $\hat\tee_0$. The right columns report the $L_2$-distances to the  ERM $\hat\tee$.  

From Tables \ref{table:np}--\ref{table:np_qr}, we can see that the proposed Dis-FONE $\teedis$ achieves similar errors as the ERM $\hat\tee$ in all cases, and outperforms DC-SGD  and SGD especially when $p$ is large.   We also provide Figure \ref{fig:N} that captures the performance of the estimators in terms of their $L_2$-errors when the total sample size $N$ increases. From Figure \ref{fig:N}, we can see that the estimation error for each method decreases as $N$ increases. Moreover, the $L_2$-error of Dis-FONE is very close to the ERM as $N$ increases, while there is a significant gap between DC-SGD and the ERM.

\begin{figure}[!t]
	\centering
	\subfigure[Logistic regression: $L_2$-distance to $\tee^*$]{\includegraphics[width = 0.47\textwidth]{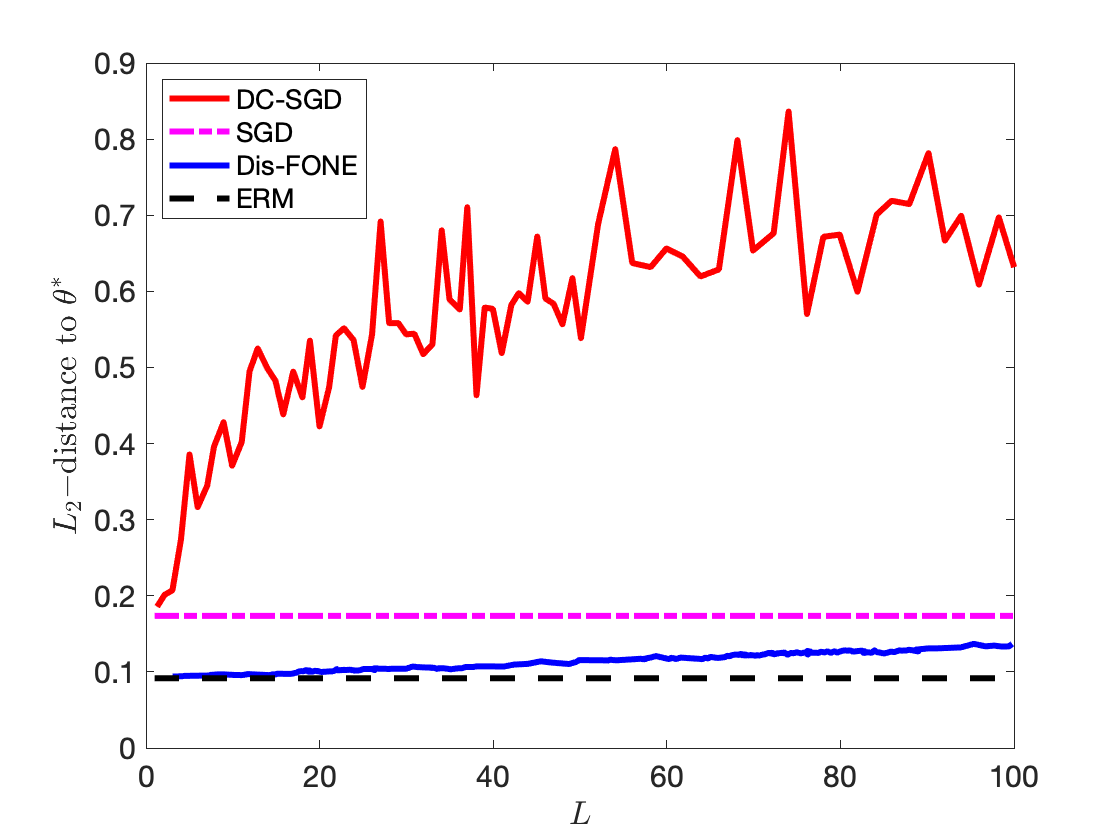}\label{fig:L1}}
	\subfigure[Logistic regression: $L_2$-distance to $\hat\tee$]{\includegraphics[width = 0.47\textwidth]{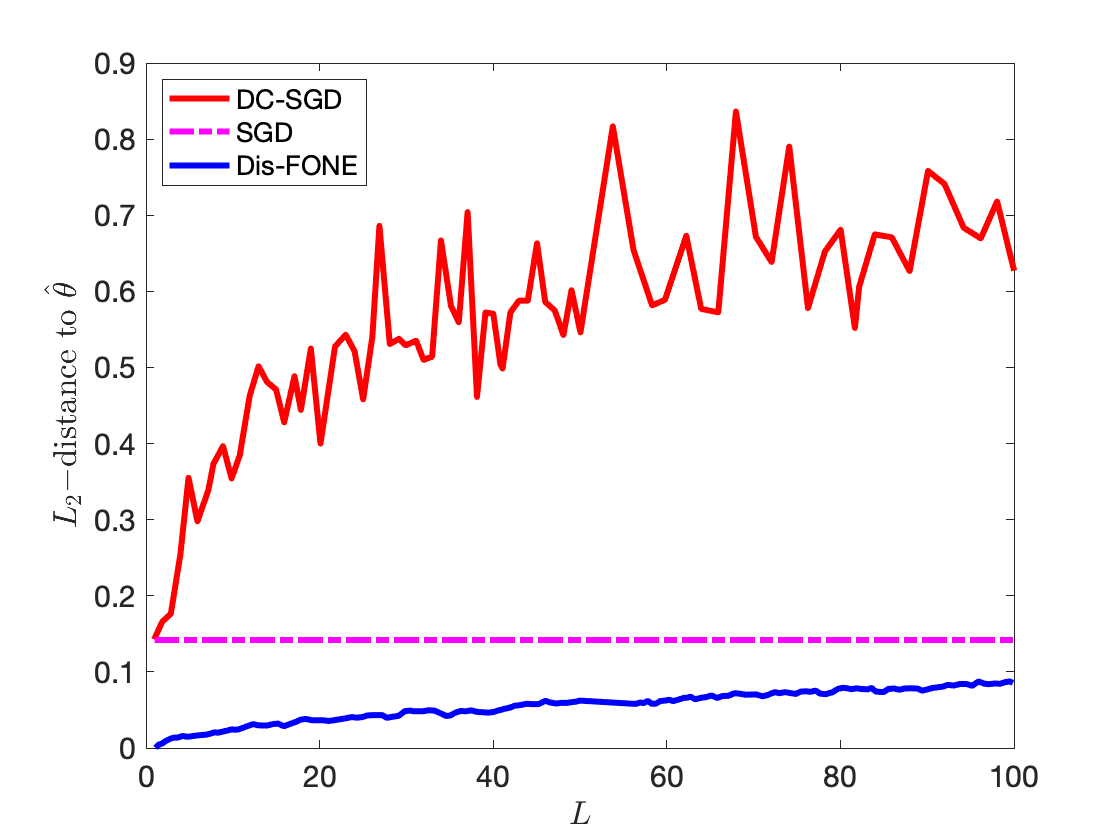}\label{fig:L2}}
	\subfigure[Quantile regression: $L_2$-distance to $\tee^*$]{\includegraphics[width = 0.47\textwidth]{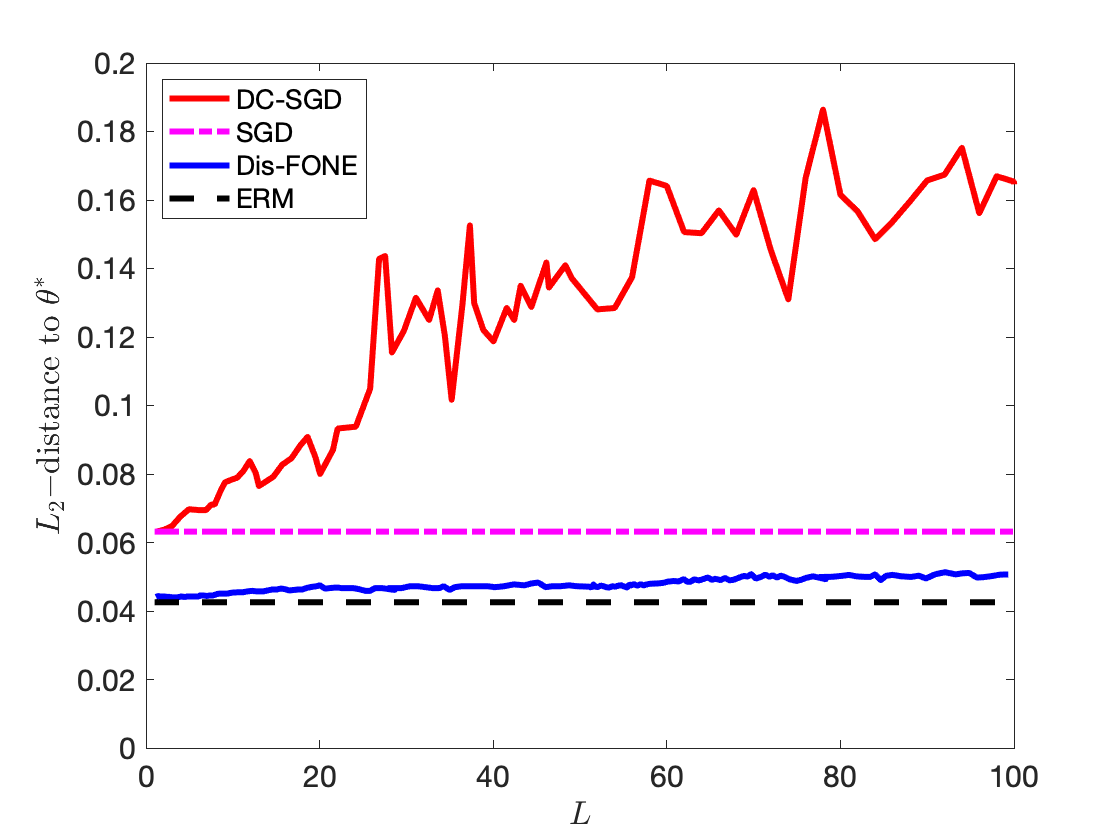}\label{fig:L3}}
	\subfigure[Quantile regression: $L_2$-distance to $\hat\tee$]{\includegraphics[width = 0.47\textwidth]{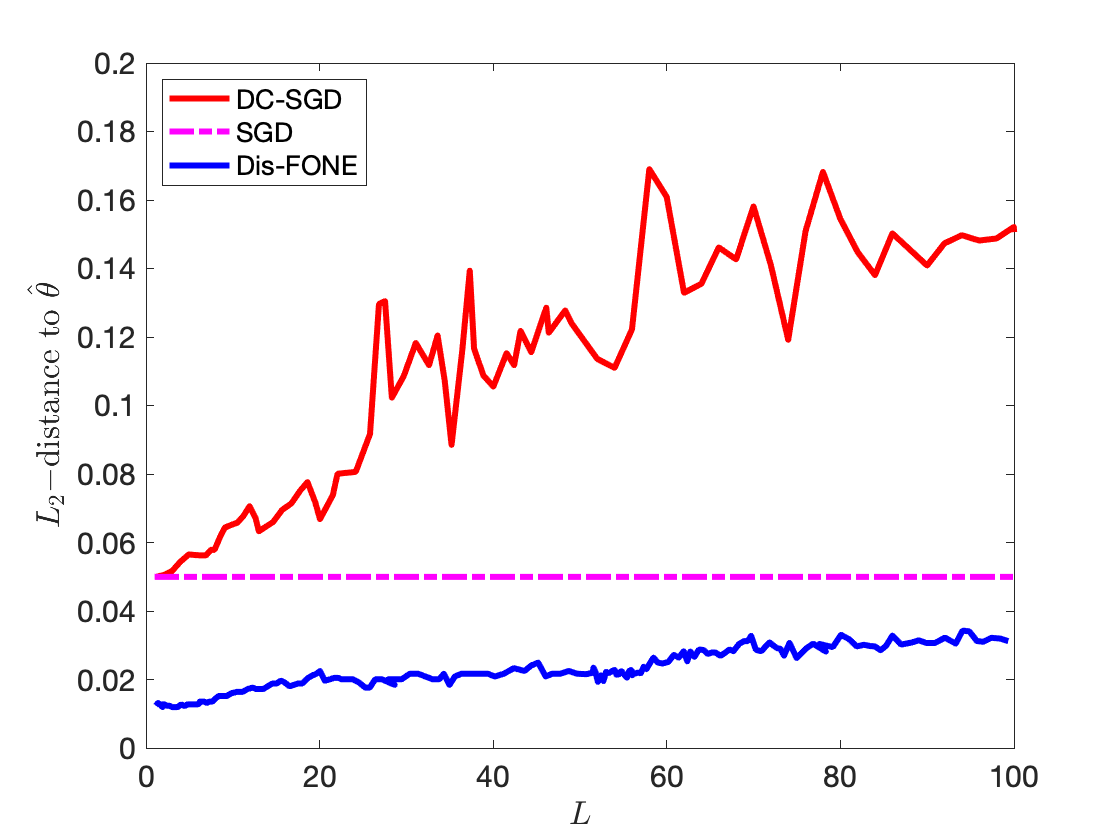}\label{fig:L4}}
	\caption{Comparison of $L_2$-errors when the number of machines $L$ increases. Here the total sample size $N=10^5$ and the dimension $p=100$. Denote by $\teedis$ the Dis-FONE with   $K=20$ in the logistic regression case and  $K=80$ in the quantile regression case.}
	\label{fig:L}
\end{figure}

\subsubsection{Effect on the number of machines $L$}

\label{sec:exp_l}
 For the effect on the number of machines $L$, we fix the total sample size $N=10^5$ and the dimension $p=100$ and vary the number of machines $L$ from $1$ to $200$, and plot the $L_2$-errors in Figure \ref{fig:L}. From  Figure \ref{fig:L}, the $L_2$-error of DC-SGD increases as $L$ increases (i.e., each machine has fewer samples). In contrast, the $L_2$-error of Dis-FONE versus $L$ is almost flat, and is quite close to ERM even when $L$ is large. This is consistent with our theoretical result that DC-SGD will fail when $L$ is large. The SGD estimator, which is the $L=1$ case of DC-SGD (and thus its error is irrelevant of $L$ and is presented by a horizontal line), provides moderate accuracy. Further increasing $L$ would lead to an excessively small local sample size (e.g., when $N=10^5$ and $L>200$, we will have the local sample size $n<500$, and thus is  unrealistic in practice, considering that the dimensionality $p=100$.

\begin{figure}[!t]
	\centering
	\subfigure[Logistic regression: $L_2$-distance to $\tee^*$]{\includegraphics[width = 0.47\textwidth]{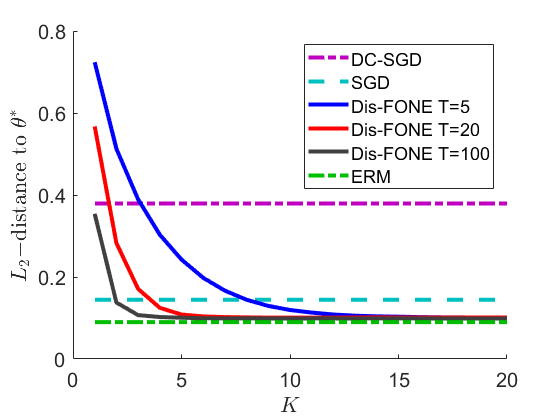}\label{fig:K1}}
	\subfigure[Logistic regression: $L_2$-distance to $\hat\tee$]{\includegraphics[width = 0.47\textwidth]{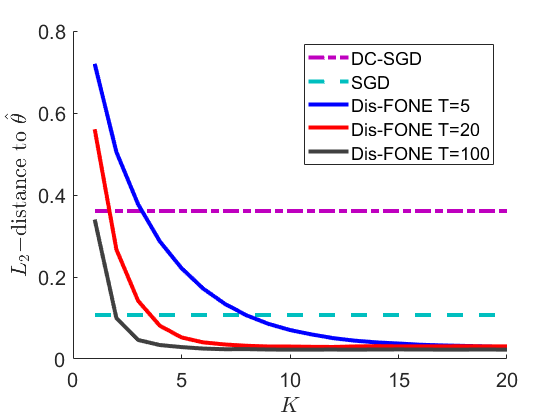}\label{fig:K2}}
	\subfigure[Quantile regression: $L_2$-distance to $\tee^*$]{\includegraphics[width = 0.47\textwidth]{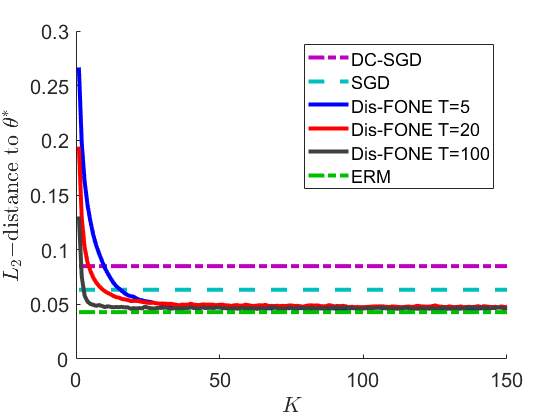}\label{fig:K3}}
	\subfigure[Quantile regression: $L_2$-distance to $\hat\tee$]{\includegraphics[width = 0.47\textwidth]{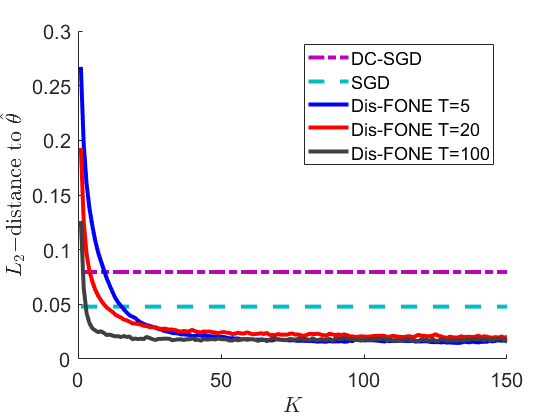}\label{fig:K4}}
	\caption{Comparison of $L_2$-errors when the number of rounds $K$ in Dis-FONE increases. The $x$-axis is the number of rounds $K$ in Dis-FONE. Here the total sample size $N=10^5$, the dimension $p=100$,  and the number of machines $L=20$. The errors of DC-SGD, SGD, and ERM are presented by the horizontal lines since their performance is irrelevant of $K$.}
	\label{fig:K}
	\vspace{-2mm}
\end{figure}

\subsubsection{Effect of $K$ and $T$ in Dis-FONE}
\label{sec:T_K}
For Dis-FONE, we provide the comparison of the estimator errors with different numbers of rounds $K$ and numbers of inner iterations $T$. In Figure \ref{fig:K}, we fix the total sample size $N=10^5$, the dimension $p=100$, the number of machines $L=20$  and vary $T$ from $\{5,20,100\}$. The $x$-axis in Figure \ref{fig:K} is the number of rounds $K$. For all three cases of $T$, the performance of Dis-FONE is quite desirable and reaches the accuracy of the ERM when $K$ becomes larger. When $T$ is smaller, it requires a larger $K$ for Dis-FONE to converge. In other words, we need to perform more rounds of Dis-FONE to achieve the same accuracy. 

\begin{figure}[!t]
	\centering
	\subfigure[Logistic regression: $L_2$-distance to $\tee^*$]{\includegraphics[width = 0.47\textwidth]{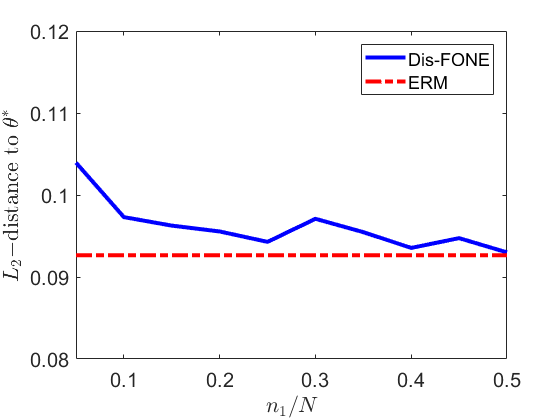}\label{fig:n11}}
	\subfigure[Logistic regression: $L_2$-distance to $\hat\tee$]{\includegraphics[width = 0.47\textwidth]{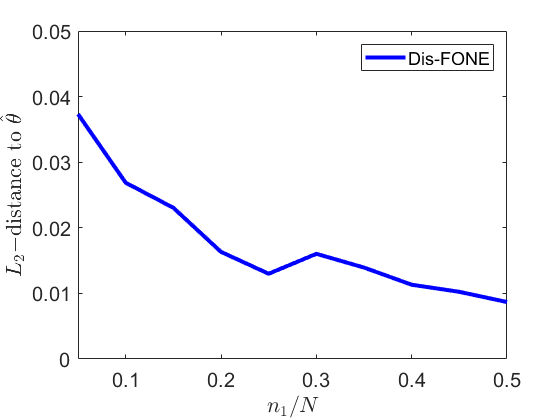}\label{fig:n12}}
	\subfigure[Quantile regression: $L_2$-distance to $\tee^*$]{\includegraphics[width = 0.47\textwidth]{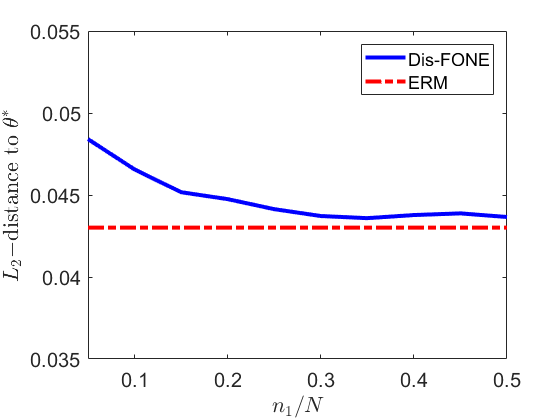}\label{fig:n13}}
	\subfigure[Quantile regression: $L_2$-distance to $\hat\tee$]{\includegraphics[width = 0.47\textwidth]{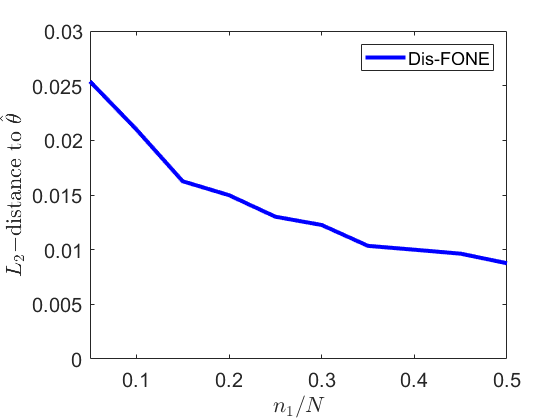}\label{fig:n14}}
	\caption{Comparison of $L_2$-errors when the sub-sample size of the first machine $n_1$ in Dis-FONE increases. The $x$-axis is the ratio of $n_1$ to the total sample size $N$. Here the total sample size $N=10^5$, the dimension $p=100$,  and the number of machines $L=20$. }
	\label{fig:n1} 
\end{figure}

\subsubsection{Effect on the unbalanced data partition}
\label{sec:n1}
In previous simulation studies, the entire dataset is evenly separated on different machines.  As one can see from Algorithm \ref{algo:distributed_FONE} and Theorem \ref{thm:d-fone-non}, the sub-sample size on the first machine $n_1$ plays a different role in Dis-FONE than those on the other machines $n_2,n_3,\dots, n_L$. In Figure \ref{fig:n1}, we investigate the effect of $n_1$ by varying $n_1$ from $N/L$ (the case of evenly distributed) to $10\times N/L$. Let the remaining data points be evenly distributed on the other machines, i.e., $n_2=n_3=\dots=n_L=(N-n_1)/(L-1)$.  We set $N=10^5$ and $L=20$. From Figure \ref{fig:n1}, the $L_2$-error of Dis-FONE gets much closer to ERM $\hat\tee$ in \eqref{eq:erm} when the largest sub-sample size $n_1$ increases, which is consistent with our theoretical results.

In Appendix G, we further investigate the cases of correlated design, the effect of the quality of the initial estimator, and different choices of the stepsizes (see Section G for more details). We also conduct simulations to compare our proposed methods (DC-SGD and Dis-FONE) to the existing methods in Section G.4 in terms of statistical accuracy and computation time. 
For logistic regression, we compare our algorithm with CSL \citep{jordan2016communication} for logistic regression.
Both Dis-FONE and CSL methods achieve nearly optimal performance as compared to the ERM, while Dis-FONE accelerates CSL. For quantile regression, we compare our methods with DC-QR \citep{volgushev2017distributed}. 
Dis-FONE outperforms DC-QR in terms of both computation time and statistical accuracy since DC-QR suffers from the restriction on the sub-sample size analogous to the case of DC-SGD. 

\subsubsection{Experiments on statistical inference}
\label{sec:exp_unitw}

\begin{table}[!t]
	\centering
	\caption{Left columns: Coverage rates and average confidence interval length in the brackets; Right columns: Square roots of the ratios of the estimated variance to the true limiting variance of ERM $\hat\tee$. The sample size $n\in\{10^5,2\times 10^5,5\times 10^5\}$ and dimension $p\in\{100,200,500\}$. The  multiplier $\tau_n=((p\log n)/n)^{1/2}$, the step-size $\eta_n=(p\log n)/n$ for logistic regression, and $\tau_{n}=((p\log n)/n)^{1/3}$, $\eta_n=((p\log n)/n)^{2/3}$ for quantile regression, respectively.}
	\vspace{.25cm}
	\begin{tabular}{cl|ccc|ccc}
		\hline
		Model &\qquad$n$ & \multicolumn{3}{c|}{Coverage rates (Avg length)} &\multicolumn{3}{c}{Square root ratio} \\
		&&$p=100$ & $p=200$ & $p=500$&$p=100$ & $p=200$ & $p=500$\\
		\hline
		\multicolumn{2}{l|}{Logistic}&&&&&\\
		& $n=10^5$ 			&94.37 (0.127) & 93.76 (0.174) & 91.67 (0.271) & 1.043 & 1.033 & 1.027\\
		& $n=2\times 10^5$ 	&94.49 (0.131) & 94.10 (0.168) & 92.14 (0.265) & 1.041 & 1.027 & 1.019\\
		& $n=5\times 10^5$ 	&94.71 (0.128) & 94.23 (0.167) & 92.31 (0.254) & 1.017 & 1.017 & 1.014\\
		\hline
		\multicolumn{2}{l|}{Quantile}&&&&&\\
		& $n=10^5$ 			&94.57 (0.160) & 94.11 (0.221) & 92.96 (0.358) & 1.042 & 1.023 & 1.027\\
		& $n=2\times 10^5$ 	&94.64 (0.159) & 94.24 (0.204) & 93.04 (0.349) & 1.007 & 1.004 & 1.004\\
		& $n=5\times 10^5$ 	&94.67 (0.154) & 94.59 (0.191) & 93.47 (0.344) & 1.005 & 1.002 & 1.003\\
		\hline
	\end{tabular}
	\label{table:sinvw}
\end{table}

In this section, we provide simulation studies for estimating $\S^{-1}\w$, where $\S$ is the population Hessian matrix of the underlying regression model and $\|\w\|_2=1$. As we illustrate in Section \ref{sec:unitw}, this estimator plays an important role in estimating the limiting variance of the ERM.  

In this experiment, we specify $\w=\boldsymbol{1}_p/\sqrt{p}$, the sample size $n\in\{10^5,2\times 10^5,5\times 10^5\}$, the dimension $p\in\{100,200,500\}$.  According to Theorems \ref{thm:sinvw} and \ref{thm:sinvw-non}, we set 
the multiplier $\tau_n=((p\log n)/n)^{1/2}$, the step-size $\eta_n=(p\log n)/n$ for logistic regression, and 
$\tau_{n}=((p\log n)/n)^{1/3}$, $\eta_n=((p\log n)/n)^{2/3}$ for quantile regression, respectively. 

Given $\hat{\S^{-1}\w}$, we are able to compute the estimator of limiting variance $\w'\S^{-1}\A\S^{-1}\w$ using \eqref{eq:imp}. Based on that and \eqref{eq:asy}, we  construct the 95\% confidence interval  for $\w'\tee^*$ as follows,
\begin{equation}\label{eq:conf}
\w'\hat\tee\pm \Phi^{-1}(0.975)\sqrt{\w'\S^{-1}\A\S^{-1}\w/n},
\end{equation}
where $\Phi(\cdot)$ is the CDF of the standard normal random variable. 

The left columns in Table \ref{table:sinvw} present the average coverage rates of the confidence intervals of $\w'\tee^*$ constructed by \eqref{eq:conf} and their average interval lengths. In the right columns of Table \ref{table:sinvw}, we report the square root of the ratio between the estimated variance and the true limiting variance, i.e., \[
\sqrt{(\widehat{\S^{-1}\w})'\hat\A(\widehat{\S^{-1}\w})\Big/\w'\S^{-1}\A\S^{-1}\w}.
\]
From Table \ref{table:sinvw}, our estimator achieves good performance for both logistic and quantile regression models. As the sample size $n$ increases, the coverage rates become closer to the nominal level and the ratios get closer to $1$.

\begin{figure}[!t]
	\centering
	\subfigure[$\tau=0.25$: $L_2$-distance to $\hat\tee$]{\includegraphics[width = 0.47\textwidth]{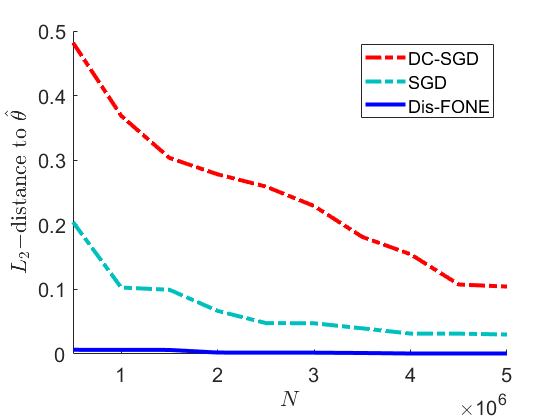}\label{fig:real_qr_1}}
	\subfigure[$\tau=0.5$: $L_2$-distance to $\hat\tee$]{\includegraphics[width = 0.47\textwidth]{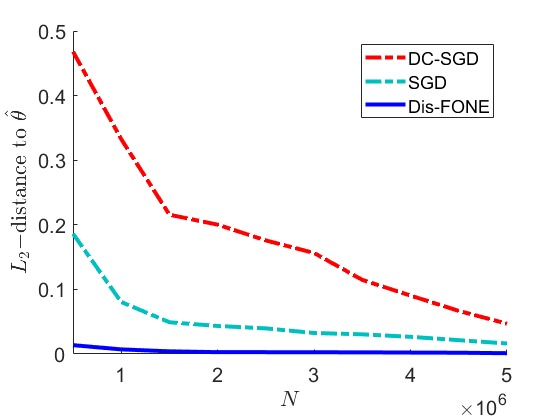}\label{fig:real_qr_2}}
	\caption{Real data analysis: comparison of $L_2$-errors when the sample size $N$ increases.} 
	\label{fig:real_qr}
	\vspace{-2mm}
\end{figure}
\subsection{Real Data Analysis - Census 2000 Data}\label{sec:real_data}
In this section, we provide real data analysis of our proposed methods. We consider the sampled U.S. 2000 Census dataset\footnote{U.S. Census, \url{http://www.census.gov/census2000/PUMS5.html}}, consisting of annual salary and related features on employed full-time wage and salary workers who reported that they worked 40 weeks or more and worked 35 hours or more per week. The U.S. 2000 Census dataset is a widely used dataset in quantile regression literature, see, e.g., \cite{angrist2006quantile,yang2013quantile}. The entire sample size is $5\times 10^6$ and the dimension $p=11$. We perform a quantile regression on the dataset, which treats the annual salary as the response variable, with two different quantile levels $\tau=0.25$ and $\tau=0.5$ (i.e., the least absolute deviations regression). We use $L=100$ machines/nodes and vary the total sample size $N$ from $5\times 10^5$ to $5\times 10^6$ to compare our Dis-FONE with DC-SGD in a distributed environment. In DC-SGD, we set $\alpha=1$ in the step-size $r_i=c_0/\max(i^\alpha,p)$. In  Dis-FONE, we set  $T=100$, $K=20$,  and step-size $\eta_n=c_0'm/n$. The constants $c_0$ and $c_0'$ is chosen in the say way as in the simulation study. More particularly, we choose the best $c_0$ and $c_0'$ that achieves the smallest objective function in \eqref{eq:erm}  with $\tee=\teeSGD^{(1)}$ and $\tee=\hat\tee_{\mathrm{dis},1}$ using data points from the first machine. The ERM estimator is computed by solving the quantile regression using the interior-point method with pooled data on a single powerful machine.

\begin{figure}[!t]
	\centering
	\subfigure[$\tau=0.25$: relative errors of objective values]{\includegraphics[width = 0.47\textwidth]{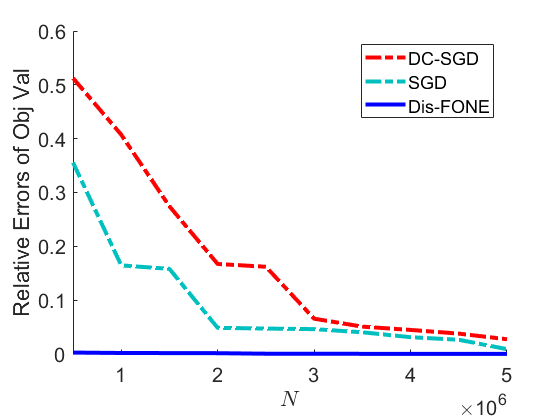}\label{fig:real_qr_3}}
	\subfigure[$\tau=0.5$: relative errors of objective values]{\includegraphics[width = 0.47\textwidth]{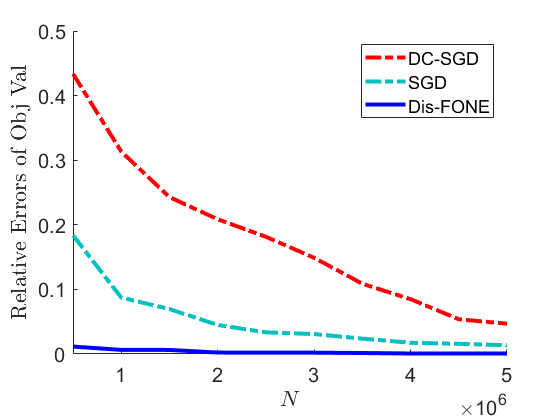}\label{fig:real_qr_4}}
	\caption{Real data analysis: comparison of the relative errors of objective values $|\hat{f}_{\text{method}}-\hat{f}_{\text{ERM}}|/|\hat{f}_{\text{ERM}}|$ on the testing data when the sample size $N$ increases.}
	\label{fig:real_qr_}
	\vspace{-2mm}
\end{figure}

From Figure \ref{fig:real_qr}, our proposed Dis-FONE $\teedis$ is very close to the ERM estimator and outperforms both DC-SGD and SGD estimators. As $N$ increases, both DC-SGD and SGD estimators are closer to the ERM estimator. In addition to estimation accuracy, we further investigate the performance on the testing set. For a given $N$, we split the data into the training set (with $0.8 N$ samples) and testing set (with $0.2N$). We estimate $\hat\tee_{\text{method}}$ using an estimation method, such as, DC-SGD, SGD, Dis-FONE, on the training set and evaluate the quantile objective value on the testing data. Denote the obtained quantile objective value on the testing set using a given method and the ERM by $\hat{f}_{\text{method}}$ and $\hat{f}_{\text{ERM}}$, respectively. 
We  report the relative errors of objective values $|\hat{f}_{\text{method}}-\hat{f}_{\text{ERM}}|/|\hat{f}_{\text{ERM}}|$ and from Figure \ref{fig:real_qr_}, the relative errors of Dis-FONE are very close to zero.

\section{Conclusions}
\label{sec:conc}

This paper studies general distributed estimation and inference problems based on stochastic subgradient descent. We propose an efficient First-Order Newton-type Estimator (FONE) for estimating $\S^{-1} \w$ and its distributed version. The key idea behind our method is to use stochastic gradient information to approximate the Newton step. We further characterize the theoretical properties when using FONE for distributed estimation and inference with both smooth and non-smooth loss functions. We also conduct numerical studies to demonstrate the performance of the proposed distributed FONE. The proposed FONE of  $\S^{-1} \w$ is a general estimator, which could find applications to other statistical estimation problems. While this paper focuses on convex loss functions, the proposed methods can be directly applied to non-convex objectives. It would be an interesting future direction to derive the convergence rates for non-convex settings.

\section*{Acknowledgement}
The authors are very grateful to anonymous referees and the associate editor for their detailed and constructive comments that considerably improved the quality of this paper. Xi Chen would like to thank the support from NSF via IIS-1845444. Weidong Liu is supported by National Program on Key Basic Research Project (973 Program, 2018AAA0100704), NSFC Grant No. 11825104 and 11690013, Youth Talent Support Program, and a grant from Australian Research Council.

\newpage
\appendix

\section{General Conditions and Their Verification}
{In this section, we first provide general versions of {\tt(A2-Log)} and {\tt(A2-QR)} beyond regression settings. In particular, the assumption {\tt(A2-Log)} for logistic regression is generalized to conditions {\tt(C2)} and {\tt(C3)}; and the assumption {\tt(A2-QR)} for quantile regression is generalized to conditions {\tt(C2)} and {\tt(C3$^*$)}.

Condition {\tt(C2)} is on the continuity of subgradient $g(\tee,\xii)$ and its expectation $G(\tee):=\ep g(\tee,\xii)$.
\noindent
{\bf \tt(C2).} 
Suppose that $G(\tee)$ is differentiable on $\tee$ and denote by $\S(\tee):= \nabla_{\tee} G(\tee)$.
For some constant $C_1>0$, we have 
\begin{eqnarray}\label{eq:diff_G}
\big\|\S(\tee_1)-\S(\tee_2)\big\|\leq C_1\|\tee_1-\tee_2\|_2\quad\mbox{for any $\tee_1,\tee_2 \in \R^p$.}
\end{eqnarray} 
Furthermore, let $\lambda_{\min}(\S(\tee))$ and $\lambda_{\max}(\S(\tee))$ be the minimum and maximum eigenvalue of $\S(\tee)$, respectively. We assume that $c_{1}\leq\lambda_{\min}(\S(\tee^*))\leq\lambda_{\max}(\S(\tee^*))\leq c^{-1}_{1}$ for some constant $c_{1}>0$.

\vspace{2mm}

Condition {\tt(C2)} can be easily verified in our two motivating examples of logistic regression and quantile regression (see Section \ref{sec:verify} below). It is worthwhile to note that $\S$ defined in \eqref{eq:sigma} is a brief notation for $\S(\tee^*)$.  The minimum eigenvalue condition on 
$\S(\tee^*)$ (i.e., $\lambda_{\min}(\S(\tee^*)) \geq c_1$) ensures that the population risk $F(\tee)$ is locally strongly convex at $\tee=\tee^*$.

\noindent {\bf\tt(C3)}. (For smooth loss function $f$) For $\v\in \R^{p}$, assume that $$\big|\v{'}[g( \tee_{1},\xii)-g(\tee_{2},\xii)]\big|\leq U(\v,\tee_{1},\tee_{2})\|\tee_{1}-\tee_{2}\|_2,$$  where $U(\v,\tee_{1},\tee_{2})$ satisfies that 
\[
\sup_{\|\v\|_2=1}\sup_{\tee_{1},\tee_{2}}\ep \exp\big(t_0 U(\v,\tee_{1},\tee_{2})\big)\leq C, \qquad \sup_{\|\v\|_2=1}  \ep \sup_{\tee_{1},\tee_{2}}U(\v,\tee_{1},\tee_{2})\leq p^{c_2},
\]
for some $c_2, t_0,C>0$. Moreover, 
\emph{one of the following two conditions} on $g(\tee, \xii)$  holds,
\begin{enumerate}
	\item  $\sup_{\|\v\|_2=1}\ep \sup_{\tee}\exp(t_0|\v'g(\tee,\xii)|)\leq C$ for some $t_0, C>0$;
	\item  $\sup_{\|\v\|_2=1}\ep \exp(t_0 |\v'g(\tee^{*},\xii)|)\leq C$ and $c_1\leq\lambda_{\min}(\S(\tee))\leq \lambda_{\max}(\S(\tee))\leq c_1^{-1}$ uniformly in $\tee$ for some $t_0, c_1, C>0$.
\end{enumerate} 

\noindent {\bf\tt (C3$^{*}$)}. (For non-smooth loss function $f$)  Suppose that for some constant $c_2, c_3,c_4>0$,
\begin{equation*}
	\sup_{\tee_{1}:\|\tee_{1}-\tee^*\|_2 \leq c_4}\ep\big{\{}\sup_{\tee_{2}:\|\tee_{1}-\tee_{2}\|_{2}\leq n^{-M},\; \|\tee_{2}-\tee^*\|_2 \leq c_4}\|g(\tee_1,\xii)-g(\tee_2,\xii)\|_2^{4}\Big{\}}\leq p^{c_2}n^{-c_3M}
\end{equation*}
for any large $M>0$. Also
\begin{eqnarray*}
	\sup_{\|\v\|_2=1}\ep (\v'(g(\tee_1,\xii)-g(\tee_2,\xii))^{2}\exp\{t_0|\v'(g(\tee_1,\xii)-g(\tee_2,\xii))|\}\leq C\|\tee_{1}-\tee_{2}\|_2
\end{eqnarray*}
and 
$
\sup_{\|\v\|_2=1}\ep \sup_{\tee}\exp(t_0|\v'g(\tee,\xii)|)\leq C 
$
for some $t_0$, $C>0$.

In Condition {\tt(C3)}, we only require either one of the two bullets of the conditions on $g(\tee, \xii)$ to hold. The second bullet in Condition {\tt(C3)} requires the moment condition on the subgradient $g(\tee^*, \xii)$ holds at the true parameter $\tee^*$, and weakens the uniform moment condition in the first bullet. On the other hand, it imposes an extra condition on the uniform eigenvalue bound of the matrix $\S(\tee)$, which is equivalent to assuming that the loss function $f$ is strongly convex on its entire domain. It is also worthwhile noting that the second bullet covers the case of linear regression where $\S(\tee)=\X\X'/n$ for all $\tee$.
}

\subsection{Verification of Conditions on Motivating Examples}
In this section, we provide verification of the conditions {\tt(C2)}, {\tt(C3)} and {\tt(C3$^*$)} on Examples \ref{ex:log} and \ref{ex:qr}.
\label{sec:verify}
\begin{customexp}{\ref{ex:log}}\label{rem:logistic}  For a logistic regression model with $\xii=(Y, \X)$,
	\begin{eqnarray*}
		\pr(Y=1|\X)=1-\pr(Y=-1|\X)=\frac{1}{1+\mathrm{exp}(-\X'\tee^*)}.
	\end{eqnarray*}
	We have $f(\tee,\xii)=\log(1+\exp(-Y\X'\tee))$, and $g(\tee,\xii)=\frac{-Y\X}{1+\exp(Y\X'\tee)}$. 
	Note that $G(\tee)=\ep\left(\frac{\X}{1+e^{-\X'\tee^*}}-\frac{\X}{1+e^{-\X'\tee}}\right)$ is  differentiable in $\tee$. Moreover, we have,
	\[
	\S(\tee)=\ep\frac{\X\X'}{[1+\exp({\X'\tee})][1+\exp(-{\X'\tee})]}.
	\]
\end{customexp}

\begin{proposition} In Example \ref{ex:log}, assume that $\tilde{c}_{1}\leq \lambda_{\min}(\ep(\X\X^{'}))\leq \lambda_{\max}(\ep(\X\X^{'}))\leq \tilde{c}^{-1}_{1}$ for some $\tilde{c}_{1}>0$
	and $\sup_{\|\v\|_{2}=1}\ep |\v^{'}\X|^{3}\leq \tilde{C}_{1}$ for some $\tilde{C}_{1}>0$.
	
	\begin{enumerate}
		\item[(1)]
		We have $\lambda_{\max}(\S(\tee))$ is bounded uniformly in $\tee$. Furthermore, if $\|\tee^{*}\|_{2}\leq \tilde{C}_{2}$, then $\lambda_{\min}(\S(\tee^{*}))\geq c_{1}$
		for some $c_{1}>0$ and (C2) holds.
		
		\item[(2)] If  the covariates $\X$ satisfy $\sup_{\|\v\|_{2}=1}\ep \exp(t_{0} (\v^{'}\X)^{2})\leq \tilde{C}_{2}$ for some $t_{0},\tilde{C}_{2}>0$,  then (C3) holds.
	\end{enumerate}
\end{proposition}

\noindent{\bf Proof.} Note that
\begin{eqnarray*}
	\|\S(\tee)\|^{2}_{2}=\sup_{\|\v\|_{2}=1}\ep \frac{(\v^{'}\X)^{2}}{[1+\exp({\X'\tee})][1+\exp(-{\X'\tee})]}\leq \sup_{\|\v\|_{2}=1}\ep(\v^{'}\X)^{2}\leq \tilde{c}^{-1}_{1}.
\end{eqnarray*}
That is, $\lambda_{\max}(\S(\tee))$ is bounded uniformly in $\tee$. Also,
\begin{eqnarray*}
	\lambda_{\min}(\S(\tee^{*}))&=&\min_{\|\v\|_{2}=1}\ep \frac{(\v^{'}\X)^{2}}{[1+\exp({\X'\tee^{*}})][1+\exp(-{\X'\tee^{*}})]}\cr
	&\geq&\min_{\|\v\|_{2}=1}\ep \frac{(\v^{'}\X)^{2}}{2(1+e^{M})}I\{|\X^{'}\tee^{*}|\leq M\}\cr
	&=&\frac{1}{2(1+e^{M})}\min_{\|\v\|_{2}=1}\Big{(}\ep (\v^{'}\X)^{2}-\ep (\v^{'}\X)^{2}I\{|\X^{'}\tee^{*}|> M\}\Big{)}\cr
	&\geq&\frac{1}{2(1+e^{M})}\min_{\|\v\|_{2}=1}\Big{(}\ep (\v^{'}\X)^{2}-\frac{\tilde{C}_{1}\tilde{C}_{2}}{M}\Big{)}\cr
	&=&\frac{1}{2(1+e^{M})}\Big{(}\tilde{c}_{1}-M^{-1}\tilde{C}_{1}\tilde{C}_{2}\Big{)}.
\end{eqnarray*}
Now let $M$ be a constant that satisfies $M>\tilde{C}_{1}\tilde{C}_{2}/\tilde{c}_{1}$. This yields that $\lambda_{\min}(\S(\tee^{*}))\geq c_{1}$ for some $c_{1}>0$.
By noting that the derivative of $(1+e^{x})^{-1}(1+e^{-x})^{-1}$ is bounded by $3$, we have 
\begin{eqnarray*}
	\|\S(\tee_{1})-\S(\tee_{2})\|\leq 3\sup_{\|\v\|_{2}=1}\ep (\v^{'}\X)^{2}|\X^{'}(\tee_{1}-\tee_{2})|\leq 3\tilde{C}_{1}\|\tee_{1}-\tee_{2}\|_{2}.
\end{eqnarray*}
This proves  (C2). Similarly, the derivative of $(1+e^{x})^{-1}$ is bounded by 1, and hence for $\tee_{1}\neq \tee_{2}$,
\begin{eqnarray*}
	|\v^{'}g(\tee_{1},\xii)-\v^{'}g(\tee_{2},\xii)|\leq |\v^{'}\X||\X^{'}(\tee_{1}-\tee_{2})|\leq U(\v,\tee_{1},\tee_{2})\|\tee_{1}-\tee_{2}\|_{2},
\end{eqnarray*}
where $U(\v,\tee_{1},\tee_{2})= |\v^{'}\X||\X^{'}(\tee_{1}-\tee_{2})|/\|\tee_{1}-\tee_{2}\|_{2}$. It is easy to see that
\[
\sup_{\|\v\|_2=1}\sup_{\tee_1,\tee_2}\ep\exp(t_0U(\v,\tee_1,\tee_2))\leq \sup_{\|\v\|_2=1}\ep\exp(t_0(\v'\X)^2)\leq \tilde{C}_2,
\]
and 
\[
\sup_{\|\v\|_2=1}\ep\sup_{\tee_1,\tee_2}U(\v,\tee_1,\tee_2)\leq\sup_{\|\v\|_2=1}\ep\sup_{\|\tee\|_2=1}|\v'\X||\tee'\X|\leq Cp.
\]
Therefore, $U(\v,\tee_{1},\tee_{2})$ satisfies (C3). 
At the meantime, since $|\v' g(\tee,\xii)|\leq |\v'\X|$ for any $\tee$, and $\sup_{\|\v\|_{2}=1}\ep \exp(t_{0} (\v^{'}\X)^{2})\leq \tilde{C}_{2}$ for some $t_{0},\tilde{C}_{2}>0$, bullet (1) of (C3) holds for $g(\tee,\xii)$.\qed

\begin{customexp}{\ref{ex:qr}}\label{rem:quantile}  For a quantile regression model,
	\begin{eqnarray*}
		y=\X'\tee^*+\epsilon,   \qquad \pr(\epsilon\leq 0|\X)=\tau.
	\end{eqnarray*} 
	We have the non-smooth quantile loss $f(\tee,\xii)=\ell(y-\X'\tee)$ with $\ell(x)=x(\tau-I\{x\leq 0\}))$, and its subgradient $g(\tee,\xii)=\X(I\{y\leq \X'\tee\}-\tau)$.
	Then $G(\tee)=\ep[\X(\pr(\epsilon\leq \X'(\tee-\tee^*)|\X)-\tau)]$. 
	Furthermore, we have  $\S(\tee)=\ep[\X\X'\rho_{\X}(\X'(\tee-\tee^*))]$,  where $\rho_{\X}(\cdot)$ is the density function of $\epsilon$ given $\X$. 
\end{customexp}

\begin{proposition}  Assume that 
	\[
	c_{1}\leq \lambda_{\min}(\ep[\X\X'\rho_{\X}(0)])\leq \lambda_{\max}(\ep[\X\X'\rho_{\X}(0)])\leq c^{-1}_{1}
	\]
	for some $c_{1}>0$ and $\sup_{\|\v\|_{2}=1}\ep |\v^{'}\X|^{3}\leq \tilde{C}_{1}$ for some $\tilde{C}_{1}>0$.
	The density function $\rho_{\X}(x)$ is bounded and satisfies $|\rho_{\X}(x_{1})-\rho_{\X}(x_{2})|\leq \tilde{C}|x_{1}-x_{2}|$ for some $\tilde{C}>0$.   Then (C2) holds. Furthermore, 
	if the covariates $\X$ satisfy $\sup_{\|\v\|_{2}=1}\ep \exp(t_{0} |\v^{'}\X|)\leq \tilde{C}_{2}$, then (C3$^{*}$) holds.
\end{proposition}

\noindent{\bf Proof}. By the Lipschitz condition on $\rho_{\X}(x)$, we have
\begin{eqnarray*}
	\|\S(\tee_{1})-\S(\tee_{2})\|_{2}\leq \tilde{C}\sup_{\|\v\|_{2}=1}\ep (\v^{'}\X)^{2}|\X^{'}(\tee_{1}-\tee_{2})|\leq  \tilde{C}_{1}\tilde{C}\|\tee_{1}-\tee_{2}\|_{2}.
\end{eqnarray*}
Hence (C2) holds. Now we prove (C3$^{*}$). Since $\rho_{\X}(x)$ is bounded, we have
\begin{eqnarray*}
	&&\ep\big[\sup_{\tee_{2}:\|\tee_{1}-\tee_{2}\|_2\leq n^{-M},\; \|\tee_{2}-\tee^*\|_2 \leq c_4}\|g(\tee_1,\xii)-g(\tee_2,\xii)\|_2^{4}\big]\cr
	&&\leq\ep\Big[\|\X\|_2^4 I\big\{|\X'(\tee_1-\tee^*)|-\|\X\|_2n^{-M}\leq \epsilon\leq |\X'(\tee_1-\tee^*)|+\|\X\|_2n^{-M}\big\}\Big]\cr
	&&\leq 2\tilde{C}\ep\big[\|\X\|_2^5 n^{-M}\big]\cr
	&&\leq 2\tilde{C}p^{5/2}n^{-M}.
\end{eqnarray*}
Again, 
\begin{eqnarray*}
	&&\ep (\v'(g(\tee_1,\xii)-g(\tee_2,\xii))^{2}\exp\{t_{0}|\v'(g(\tee_1,\xii)-g(\tee_2,\xii))|\}\cr
	&&\quad\leq\ep \Big{[}(\v'\X)^{2} (I\{ \epsilon\leq \X^{'}(\tee_{1}-\tee^{*})\}-I\{ \epsilon\leq \X^{'}(\tee_{2}-\tee^{*})\})^2\exp\{t_{0}|\v'\X|\}\Big{]}\cr
	&&\quad\leq \ep \Big{[}(\v'\X)^{2}I\big\{\X^{'}(\tee_{1}-\tee^{*})\leq \epsilon\leq \X^{'}(\tee_{2}-\tee^{*})\big\}\exp\{t_{0}|\v'\X|\}\Big{]}\cr
	&&\qquad +\ep \Big{[}(\v'\X)^{2}I\big\{\X^{'}(\tee_{2}-\tee^{*})\leq \epsilon\leq \X^{'}(\tee_{1}-\tee^{*})\big\}\exp\{t_{0}|\v'\X|\}\Big{]}\cr
	&&\quad\leq 2\tilde{C}\ep \Big{[}(\v'\X)^{2}|\X^{'}(\tee_{1}-\tee_{2})|\exp\{t_{0}|\v'\X|\}\Big{]}\cr
	&&\quad\leq C\|\tee_{1}-\tee_{2}\|_{2},
\end{eqnarray*}
for some $C>0$.
This ensures that (C3$^{*}$) holds.\qed

\section{Theory of Mini-batch SGD}
\label{sec:theory_sgd}
Before we prove the theory of DC-SGD, we first provide some theoretical results of the mini-batch SGD in the diverging $p$ case. In particular, let $\ep_{0}(\cdot)$ be the expectation to $\{\xii_{i},1\leq i\leq n\}$ given the initial estimator $\hat\tee_{0}$. Let us denote the solution of mini-batch SGD in \eqref{eq:SGD_mini} with $s=n/m$ iterations by $\teeSGD$. We obtain the consistency result of the mini-batch SGD in the diverging $p$ case. Recall that $r_{i}=c_{0}/\max(i^{\alpha},p)$ for some $0<\alpha\leq 1$ and $c_{0}$ is a sufficiently large constant.

\begin{theorem} \label{thm:sgd}Assume  (C1), (C2), (C3) or (C3$^*$) hold and the initial estimator $\hat\tee_0$ is independent to $\{\xii_i, i=1,2,\dots, n\}$.  On the event $\{\|\hat\tee_{0}-\tee^*\|_2\leq d_n\}$ with $d_n\rightarrow 0$, the mini-batch SGD estimator satisfies 
	\[	
	\ep_{0}\|\teeSGD-\tee^*\|_2^{2}=O(\frac{p}{m^{1-\alpha}n^{\alpha}}) \quad \text{and} \quad \|\ep_{0}(\teeSGD)-\tee^*\|_2= O(\frac{p}{m^{1-\alpha}n^{\alpha}}).
	\] Furthermore, if 
	$\|\hat\tee_{0}-\tee^*\|_2=o_{\pr}(1)$,  then  $\|\teeSGD-\tee^*\|_2^{2}=O_{\pr}(\frac{p}{m^{1-\alpha}n^{\alpha}})$.
\end{theorem}

Theorem \ref{thm:sgd} characterizes both the mean squared error and the bias of the obtained estimator from SGD. When the decaying rate of the step-size $\alpha=1$, the convergence rate is not related to $m$, and it  achieves the same rate as the ERM $\hat\tee$ in \eqref{eq:erm} (i.e., $O(\sqrt{p/n})$).

We note that Theorem \ref{thm:sgd} requires a consistent initial estimator $\hat\tee_0$. In practice, we can always use a small separate subset of samples to construct the initial estimator by minimizing the empirical risk.

In contrast to  the fixed $p$ setting where an arbitrary initialization can be used, a consistent initial estimator is almost necessary to ensure the convergence in the diverging $p$ case, which is shown in the following proposition:

\begin{proposition}\label{prop:sgd1} Assume that the initial estimator $\hat\tee_{0}$ is independent to $\{\xii_i,i=1,2,\dots, n\}$ and satisfies $\ep\|\hat\tee_{0}-\tee^*\|^{2}_{2}\geq p^{2\nu}$ for some $\nu>0$, the step-size $r_{i}\leq C/i^{\alpha}$ for some $0<\alpha\leq 1$ and the batch size $m\geq 1$. Suppose that
	$\sup_{\|\v\|_{1}=1}\sup_{\tee}\ep(\v' g(\tee,\xii))^{2}\leq C$.  We have $\ep\|\teeSGD-\tee^*\|_2^{2}\geq Cp^{2\nu}$ for all $n\leq m\exp(o(p^{\nu}))$ when $\alpha=1$ and for all $n=o(mp^{\nu/(1-\alpha)})$ when $0<\alpha<1$.
\end{proposition}

We note that Proposition \ref{prop:sgd1} provides a lower bound result, which shows that in the diverging $p$ case, a standard mini-batch SGD with a random initialization will not converge with high probability.  Indeed, a random initial estimator $\hat\tee_{0}$ will incur an error $\ep\|\hat\tee_{0}-\tee^*\|_{2}^{2}\approx p$.  When $p=n^{\kappa}$ for some $\kappa>0$, the exponential relationship  $n\leq m\exp(o(p^{\nu}))$ holds with $\nu=0.5$ and thus Proposition \ref{prop:sgd1} implies that $\teeSGD$  has a large mean squared error that is at least on the order of ${p}$.   Proposition \ref{prop:sgd1} indicates that a good initialization is crucial for SGD when $p$ is diverging along with $n$.
\subsection{Proof of Theorem \ref{thm:sgd}}
\label{suppsec:thmsgd1}
By Condition (C1), we have $\sqrt{(p\log n)/m}\rightarrow 0$. 
Without loss of generality, we can assume that $\sqrt{(p\log n)/m}=o(d_n)$. Let $\de_{i}=\z_{i}-\tee^*$ and $\bar{g}(\tee,\xii)=g(\tee,\xii)-\ep g(\tee,\xii)$. Define $$\Theta_{0}=\{\tee\in \R^{p}: \|\tee-\tee^*\|_2\leq c_4\},$$
where $c_4$ is given in \eqref{eq:Theta0}. Define the events
$\mathcal{F}_{i}=\{\|\de_{i-1}\|_2\leq d_n\}$, and
\begin{eqnarray*}
	\mathcal{C}_{i}&=&\Big{\{}\sup_{\tee\in\Theta_{0}}\Big{\|}\frac{1}{m}\sum_{j\in H_i}\bar{g}(\tee,\xii_j)\Big{\|}_2\leq C\sqrt{\frac{p\log n}{m}}\Big{\}},
\end{eqnarray*}
where $C$ is sufficiently large. From the SGD updating rule \eqref{eq:SGD_mini}, we have
\begin{equation}\label{eq_extract}
\|\de_{i}\|_2^{2}=\|\de_{i-1}\|_2^{2}-2\frac{r_{i}}{m}\sum_{j\in H_{i}}\de'_{i-1}g(\z_{i-1},\xii_j)+\Big{\|}\frac{r_i}{m}\sum_{j\in H_{i}}g(\z_{i-1},\xii_j)\Big{\|}^{2}_2.
\end{equation}
Note that by the mean value theorem (see \citet[Chapter XIII, Theorem 4.2]{lang2012real}), $G(\z_{i-1})-G(\tee^{*})=\int_{0}^{1}\S(\tee^{*}+t\de_{i-1})dt\de_{i-1}=:\tilde{\S}(\z_{i-1})\de_{i-1}$ and $\|\tilde{\S}(\z_{i-1})-\S(\tee^{*})\|\leq C\|\de_{i-1}\|_{2}$ by (C2).
Since $G(\tee)=\ep g(\tee,\xii)$ and $G(\tee^*)=0$ by \eqref{eq:SA},
\begin{eqnarray*}\label{a1}
	\frac{r_i}{m}\sum_{j\in H_{i}}\de'_{i-1}{g}(\z_{i-1},\xii_j)&=&r_i\de'_{i-1}G(\z_{i-1})+\frac{r_i}{m}\sum_{j\in H_{i}}\de'_{i-1}\bar{g}(\z_{i-1},\xii_j)\cr
	&=&r_i\de'_{i-1}\Big(G(\z_{i-1})-G(\tee^*)\Big)+\frac{r_i}{m}\sum_{j\in H_{i}}\de'_{i-1}\bar{g}(\z_{i-1},\xii_j)\cr
	&\geq&c_1r_i\|\de_{i-1}\|_2^2-C_1r_i\|\de_{i-1}\|_2^3-r_i\big\|\de_{i-1}\big\|_2\Big\|\frac{1}{m}\sum_{j\in H_{i}}\bar{g}(\z_{i-1},\xii_j)\Big\|_2.
\end{eqnarray*}
Similarly,
\begin{eqnarray*}
	\frac{r_i}{m}\sum_{j\in H_{i}}g(\z_{i-1},\xii_j)&=&r_i\big(G(\z_{i-1})-G(\tee^*)\big)+\frac{r_i}{m}\sum_{j\in H_{i}}\bar{g}(\z_{i-1},\xii_j)
\end{eqnarray*}
and
\begin{eqnarray*}
	\Big{\|}\frac{r_{i}}{m}\sum_{j\in H_{i}}g(\z_{i-1},\xii_j)\Big{\|}^{2}_2&\leq &2r_i^2\|G(\z_{i-1})-G(\tee^*)\|_2^2+2\Big{\|}\frac{r_{i}}{m}\sum_{j\in H_{i}}\bar{g}(\z_{i-1},\xii_j)\Big{\|}^{2}_2\cr
	&\leq &r_i^2c_1^{-2}\|\de_{i-1}\|_{2}^2+C_1^2r_i^2\|\de_{i-1}\|_{2}^4+2\Big{\|}\frac{r_{i}}{m}\sum_{j\in H_{i}}\bar{g}(\z_{i-1},\xii_j)\Big{\|}^{2}_2.
\end{eqnarray*}
Therefore, on $\mathcal{C}_{i}\cap \mathcal{F}_{i}$, since $\sup_{i\geq 1}r_i=o(1)$,
\begin{eqnarray*}
	\|\de_{i}\|_2^{2}&\leq& (1- c_1r_i)\|\de_{i-1}\|_{2}^2+C\Big{(}r_id_n\sqrt{\frac{p\log n}{m}}+r_i^2\frac{p\log n}{m}+r_id_n^3+r_i^2d_n^2+r_i^2d_n^4\Big{)}\cr
	&\leq&(1- c_1r_i/2)d_n^2+Cr_i\frac{p\log n}{m},
\end{eqnarray*}
where we used the inequality $d_n\sqrt{\frac{p\log n}{m}}\leq t d^{2}_{n}+t^{-1}(p\log n)/m$ for any small $t>0$.
Note that $\sqrt{(p\log n)/m}=o(d_n)$. Therefore, $ \mathcal{F}_{i}\cap\mathcal{C}_i\subset \{\|\de_i\|_2\leq d_n\}= \mathcal{F}_{i+1}$. 
Combining the above arguments for $j=1,2,\dots, i$,  on the event $\{\|\hat\tee_0-\tee^*\|_2\leq d_n\}\cap (\cap_{k=1}^i\mathcal{C}_k)$, we have $\max_{1\leq j\leq i}\|\de_j\|_2\leq d_n$. 

We now assume $\sup_{\|\v\|_2=1}\ep \sup_{\tee}\exp(t_{0}|\v'g(\tee,\xii)|)\leq C$ (by Condition (C3) bullet 1 or (C3$^*$)). We have
\begin{eqnarray*}
	\|\de_{i}\|_{2}\leq \|\de_{i-1}\|_{2}+\frac{r_{i}}{m}\sum_{j\in H_{i}}\sup_{\tee}\|g(\tee,\xii_{j})\|_{2}\leq C\frac{1}{m}\sum_{j=1}^{n}\sup_{\tee}\|g(\tee,\xii_{j})\|_{2}.
\end{eqnarray*}
Thus $\ep_{0}\|\de_{i}\|^{6}_{2}\leq Cn^{6}$ and $\ep_{0}\|\de_{i}\|^{8}_{2}\leq Cn^{8}$.
Recall that $\ep_0(\cdot)$ is denoted by the expectation to $\{\xii_i\}$ given the initial estimator $\hat\tee_0$. By $G(\tee^*)=0$,
\begin{eqnarray*}
	\ep_0 \Big[\frac{1}{m}\sum_{j\in H_{i}}\de'_{i-1}g(\z_{i-1},\xii_j)\Big{]}=
	\ep_{0} \big[\de'_{i-1}G(\z_{i-1})\big]=\ep_{0}\big[\de'_{i-1}\tilde{\S}(\z_{i-1})\de_{i-1}\big],
\end{eqnarray*}
Then on the event $\{\|\hat\tee_0-\tee^*\|_2\leq d_n\}$, by (C2), Lemma \ref{lem:ct} (see Section \ref{sec:tech_lemma} below), and $\ep_0\|\de_{i-1}\|^{6}_{2}\leq Cn^{6}$,
\begin{eqnarray}\label{acd}
\ep_0\big[\de'_{i-1}\tilde{\S}(\z_{i-1})\de_{i-1}\big]&\geq& \ep_0\big[\de'_{i-1}\S\de_{i-1}\big]-C_1d_n\ep_0\|\de_{i-1}\|^2_2I\{\cap_{k=1}^i\mathcal{C}_k\}\cr
&&-\ep_0\big[\big{|}\de'_{i-1}\big(\tilde{\S}(\z_{i-1})-\S\big)\de_{i-1}\big{|}\big]I\{\{\cap_{k=1}^i\mathcal{C}_k\}^c\}\cr
&\geq & c_1\ep_0\|\de_{i-1}\|_2^2-C_1d_n\ep_0\|\de_{i-1}\|_2^2-C_1\ep_0\big{[}\|\de_{i-1}\|_2^3I\{\{\cap_{k=1}^i\mathcal{C}_i\}^c\}\big{]}\cr
&\geq&2^{-1}c_1\ep_0\|\de_{i-1}\|_2^2-C_1\ep_0\big{[}\|\de_{i-1}\|_2^3I\{\{\cap_{k=1}^i\mathcal{C}_i\}^c\}\big{]}\cr
&\geq&2^{-1}c_1\ep_0\|\de_{i-1}\|_2^2-Cn^{3-\gamma}
\end{eqnarray}
for any $\gamma>0$.
Also,  on the event $\{\|\hat\tee_0-\tee^*\|_2\leq d_n\}$,
\begin{eqnarray}\label{eq:a00}
\ep_0\Big{\|}\frac{1}{m}\sum_{j\in H_{i}}g(\z_{i-1},\xii_j)\Big{\|}^{2}_2&= &\ep_0\|G(\z_{i-1})\|_2^2+\ep_0\Big{\|}\frac{1}{m}\sum_{j\in H_{i}}\bar{g}(\z_{i-1},\xii_j)\Big{\|}^{2}_2\cr
&\leq& c_1^{-1}\ep_0\|\de_{i-1}\|_2^2+C\ep_0\|\de_{i-1}\|_2^4+\frac{Cp}{m}.
\end{eqnarray}
Moreover, on the event $\{\|\hat\tee_0-\tee^*\|_2\leq d_n\}$,
\begin{eqnarray*}
	\ep_0\|\de_{i-1}\|_2^4&\leq& d_n^2\ep_0\|\de_{i-1}\|_2^2+(\ep_0\|\de_{i-1}\|_2^8\cdot\pr_{0}\{\|\de_{i-1}\|_2>d_n\})^{1/2}\cr
	&\leq& d_n^2\ep_0\|\de_{i-1}\|_2^2+(\ep_0\|\de_{i-1}\|_2^8\cdot\pr(\cup_{k=1}^{i-1}\mathcal{C}^{c}_{k}))^{1/2}\cr
	&\leq& d_n^2\ep_0\|\de_{i-1}\|_2^2+C n^{4-\gamma}
\end{eqnarray*} 
for any $\gamma>0$.
Therefore, on the event $\{\|\hat\tee_0-\tee^*\|_2\leq d_n\}$,
\begin{eqnarray}\label{eq_extract2}
\ep_0\big[\|\de_{i}\|_2^{2}\big]&\leq &(1-cr_i/2)\ep_0\|\de_{i-1}\|_2^{2}+Cr^{2}_i\frac{p}{m},
\end{eqnarray}
which by Lemma \ref{lem:an} (see Section \ref{sec:tech_lemma} below) implies that for any $\tau>0$, $\gamma>0$ and all $i\geq p^{1/\alpha+\tau}$, $\ep_{0} \|\de_{i}\|_2^{2}\leq C_{1}(p/(i^{\alpha}m)+i^{-\gamma})$.  By (C1), we have
$s=n/m\geq p^{1/\alpha+\tau_{2}}$. That is, $\ep_{0} \|\de_{s}\|_2^{2}\leq C_{1}p/(n^{\alpha}m^{1-\alpha})$.

Now consider the setting that Condition (C3) holds with bullet 2:  $c_1\leq\lambda_{\min}(\S(\tee))\leq \lambda_{\max}(\S(\tee))\leq c_1^{-1}$ uniformly in $\tee$.  Then we have $c_1\leq\lambda_{\min}(\tilde{\S}(\z_{i-1}))\leq\lambda_{\max}(\tilde{\S}(\z_{i-1}))\leq c_1^{-1}$. Therefore
\begin{eqnarray*}
	\ep_0 \Big[\frac{1}{m}\sum_{j\in H_{i}}\de'_{i-1}g(\z_{i-1},\xii_j)\Big{]}=\ep_{0}\big[\de'_{i-1}\tilde{\S}(\z_{i-1})\de_{i-1}\big]\geq c_1\ep_0\|\de_{i-1}\|_2^2.
\end{eqnarray*}
Also by (C3),
\begin{eqnarray*}
	\ep_{0}\big{\|}\bar{g}(\z_{i-1},\xii_j)\big{\|}^{2}_2&=&\ep_{0}\Big{(}\ep_{0}\big{[}\big{\|}\bar{g}(\z_{i-1},\xii_j)\big{\|}^{2}_2\big{|}\z_{i-1}\big{]}\Big{)}\cr
	&\leq&2\ep_{0}\Big{(}\ep_{0}\big{[}\big{\|}\bar{g}(\z_{i-1},\xii_j)-\bar{g}(\tee^{*},\xii_{j})\big{\|}^{2}_2\big{|}\z_{i-1}\big{]}\Big{)}+Cp\cr
	&\leq&Cp\ep_{0}\|\de_{i-1}\|^{2}_{2}+Cp.
\end{eqnarray*}
So
\begin{eqnarray*}
	\ep_0\Big{\|}\frac{1}{m}\sum_{j\in H_{i}}g(\z_{i-1},\xii_j)\Big{\|}^{2}_2&\leq& 2\ep_0\|G(\z_{i-1})\|_2^2+2\ep_0\Big{\|}\frac{1}{m}\sum_{j\in H_{i}}\bar{g}(\z_{i-1},\xii_j)\Big{\|}^{2}_2\cr
	&\leq&C\ep_{0}\|\de_{i-1}\|^{2}_{2}+C\frac{p}{m}\ep_{0}\|\de_{i-1}\|^{2}_{2}+C\frac{p}{m}.
\end{eqnarray*}
That is, (\ref{eq_extract2}) still holds and  $\ep_{0} \|\de_{s}\|_2^{2}\leq Cp/(n^{\alpha}m^{1-\alpha})$.

We now consider the bias of $\ep_{0}\z_{i}$.
We have
\begin{eqnarray*}
	\ep_0\Big[\frac{1}{m}\sum_{j\in H_{i}}g(\z_{i-1},\xii_j)\Big]= \ep_{0}\Big{(}G(\z_{i-1})-G(\tee^*)\Big{)}=\S\ep_{0}\de_{i-1}+\ep_{0}(\tilde{\S}(\z_{i-1})-\S)\de_{i-1}
\end{eqnarray*}
and $\|(\tilde{\S}(\z_{i-1})-\S)\de_{i-1}\|_{2}\leq C\|\de_{i-1}\|^{2}_{2}$.
Therefore, on the event $\{\|\hat\tee_0-\tee^*\|_2\leq d_n\}$, for any $\tau>0$, $0<\mu<1$ and $i\geq \max(p^{1/\alpha+\tau/2},(n/m)^{\mu})$,
\begin{eqnarray*}
	\|\ep_0\de_i\|_2&\leq& \|\boldsymbol{I}-r_i\S\|\|\ep_0\de_{i-1}\|_2+Cr_i\ep_0\|\de_{i-1}\|_2^2\cr
	&\leq&(1-c_{1}r_{i})\|\ep_0\de_{i-1}\|_2+Cr_i\ep_0\|\de_{i-1}\|_2^2\cr
	&\leq&(1-c_{1}r_{i})\|\ep_0\de_{i-1}\|_2+Cr^{2}_i \frac{p}{m}+Ci^{-\gamma}\cr
	&\leq&(1-c_{1}r_{i})\|\ep_0\de_{i-1}\|_2+Cr^{2}_i \frac{p}{m},
\end{eqnarray*}
by noting that $\gamma>0$ can be arbitrarily large.
Let $q_{\alpha}= \max(p^{1/\alpha+\tau/2},(n/m)^{\mu})$. Then for  any $\gamma>0$,
\begin{eqnarray*}
	\|\ep_0\de_s\|_2&\leq& \prod_{j=q_{\alpha}+1}^{s}(1-c_{1}r_{j})\|\ep_0\de_{q_{\alpha}}\|_2+\frac{p}{m}\sum\limits_{k=q_{\alpha}+1}^{s}r_{k}^2\prod_{j=k}^{i-1}(1-cr_{j+1})\cr
	&\leq& C(q_{\alpha}/s)^{\tilde{c}}+Cr_{s}\frac{p}{m}+Cs^{-\gamma},
\end{eqnarray*}
where $\tilde{c}$ is sufficiently large.
Therefore, by Lemma \ref{lem:an}, $ \|\ep_0\de_s\|_2\leq C_{1}p/(n^{\alpha}m^{1-\alpha})$. 
\qed

\subsection{Proof of Proposition \ref{prop:sgd1}}\label{suppsec:propsgd1}
Since $\sup_{\|\v\|_2=1}\sup_{\tee}\ep(\v'g(\tee,\xii))^{2}\leq C$, by the independence between $\xii_{j}$ and $\z_{i-1}$, we have $\ep(\de'_{i-1}g(\z_{i-1},\xii_j))^{2}\leq C\|\de_{i-1}\|_2^2$.
By \eqref{eq_extract}, we have
\begin{eqnarray*}
	\ep \|\de_{i}\|_{2}^{2}&\geq& \ep \|\de_{i-1}\|^{2}_{2}-\frac{2r_{i}}{m}\sum_{j\in H_{i}}\ep \de'_{i-1}g(\tee_{i-1},\xii_j)\cr
	&\geq&\ep \|\de_{i-1}\|_{2}^{2}-Cr_{i}\sqrt{\ep \|\de_{i-1}\|^{2}_{2}}\cr
	&\geq& \min\Big{(}(1-Cr_{i}/p^{\nu})p^{2\nu},\ep \|\de_{i-1}\|_{2}^{2}-Cr_{i}p^{\nu}\Big{)}\cr
	&\geq&\min\Big{(}(1-Cr_{i}/p^{\nu})p^{2\nu},(1-Cr_{i-1}/p^{\nu})p^{2\nu}-Cr_{i}p^{\nu},\cr
	&&\quad\ep \|\de_{i-2}\|_{2}^{2}-Cr_{i}p^{\nu}-Cr_{i-1}p^{\nu}\Big{)}\cr
	&\geq&(1-C/p^{\nu})p^{2\nu}-C\sum_{j=1}^{i}r_{j}p^{\nu}.
\end{eqnarray*}
Note that $\sum_{j=1}^{i}r_{j}=O(i^{1-\alpha})$ when $0<\alpha<1$ and $\sum_{j=1}^{i}r_{j}=O(\log i)$ when $\alpha=1$,
So if $\alpha=1$ and $\log (n/m)=o(p^{\nu})$, or  if $0<\alpha<1$ and $n/m=o(p^{\nu/(1-\alpha)})$, we have $\ep \|\de_{s}\|_2^{2}\geq Cp^{2\nu}$.
\qed

\section{Proofs for Results of DC-SGD in Section \ref{sec:th-sgd}}
\label{sec:proof_dcsgd}

\subsection{Proof of Theorem \ref{th_dc}}
\label{sec:proof_dcsgd1}
\begin{customthm}{\ref{th_dc}} Assume (C1), (C2), (C3) or (C3$^*$) hold, suppose the initial estimator $\hat\tee_0$ is independent to $\{\xii_i, i=1,2,\dots, N\}$.
		On the event $\{\|\hat\tee_{0}-\tee^*\|_2\leq d_n\}$ with $d_n\rightarrow 0$, the DC-SGD estimator achieves the following convergence rate:
		\begin{equation}
		\ep_{0}\|\teeDC-\tee^*\|_2^{2}=O\left(\frac{p}{L^{1-\alpha}m^{1-\alpha}N^{\alpha}}  +\frac{p^{2}L^{2\alpha}}{m^{2-2\alpha}N^{2\alpha}}\right).
		\end{equation}
\end{customthm}
\begin{proof}
Denote by $\teeSGD^{(k)}$ the local mini-batch SGD estimator on machine $k$. Since $\xii_i$'s are \emph{i.i.d.} and independent to the initial estimator $\hat\tee_0$, by $N=nL$ and Theorem \ref{thm:sgd},
\begin{eqnarray}
\ep_0\|\teeDC-\tee^*\|_2^2&=&\ep_0\Big\|\frac1L\sum\limits_{k=1}^L(\teeSGD^{(k)}-\tee^*)\Big\|^2_2\cr
&\leq &\ep_0\Big\|\frac1L\sum\limits_{k=1}^L\Big{\{}(\teeSGD^{(k)}-\tee^*)-\ep_{0}(\teeSGD^{(k)}-\tee^*)\Big{\}}\Big{\|}_2^2\cr
& &+\big\|\ep_{0}(\teeSGD^{(1)}-\tee^*)\big\|_2^2\cr
&=&O\Big(\frac{p}{Lm^{1-\alpha}n^{\alpha}}\Big)+O\Big(\big(\dfrac{p}{m^{1-\alpha}n^{\alpha}}\big)^2\Big)\cr
&=&O\Big(\frac{p}{L^{1-\alpha}m^{1-\alpha}N^{\alpha}}+\frac{p^{2}}{L^{-2\alpha}m^{2-2\alpha}N^{2\alpha}}\Big).
\end{eqnarray}
\end{proof}

\subsection{Proof of the lower bound of bias for Example \ref{ex:log}}\label{sec:proof_dcsgd2}
We first provide an upper bound for $\ep_{0} \|\de_{i-1}\|_2^{3}$. On the event $\{\|\hat{\tee}_{0}-\tee^{*}\|_2\leq d_n\}$, 
\begin{eqnarray*}
	\ep_{0}\|\de_{i-1}\|^{3}_{2}&=&\ep_{0}\|\de_{i-1}\|^{3}_{2}I\{\cap_{j=1}^{i}\mathcal{C}_{j}\}+\ep_{0}\|\de_{i-1}\|^{3}_{2}I\{\{\cap_{j=1}^{i}\mathcal{C}_{j}\}^{c}\}\cr
	&\leq& \min(d_n^{3},d_n\ep_{0}\|\de_{i-1}\|^{2}_{2})+Cn^{3-\gamma}
\end{eqnarray*}
for any $\gamma>0$. Therefore $\max_{1\leq i\leq s}\ep_{0}\|\de_{i-1}\|^{3}_{2}=o(1)$ and
\begin{eqnarray}\label{ctt}
\ep_{0}\|\de_{i-1}\|^{3}_{2}=o(1)\ep_{0}\|\de_{i-1}\|^{2}_{2}+O(n^{3-\gamma}).
\end{eqnarray}

We next prove that $\ep_{0}\|\de_{i}\|^{2}_{2}\geq cr_{i}p/m$ for any $\tau>0$ and $i\geq p^{1/\alpha+\tau}$. Recall that
\begin{eqnarray}
\|\de_{i}\|_2^{2}&=&\|\de_{i-1}\|_2^{2}-2\frac{r_{i}}{m}\sum_{j\in H_{i}}\de'_{i-1}g(\z_{i-1},\xii_j)+\Big{\|}\frac{r_i}{m}\sum_{j\in H_{i}}g(\z_{i-1},\xii_j)\Big{\|}^{2}_2\cr
&=:&\|\de_{i-1}\|_2^{2}-2r_{i} U_{1}+r^{2}_{i} U_{2}.
\end{eqnarray}
Note that $\ep_0    \de'_{i-1}g(\z_{i-1},\xii_j)=\ep_{0}\de'_{i-1}G(\z_{i-1})$.
Hence from the proof of (\ref{acd}) and (\ref{ctt}),
\begin{eqnarray*}
	\ep_0 U_{1}=\frac{1}{m}\sum\limits_{j\in H_{i}}\ep_0    \de'_{i-1}g(\z_{i-1},\xii_j)\leq C\ep_{0}\|\de_{i-1}\|^{2}_{2}+Cn^{3-\gamma}
\end{eqnarray*}
for any sufficiently large $\gamma>0$. For $U_2$, 
\begin{eqnarray*}
	\ep_0\Big{\|}\frac{1}{m}\sum_{j\in H_{i}}g(\z_{i-1},\xii_j)\Big{\|}^{2}_2&\geq &\ep_0\Big{\|}\frac{1}{m}\sum_{j\in H_{i}}\bar{g}(\z_{i-1},\xii_j)\Big{\|}^{2}_2\cr
	&=& \frac{\sum_{j\in H_{i}}\Big{(}\ep_{0}\|g(\z_{i-1},\xii_j)\|^{2}_2-\ep_{0}\|G(\z_{i-1})\|^{2}_{2}\Big{)}}{m^{2}}
\end{eqnarray*}
Recall that $G(\tee)=\ep g(\tee,\xii)=\ep\left(\frac{\X}{1+e^{-\X'\tee^*}}-\frac{\X}{1+e^{-\X'\tee}}\right)$. We have 
\begin{eqnarray*}
	\|G(\tee)\|^{2}_{2}=\Big{\|}\ep\big(\frac{\X}{1+e^{-\X'\tee^*}}-\frac{\X}{1+e^{-\X'\tee}}\big) \Big{\|}^{2}_{2}\leq Cp\|\tee-\tee^{*}\|^{2}_{2}.
\end{eqnarray*}
So we have $\ep_{0}\|G(\z_{i-1})\|^{2}_{2}\leq Cp\ep_{0}\|\de_{i-1}\|^{2}_{2}=o(p)$. Also
\begin{eqnarray*}
	\ep_{0}\|g(\z_{i-1},\xii_j)\|^{2}_2&=&\ep_{0} \frac{\|\X_{j}\|^{2}_{2}}{(1+e^{-\X'_{j}\z_{i-1}})^{2}}\geq \ep \frac{\|\X_{j}\|^{2}_{2}}{(1+e^{-\X'_{j}\tee^{*}})^{2}}
	-Cp\ep_{0}\|\de_{i-1}\|_{2}\geq Cp.
\end{eqnarray*}
This yields that
\begin{eqnarray*}
	\ep_{0}\|\de_{i}\|^{2}_{2}\geq (1-cr_{i})\ep_{0}\|\de_{i-1}\|^{2}_{2}+Cr^{2}_{i}\frac{p}{m}
\end{eqnarray*}
for some positive constants $c$ and $C$. Then   by Lemma \ref{lem:an}, $\ep_{0}\|\de_{i}\|^{2}_{2}\geq c_1r_{i}p/m$ for all $i\geq p^{1/\alpha+\tau/2}$.

Now by $\tee^{*}=(1,0,...0)'$, $\ep X_{i}=0$ for $1\leq i\leq p-1$ and Taylor's formulation, we have
\begin{eqnarray*}
	\ep_{0}\delta_{i,1}&=&\ep_{0} \delta_{i-1,1}-\frac{r_{i}}{m}\sum_{j\in H_{i}}\ep_{0}g(\z_{i-1},\xii_{j})\cr
	&=&\ep_{0} \delta_{i-1,1}-r_{i}\frac{e}{(1+e)^{2}}\ep_{0}\delta_{i-1,1}+r_{i}\frac{e^{2}-e}{2(1+e)^{3}}\ep_{0}\de'_{i-1}\S\de_{i-1}+O(r_{i})\ep_{0}\|\de_{i-1}\|^{3}_{2}.
\end{eqnarray*}
By (\ref{ctt}), we have
\begin{eqnarray*}
	\ep_{0}\delta_{i,1}\geq (1-cr_{i})\ep_{0}\delta_{i-1,1}+Cr^{2}_{i}p/m
\end{eqnarray*}
for some positive $c$ and $C$ and all $i\geq p^{1/\alpha+\tau/2}$. Noting that 
$\prod_{j=p^{1/\alpha+\tau/2}+1}^{s}(1-cr_{j})=O(n^{-\gamma})$ for any $\gamma>0$ by letting $c_{0}$ in $r_{i}$ be sufficiently large,
by the proof of the second claim in Lemma  \ref{lem:an},
\begin{eqnarray*}
	\ep_{0}\delta_{s,1}&\geq& C\frac{p}{m}\sum\limits_{k=p^{1/\alpha+\tau/2}+1}^{s}r_{k}^2\prod_{j=k}^{i-1}(1-cr_{j+1})\cr
	& &+\ep_{0}\delta_{p^{1/\alpha+\tau/2},1}\prod_{j=p^{1/\alpha+\tau/2}+1}^{s}(1-cr_{j})\cr
	&\geq& Cr_{s}p/m,
\end{eqnarray*}
which completes the proof.
\qed

\subsection{Proof of the lower bound of bias for Example \ref{ex:qr}}\label{sec:proof_dcsgd3}

As above, we can show that $\max_{1\leq i\leq s}\ep_{0}\|\de_{i-1}\|^{3}_{2}=o(1)$ and $\ep_{0}\|\de_{i-1}\|^{3}_{2}=o(1)\ep_{0}\|\de_{i-1}\|^{2}_{2}+O(n^{3-\gamma})$.
Also, similarly,
\begin{eqnarray*}
	\ep_0 U_{1}\leq C\ep_{0}\|\de_{i-1}\|^{2}_{2}+Cn^{3-\gamma}
\end{eqnarray*}
for any sufficiently large $\gamma>0$.
Note that
\begin{eqnarray*}
	\ep_{0}\|G(\z_{i-1})\|^{2}_{2}\leq\ep_{0}\|\X_{j}\|^{2}_{2}\Big{(}F(\X'_{j}\de_{i-1})-\tau\Big{)}^{2}\leq C\ep_{0}\|\X_{j}\|^{2}_{2}(\X'_{j}\de_{i-1})^{2}
	\leq Cp\ep_{0}\|\de_{i-1}\|^{2}_{2}.
\end{eqnarray*}
Also
\begin{eqnarray*}
	\ep_{0}\|g(\z_{i-1},\xii_j)\|^{2}_2&=&\ep_{0}\|\X_{j}\|^{2}_{2}(F(\X'_{j}\de_{i-1})+\tau^{2}-2\tau F(\X'_{j}\de_{i-1}))\cr
	&\geq&\tau(1-\tau)\ep_{0}\|\X_{j}\|^{2}_{2}-Cp\ep_{0}\|\de_{i-1}\|_{2}\cr
	&\geq&Cp.
\end{eqnarray*}
Then   by Lemma \ref{lem:an}, we  have $\ep_{0}\|\de_{i}\|^{2}_{2}\geq cr_{i}p/m$ for all $i\geq p^{1/\alpha+\tau/2}$.

Since $\ep X_{i}=0$ for $1\leq i\leq p-1$, we have for  $i\geq p^{1/\alpha+\tau/2}$,
\begin{eqnarray*}
	\ep_{0}\delta_{i,1}&=&\ep_{0} \delta_{i-1,1}-\frac{r_{i}}{m}\sum_{j\in H_{i}}\ep_{0}g(\z_{i-1},\xii_{j})\cr
	&=&\ep_{0} \delta_{i-1,1}-\frac{r_{i}}{m}\sum_{j\in H_{i}}\ep_{0}[F(\X'_{j}\de_{i-1})-F(0)]\cr
	&=&(1-r_{i} F'(0))\ep_{0} \delta_{i-1,1}+r_{i}F^{''}(0)\ep_{0}\de'_{i-1}\S\de_{i-1}+O(r_{i}\ep_{0} \|\de_{i-1}\|^{3})\cr
	&\geq&(1-r_{i} F'(0))\ep_{0} \delta_{i-1,1}+cF^{''}(0)r^{2}_{i}p/m.
\end{eqnarray*}
So we have $\ep_{0}\delta_{s,1}\geq Cr_{s}p/m$.
\qed

\section{Proofs for Results of FONE in Section \ref{sec:FONE}}
\label{sec:proof_fone}
\subsection{Proof of Proposition \ref{prop:fone}}\label{suppsec:d1}
\begin{customprop}{\ref{prop:fone}}[On $\teeFONE$ for $\S^{-1} \a$ for smooth loss function $f$]
	Assume (C1$^{*}$), (C2) and (C3) hold. Suppose that the initial estimator satisfies $\|\hat{\tee}_{0}-\tee^{*}\|_{2}=O_{\pr}(d_n)$, and $\|\a\|_2=O(\tau_{n})$ (or $O_{\pr}(\tau_{n})$ for the random case). The iteration number $T$ and step-size $\eta_{n}$ satisfy $\log n=o(\eta_{n}T)$ and $T=O(n^{A})$ for some $A>0$. We have
	\begin{equation}
	\|\teeFONE-\S^{-1}\a\|_2=O_{\pr}\big{(}\tau_{n}d_n+\tau^{2}_{n}+\sqrt{\frac{p\log n}{n}}\tau_{n}+\sqrt{\eta_{n}}\tau_{n}+n^{-\gamma}\big{)}
	\end{equation}
	for any large $\gamma>0$. 
\end{customprop}
\begin{proof}
Define
\begin{eqnarray*}
	\mathcal{E}_{t}=\Big{\{}\sup_{
		\mbox{\tiny$
			\begin{array}{c}
			\|\tee_{1}-\tee^*\|_2\leq c_{4},\\
			\|\tee_{2}-\tee^*\|_2\leq c_{4}
			\end{array}$
	}}\frac{\Big{\|}\frac{1}{m}\sum_{i\in B_{t}}[\bar{g}(\tee_{1},\xii_i)-\bar{g}(\tee_{2},\xii_i)]\Big{\|}_2}{\sqrt{\|\tee_{1}-\tee_{2}\|_2^{2}+n^{-\gamma_{2}}}}\leq c\sqrt{\frac{p\log n}{m}}\Big{\}}.
\end{eqnarray*}
In Lemma \ref{le1} (see Section \ref{sec:tech_lemma} below), take $\u=(\tee'_{1},\tee'_{2})'$, $\u_{0}=(\tee'_{0},\tee'_{0})'$, $q=2p$ and $h(\u,\xii)=\bar{g}(\tee_{1},\xii)-\bar{g}(\tee_{2},\xii)$.
Then (C3)  implies that  (B1)--(B4) hold with $\alpha=1$, $b(\u)=C\|\tee_{1}-\tee_{2}\|_2^{2}$ and $b(\u)$ satisfies $|b(\u_{1})-b(\u_{2})|\leq C(1+\|\tee^*\|_2)\|\u_{1}-\u_{2}\|_2\leq
C\sqrt{p}\|\u_{1}-\u_{2}\|_2$. Therefore, by Lemma \ref{le1},
$$\pr(\mathcal{E}_{t})\geq 1-O(n^{-\gamma})$$ for any large $\gamma$. Now take $m=n$ and define
\begin{eqnarray*}
	\mathcal{E}=\Big{\{}\sup_{
		\mbox{\tiny$
			\begin{array}{c}
			\|\tee_{1}-\tee^*\|_2\leq c_{4},\\
			\|\tee_{2}-\tee^*\|_2\leq c_{4}
			\end{array}$
	}}\frac{\Big{\|}\frac{1}{n}\sum_{i=1}^{n}[\bar{g}(\tee_{1},\xii_i)-\bar{g}(\tee_{2},\xii_i)]\Big{\|}_2}{\sqrt{\|\tee_{1}-\tee_{2}\|_2^{2}+n^{-\gamma_{2}}}}\leq c\sqrt{\frac{p\log n}{n}}\Big{\}}.
\end{eqnarray*}
Then for any $\gamma_{2},\gamma>0$, $$\pr(\mathcal{E})\geq 1-O(n^{-\gamma}).$$

Recall
\begin{eqnarray*}
	\z_{t}=\z_{t-1}-\eta_{t}\Big{(}\frac{1}{m}\sum_{i\in B_{t}}[g(\z_{t-1},\xii_i)-g(\hat\tee_0,\xii_i)]+\a\Big{)}.
\end{eqnarray*}
Let the event $\mathcal{A}=\{\|\hat\tee_0-\tee^*\|_2\leq d_n,\|\a\|_2\leq \tau_{n}\}$ with $d_n,\tau_{n}\rightarrow 0$, and $\mathcal{B}_{t}=\{\|\z_{t-1}-(\hat\tee_0-\S^{-1}\a)\|_2\leq b_{n}\}$ with $b_{n}\rightarrow 0$, $\frac{p\log n}{m}\leq b_n^2$ and $\tau_{n}=o(b_{n})$. Note that on $\mathcal{A}\cap\mathcal{B}_{t}$, we have $\|\z_{t-1}-\tee^*\|_2\leq C(b_{n}+d_n)$. Define
$\mathcal{D}_{t}=\mathcal{A}\cap\cap_{i=1}^{t}\mathcal{C}_{i}$, where $\mathcal{C}_{i}$ is defined in the proof of Theorem \ref{thm:sgd}.

We first prove that on $\mathcal{D}_{t}$, $\max_{1\leq i\leq t}\|\z_{i}-(\hat\tee_0-\S^{-1}\a)\|_2\leq b_{n}.$
Let  $\tilde{\de}_{t}=\z_{t}-(\hat\tee_0-\S^{-1}\a)$ and
\begin{eqnarray*}
	\Delta(\z_{t-1})=\frac{1}{m}\sum_{i\in B_{t}}[g(\z_{t-1},\xii_i)-g(\hat\tee_0,\xii_i)]-[G(\z_{t-1})-G(\hat\tee_0)].
\end{eqnarray*}
We have
\begin{eqnarray*}
	\tilde{\de}_{t}=\tilde{\de}_{t-1}-\eta_{t}\Big{(}G(\z_{t-1})-G(\hat\tee_0)+\Delta(\z_{t-1})+\a\Big{)}
\end{eqnarray*}
and
\begin{eqnarray}\label{fone-1}
\|\tilde{\de}_{t}\|_2^{2}&=&\|\tilde{\de}_{t-1}\|_2^{2}-2\eta_{t}\tilde{\de}'_{t-1}[G(\z_{t-1})-G(\hat\tee_0)]-2\eta_{t}\tilde{\de}'_{t-1}(\Delta(\z_{t-1})+\a)\cr
& &+\eta^{2}_{t}\Big{\|}G(\z_{t-1})-G(\hat\tee_0)+\Delta(\z_{t-1})+\a\Big{\|}_2^{2}.
\end{eqnarray}
Note that $G(\z_{t-1})-G(\hat\tee_0)=\int_{0}^{1}\S(\hat\tee_{0}+t(\z_{t-1}-\hat\tee_{0}))dt(\z_{t-1}-\hat\tee_0)=:\hat{\S}(\z_{t-1})(\z_{t-1}-\hat\tee_0)$, where $\hat{\S}(\z_{t-1})$ satisfies $\|\hat{\S}(\z_{t-1})-\S(\tee^*)\|\leq \|\hat\tee_0-\tee^*\|_2+\|\z_{t-1}-\hat\tee_0\|_2$. So we have
\begin{eqnarray}\label{fon-2}
&&\tilde{\de}'_{t-1}[G(\z_{t-1})-G(\hat\tee_0)]+\tilde{\de}'_{t-1}(\Delta(\z_{t-1})+\a)\cr
&&=\tilde{\de}'_{t-1}\hat{\S}(\z_{t-1})(\z_{t-1}-\hat\tee_0)+\tilde{\de}'_{t-1}\a+\tilde{\de}'_{t-1}\Delta(\z_{t-1})\cr
&&=\tilde{\de}'_{t-1}\hat{\S}(\z_{t-1})\tilde{\de}_{t-1}-\tilde{\de}'_{t-1}[\hat{\S}(\z_{t-1})\S^{-1}-\I]\a+\tilde{\de}'_{t-1}\Delta(\z_{t-1}).
\end{eqnarray}
On $\mathcal{A}\cap\mathcal{B}_{t}$, by (C2), we have $2c_{1}^{-1}\geq\lambda_{\max}(\hat{\S}(\z_{t-1}))\geq\lambda_{\min}(\hat{\S}(\z_{t-1}))\geq c_{1}/2$ since $d_n,b_{n}\rightarrow 0$. Also
\begin{eqnarray}\label{fone-3}
&&\big{\|}\tilde{\de}'_{t-1}[\hat{\S}(\z_{t-1})\S^{-1}-\I]\a\big{\|}_2\cr
&&\quad\leq C_1\|\tilde{\de}_{t-1}\|_2\|\z_{t-1}-\tee^*\|_2\|\S^{-1}\a\|_2\cr
&&\quad\leq  C_1\|\tilde{\de}_{t-1}\|_2\big(\|\hat\tee_0-\tee^*\|_2+\|\z_{t-1}-\hat\tee_0\|_2\big)\|\S^{-1}\a\|_2\cr
&&\quad=  C_1\|\tilde{\de}_{t-1}\|_2\big(\|\hat\tee_0-\tee^*\|_2+\|\tilde{\de}_{t-1}-\S^{-1}\a\|_2\big)\|\S^{-1}\a\|_2\cr
&&\quad\leq C\Big{(}\tau_{n}\|\tilde{\de}_{t-1}\|_2\|\hat\tee_0-\tee^*\|_2+\tau^{2}_{n}\|\tilde{\de}_{t-1}\|_2+\tau_{n}\|\tilde{\de}_{t-1}\|_2^{2}\Big{)}.
\end{eqnarray}
Furthermore, on $\mathcal{D}_{t}\cap\mathcal{B}_{t}$, we have  that
\begin{eqnarray}\label{fone-4}
\|\tilde{\de}'_{t-1}\Delta(\z_{t-1})\|_2\leq C\sqrt{\frac{p\log n}{m}}\|\tilde{\de}'_{t-1}\|_2
\end{eqnarray}
and
\begin{eqnarray}\label{fone-5}
&&\big{\|}G(\z_{t-1})-G(\hat\tee_0)+\Delta(\z_{t-1})+\a\big{\|}_2^{2}\cr
&&\quad\leq C\Big{(}\|\z_{t-1}-\hat\tee_0\|_2^2+\|\Delta(\z_{t-1})\|_2^2+\tau_n^2\Big{)}\cr
&&\quad\leq C(\frac{p\log n}{m}+\tau^{2}_{n})+C\|\tilde{\de}_{t-1}\|_2^{2}.
\end{eqnarray}
Since $\eta_{t}\leq c$ for some small enough $c>0$, by (\ref{fone-1})-(\ref{fone-5}), we have, on $\mathcal{D}_{t}\cap\mathcal{B}_{t}$,
\begin{eqnarray*}
	\|\tilde{\de}_{t}\|_2^{2}&\leq& \|\tilde{\de}_{t-1}\|_2^{2}-\eta_{t}\tilde{\de}'_{t-1}\hat{\S}(\z_{t-1})\tilde{\de}_{t-1}+C\eta_t^2\|\tilde\de_{t-1}\|_2^2+\eta_t\tau_{n}\|\tilde{\de}_{t-1}\|_2^{2}\cr
	&&+C\eta_t\Big{(}\tau_{n}\|\tilde{\de}_{t-1}\|_2\|\hat\tee_0-\tee^*\|_2+\tau^{2}_{n}\|\tilde{\de}_{t-1}\|_2+\sqrt{\frac{p\log n}{m}}\|\tilde{\de}'_{t-1}\|_2\Big{)}\cr
	&&+C_{3}\eta_{t}^2(\frac{p\log n}{m}+\tau^{2}_{n})\cr
	&\leq& \|\tilde{\de}_{t-1}\|_2^{2}-C_{1}\eta_{t}\|\tilde{\de}_{t-1}\|_2^{2}+C_{2}\eta_{t}(\tau^{2}_{n}\|\hat\tee_0-\tee^*\|_2^{2}+\tau^{4}_{n}+\frac{p\log n}{m})\cr
	& &+C_{3}\eta^{2}_{t}(\frac{p\log n}{m}+\tau^{2}_{n})\cr
	&\leq&b^{2}_{n}-C_{1}\eta_{t}b^{2}_{n}+C_{2}\eta_{t}(\tau^{2}_{n}\|\hat\tee_0-\tee^*\|_2^{2}+\tau^{4}_{n}+\frac{p\log n}{m})\cr
	& &+C_{3}\eta^{2}_{t}(\frac{p\log n}{m}+\tau^{2}_{n}).
\end{eqnarray*}
Note that
\begin{eqnarray*}
	\tau^{2}_{n}\|\hat\tee_0-\tee^*\|_2^{2}+\tau^{4}_{n}+\frac{p\log n}{m}+\eta_{t}(\frac{p\log n}{m}+\tau^{2}_{n})=o(b^{2}_{n}).
\end{eqnarray*}
So we have on $\mathcal{D}_{t}\cap\mathcal{B}_{t}$,
$\|\tilde{\de}_{t}\|_2^{2}\leq b^{2}_{n}$. Combining the above arguments,
\begin{eqnarray*}
	\{\max_{1\leq i\leq t}\|\tilde{\de}_{i}\|_2>b_{n}\}\cap\mathcal{D}_{t}&&= \{\max_{1\leq i\leq t}\|\tilde{\de}_{i}\|_2>b_n,\max_{1\leq i\leq t-1}\|\tilde{\de}_{i}\|\leq b_n\}\cap\mathcal{D}_{t}\cr
	& &\quad+ \{\max_{1\leq i\leq t}\|\tilde{\de}_{i}\|_2>b_n,\max_{1\leq i\leq t-1}\|\tilde{\de}_{i}\|_2> b_n\}\cap\mathcal{D}_{t}\cr
	&&\subset\{\max_{1\leq i\leq t-1}\|\tilde{\de}_{i}\|_2> b_n\}\cap\mathcal{D}_{t}\cr
	&&\subset\{\|\tilde{\de}_{0}\|_2> b_n\}\cap\mathcal{D}_{t}=\emptyset,
\end{eqnarray*}
where the last inequality follows from $\|\tilde{\de}_{0}\|\leq b_n$ due to $\tau_{n}=o(b_n)$.
This proves that $\max_{1\leq i\leq t}\|\z_{i}-(\hat\tee_0-\tau_{n}\S^{-1}\a)\|_2\leq b_n$ on $\mathcal{D}_{t}$, i.e., $\mathcal{D}_{t}\subset\cap_{i=1}^{t+1}\mathcal{B}_{i}.$

Now let $\ep_{*}(\cdot)$ be the expectation to the random set $\{B_{t},t\geq 1\}$ given $\{\xii_1,\xii_2,\dots,\xii_n\}$.  Let
\begin{eqnarray*}
	\Delta_{n}(\z_{t-1})=\frac{1}{n}\sum_{i=1}^{n}[g(\z_{t-1},\xii_{i})-g(\hat\tee_0,\xii_{i})]-[G(\z_{t-1})-G(\hat\tee_0)].
\end{eqnarray*}
Let $\tilde{\mathcal{D}}_{t}=\mathcal{D}_{t}\cap\mathcal{E}\cap\mathcal{C}$.
As in each iteration, $B_{t}$, $1\leq t\leq T$ are independent, we have
\begin{eqnarray*}
	\ep_{*}\Big{[}\tilde{\de}'_{t-1}\Delta(\z_{t-1})I\{\tilde{\mathcal{D}}_{t-1}\}\Big{]}&=&\ep_{*}\Big{[}\ep_{*}\big[\tilde{\de}'_{t-1}\Delta(\z_{t-1})I\{\tilde{\mathcal{D}}_{t-1}\}|\{B_{i},1\leq i\leq t-1\}\big]\Big{]}\cr
	&=&\ep_{*}\Big{[}\tilde{\de}'_{t-1}\Delta_{n}(\z_{t-1})I\{\tilde{\mathcal{D}}_{t-1}\}\Big{]}.
\end{eqnarray*}
Note that $I\{\tilde{\mathcal{D}}_{t}\}=I\{\tilde{\mathcal{D}}_{t-1}\}-I\{\tilde{\mathcal{D}}_{t-1}\cap\mathcal{C}^{c}_{t}\}$.
Thus
\begin{eqnarray}\label{fone-6}
&&\ep_{*}\Big{[}\tilde{\de}'_{t-1}\Delta(\z_{t-1})I\{\tilde{\mathcal{D}}_{t}\}\Big{]}\cr
&&\quad=\ep_{*}\Big{[}\tilde{\de}'_{t-1}\Delta_{n}(\z_{t-1})I\{\tilde{\mathcal{D}}_{t-1}\}\Big{]}-
\ep_{*}\Big{[}\tilde{\de}'_{t-1}\Delta(\z_{t-1})I\{\tilde{\mathcal{D}}_{t-1}\cap\mathcal{C}^{c}_{t}\}\Big{]}.
\end{eqnarray}
By (C3), we can get
\begin{eqnarray*}
	\ep \sup_{\|\tee-\tee^*\|_2\leq c_4}\|g(\tee,\xii)\|_2^{2}\leq n^{c_5}
\end{eqnarray*}
for some $c_4,c_5>0$.  Note that on $\mathcal{D}_{t-1}$, we have $\|\tilde{\de}_{t-1}\|\leq b_n$ and
\begin{eqnarray*}
	\|\Delta(\z_{t-1})\|_2\leq 2\frac{1}{m}\sum_{i\in B_{t}}\sup_{\|\tee-\tee^*\|_2\leq C(b_{n}+d_n)}\|g(\tee,\xii_i)-G(\tee)\|_2.
\end{eqnarray*}
Hence
\begin{eqnarray*}
	\ep\Big{|}\ep_{*}\Big{[}\tilde{\de}'_{t-1}\Delta(\z_{t-1})I\{\tilde{\mathcal{D}}_{t-1}\cap\mathcal{C}^{c}_{t}\}\Big{]}\Big{|}=O(n^{c_5/2-\gamma})
\end{eqnarray*}
and
\begin{eqnarray*}
	\ep\Big{|}\ep_{*}\big{[}\big{\|}G(\z_{t-1})-G(\hat\tee_0)+\Delta(\z_{t-1})+\a\big{\|}^{2}_2I\{\tilde{\mathcal{D}}_{t-1}\cap\mathcal{E}^{c}_{t}\}\big{]}\Big{|}
	=O(n^{-\gamma})
\end{eqnarray*}
for any large $\gamma>0$ (by choosing $c$ in $\mathcal{E}_{t}$ sufficiently large). On $\mathcal{D}_{t-1}\cap\mathcal{E}$,
\begin{eqnarray}\label{fone-7}
\|\tilde{\de}'_{t-1}\Delta_{n}(\z_{t-1})\|_2&\leq& C\sqrt{\frac{p\log n}{n}}\big\|\z_{t-1}-\hat\tee_0\big\|_2\|\tilde{\de}_{t-1}\|_2+Cn^{-\gamma_{2}/2}\cr
&\leq& C\sqrt{\frac{p\log n}{n}}\|\tilde{\de}_{t-1}\|_2^{2}+C\tau_{n}\sqrt{\frac{p\log n}{n}}\|\tilde{\de}_{t-1}\|_2+Cn^{-\gamma_{2}/2}.
\end{eqnarray}
Similarly as above,  on $\mathcal{D}_{t}\cap\mathcal{E}_{t}$, we have
\begin{eqnarray}\label{fone-8}
&&\Big{\|}G(\z_{t-1})-G(\hat\tee_0)+\Delta(\z_{t-1})+\a\Big{\|}_2^{2}\cr
&&\leq C(\frac{p\log n}{m}\|\z_{t-1}-\hat\tee_0\|_2^{2}+\tau^{2}_{n})+C\|\z_{t-1}-\hat\tee_0\|_2^{2}+Cn^{-\gamma_{2}}\cr
&&\leq C\|\tilde{\de}_{t-1}\|_2^{2}+Cn^{-\gamma_{2}}+C\tau^{2}_{n}.
\end{eqnarray}
By (\ref{fone-1})-(\ref{fone-3}) and (\ref{fone-6})-(\ref{fone-8}),
\begin{eqnarray*}
	\ep[\|\tilde{\de}_{t}\|_2^{2}I\{\tilde{\mathcal{D}}_{t}\}]&\leq& (1-C_{1}\eta_{n})\ep[\|\tilde{\de}_{t-1}\|_2^{2}I\{\tilde{\mathcal{D}}_{t-1}\}]\cr
	& &+C_{2}\eta_{n}(\tau^{2}_{n}d_n^{2}+\tau^{4}_{n}+\frac{p\log n}{n}\tau^{2}_{n}+n^{-\gamma_{2}/2})\cr
	& &+C\eta^{2}_{n}(n^{-\gamma_{2}}+\tau^{2}_{n}),
\end{eqnarray*}
where we used $I\{\tilde{\mathcal{D}}_{t}\}\leq I\{\tilde{\mathcal{D}}_{t-1}\}$. This implies that
\begin{eqnarray*}
	\ep\big[\|\tilde{\de}_{t}\|_2^{2}I\{\tilde{\mathcal{D}}_{t}\}\big]&\leq &(1-C_1\eta_{n})^t\ep\big[\|\tilde\de_0\|_2^2I\{\mathcal{A}\cap\mathcal{E}\cap\mathcal{C}\}\big]\cr
	&&+\frac{1-(1-C_1\eta_{n})^t}{C_1\eta_{n}}\big[C\eta_{n}(\tau^{2}_{n}d_n^{2}+\tau^{4}_{n}+\frac{p\log n}{n}\tau^{2}_{n}+n^{-\gamma_{2}/2})\cr
	&&+C\eta^2_{n}(n^{-\gamma_{2}}+\tau^{2}_{n})\big].
\end{eqnarray*}
Note that $(1-C_1\eta_n)^t\leq \exp(-C_1\eta_nt)$. Then as long as $\log(n)=o(\eta_nt)$, 
\begin{eqnarray*}
	\ep[\|\tilde{\de}_{t}\|_2^{2}I\{\tilde{\mathcal{D}}_{t}\}]\leq C(\tau^{2}_{n}d_n^{2}+\tau^{4}_{n}+\frac{p\log n}{n}\tau^{2}_{n}+n^{-\gamma_{2}/2})+C\eta_{n}(n^{-\gamma_{2}}+\tau^{2}_{n}).
\end{eqnarray*}
Therefore, since $T=O(n^{A})$ for some $A>0$, we have $\pr(\{\mathcal{E}\cap\cap_{i=1}^{T}\mathcal{E}_{i}\}^c)=O(n^{-\gamma})$ and $\pr(\{\mathcal{C}\cap\cap_{i=1}^{T}\mathcal{C}_{i}\}^c)=O(n^{-\gamma})$ for any $\gamma>0$.  That is, when $\pr(\mathcal{A}^{c})=o(1)$, we have
\begin{eqnarray*}
	\|\tilde{\de}_{T}\|_2=O_{\pr}(\tau_{n}d_n+\tau^{2}_{n}+\sqrt{\frac{p\log n}{n}}\tau_{n}+\sqrt{\eta_n}\tau_{n}+n^{-\gamma_{2}/4}).
\end{eqnarray*}
This proves the theorem.
\end{proof}

\subsection{Proof of Proposition \ref{prop:fone_non}}\label{suppsec:d2}
\begin{customprop}{\ref{prop:fone_non}}[On $\teeFONE$ for $\S^{-1} \a$ for non-smooth loss function $f$]
	Assume the conditions in Proposition \ref{prop:fone} hold with (C3) being replaced by (C3$^{*}$).  We have
	{\small\begin{align}
		\|\teeFONE&-\S^{-1}\a\|_2\nonumber \\
		& = O_{\pr}\Big{(}\tau_{n}d_n+\tau^{2}_{n}+\sqrt{\frac{p\log n}{n}}\sqrt{\tau_{n}}+\frac{p\log n}{m}\sqrt{\eta_{n}}+\sqrt{\eta_{n}}\tau_{n}+\frac{p\log n}{n}\Big{)}. 
		\end{align}}
\end{customprop}
\begin{proof}
Note that (C3$^{*}$)  implies that (B4$^{*}$) holds with $\alpha=0$. Define
\begin{eqnarray*}
	\mathcal{E}_{t}=\Big{\{}\sup_{
		\mbox{\tiny$
			\begin{array}{c}
			\|\tee_{1}-\tee^*\|_2\leq c_4,\\
			\|\tee_{2}-\tee^*\|_2\leq c_4
			\end{array}$
	}}\frac{\Big{\|}\frac{1}{m}\sum_{i\in B_{t}}[\bar{g}(\tee_{1},\xii_i)-\bar{g}(\tee_{2},\xii_i)]\Big{\|}_2}{\sqrt{\|\tee_{1}-\tee_{2}\|_2+\frac{p\log n}{m}}}\leq c\sqrt{\frac{p\log n}{m}}\Big{\}}
\end{eqnarray*}
and
\begin{eqnarray*}
	\mathcal{E}=\Big{\{}\sup_{
		\mbox{\tiny$
			\begin{array}{c}
			\|\tee_{1}-\tee^*\|_2\leq c_4,\\
			\|\tee_{2}-\tee^*\|_2\leq c_4
			\end{array}$
	}}\frac{\Big{\|}\frac{1}{n}\sum_{i=1}^{n}[\bar{g}(\tee_{1},\xii_i)-\bar{g}(\tee_{2},\xii_i)]\Big{\|}_2}{\sqrt{\|\tee_{1}-\tee_{2}\|_2+\frac{p\log n}{n}}}\leq c\sqrt{\frac{p\log n}{n}}\Big{\}}.
\end{eqnarray*}
We have by Lemma \ref{le1} that $\pr(\mathcal{E}\cap\cap_{t=1}^{T}\mathcal{E}_{t})\geq 1-O(n^{-\gamma})$ for any $\gamma>0$.

On $\mathcal{D}_{t-1}\cap\mathcal{E}$,
\begin{eqnarray*}
	\|\tilde{\de}'_{t-1}\Delta_{n}(\z_{t-1})\|&\leq& C\sqrt{\frac{p\log n}{n}}\|\z_{t-1}-\hat\tee_0\|_2^{1/2}\|\tilde{\de}_{t-1}\|_2+C\frac{p\log n}{n}\|\tilde{\de}_{t-1}\|_2+
	Cn^{-\gamma_{2}/2}\cr
	&\leq& C\sqrt{\frac{p\log n}{n}}\|\tilde{\de}_{t-1}\|_2^{3/2}+C(\tau^{1/2}_{n}\sqrt{\frac{p\log n}{n}}+\frac{p\log n}{n})\|\tilde{\de}_{t-1}\|_2+Cn^{-\gamma_{2}/2}.
\end{eqnarray*}
Similarly, on $\mathcal{D}_{t-1}\cap\mathcal{E}_{t}$,
we have
\begin{eqnarray*}
	&&\Big{\|}G(\z_{t-1})-G(\hat\tee_0)+\Delta(\z_{t-1})+\a\Big{\|}_2^{2}\cr
	&&\leq C\Big(\frac{p\log n}{m}\|\z_{t-1}-\hat\tee_0\|_2+\big(\frac{p\log n}{m}\big)^{2}+\tau^{2}_{n}\Big)+C\|\z_{t-1}-\hat{\tee}_{0}\|_2^{2}+Cn^{-\gamma_{2}}\cr
	&&\leq C\|\tilde{\de}_{t-1}\|_2^{2}+\big(\frac{p\log n}{m}\big)^{2}+C\tau^{2}_{n}.
\end{eqnarray*}
So as the proof of Proposition \ref{prop:fone}, we can get
\begin{eqnarray*}
	\|\tilde{\de}_{T}\|^{2}=O_{\pr}\Big{(} \tau_{n}\frac{p\log n}{n}+\big(\frac{p\log n}{n}\big)^{2}+\eta_{n}\big(\frac{p\log n}{m}\big)^{2}+\tau^{2}_{n}d_n^{2}+\eta_{n}\tau^{2}_{n}+\tau^{4}_{n}\Big{)}.
\end{eqnarray*}
The proof is complete.
\end{proof}

\subsection{Proof of Theorem \ref{thm:d-fone}}\label{suppsec:d3}
\begin{customthm}{\ref{thm:d-fone}} [distributed FONE for smooth loss function $f$] Assume (C1$^{*}$), (C2) and (C3) hold, $N=O(n^{A})$ for some $A>0$. Suppose that $\|\hat{\tee}-\tee^{*}\|_{2}+\|\hat\tee_0-\tee^*\|_2=O_{\pr}(n^{-\delta_1})$ for some $\delta_1>0$. Let $\eta_n=n^{-\delta_{2}}$ for some $\delta_{2}>0$,  $\log n=o(\eta_{n}T)$, $T=O(n^{A})$ for some $A>0$, and $p\log n=o(m)$. For any $\gamma>0$, there exists $K_{0}>0$ such that for any $K\geq K_{0}$, we have $\|\teedis-\hat{\tee}\|_2=O_{\pr}(n^{-\gamma})$. 
\end{customthm}
\begin{proof}
	Since $f(\tee,\xii)$ is  differentiable, we have
	$\frac{1}{N}\sum_{i=1}^{N}g(\hat{\tee},\xii_i)=0$. Denote by 
	\begin{eqnarray*}
		\mathcal{E}_N=\left\{\sup_{
			\mbox{\tiny$
				\begin{array}{c}
				\|\tee_{1}-\tee^*\|_2\leq c_4,\\
				\|\tee_{2}-\tee^*\|_2\leq c_4
				\end{array}$
		}}\frac{\Big{\|}\frac{1}{N}\sum_{i=1}^{N}[\bar{g}(\tee_{1},\xii_i)-\bar{g}(\tee_{2},\xii_i)]\Big{\|}_2}{\sqrt{\|\tee_{1}-\tee_{2}\|_2^{2}+N^{-\gamma_{2}}}}\leq c\sqrt{\frac{p\log N}{N}}\right\}.
	\end{eqnarray*}
	By the proof of Proposition \ref{prop:fone}, we have $\pr(\mathcal{E}_{N})\geq 1-O(N^{-\gamma})$ for any large $\gamma$. Therefore, 
	on the event $\mathcal{E}_{N}\cap \{\|\hat{\tee}_{j-1}-\tee^{*}\|_{2}+\|\hat{\tee}-\tee^{*}\|_{2}\leq c_{2}\}$,`
	in the $j$-th round in Algorithm \ref{algo:distributed_FONE},
	\begin{eqnarray*}
		\a&=&\frac{1}{N}\sum_{i=1}^{N}[g(\hat\tee_{j-1},\xii_i)-g(\hat{\tee},\xii_i)]\cr
		&=&G(\hat\tee_{j-1})-G(\hat{\tee})+O\Big{(}\sqrt{\frac{p\log N}{N}}\|\hat\tee_{j-1}-\hat{\tee}\|_2+N^{-\gamma_{2}/2}\Big{)}\cr
				&=&\S(\tee^*)(\hat\tee_{j-1}-\hat{\tee})+O\Big{(}\sqrt{\frac{p\log N}{N}}\|\hat\tee_{j-1}-\hat{\tee}\|_2+N^{-\gamma_{2}/2}\Big{)}\cr
		&&+O\big(\|\hat\tee_{j-1}-\hat{\tee}\|_2^{2}+
		\|\hat\tee_{j-1}-\hat{\tee}\|_2\|\hat\tee_{j-1}-\tee^{*}\|_2\big).
	\end{eqnarray*}
	In above and throughout the paper, for a sequence of vector $\{\x_n\}$,  we write $\x_n=O(a_n)$ if $\|\x_n\|_2=O(a_n)$ for simplicity. Now, in the first round of iteration, i.e., $j=1$, we have
	$\|\hat\tee_0-\hat{\tee}\|_2=O_{\pr}(n^{-\delta_{1}})$.   Then  we can let $\tau_{n}=Cn^{-\delta_{1}}$ with some large constant $C$. By Proposition \ref{prop:fone}, we have
	\begin{eqnarray*}
		\|\hat{\tee}_{1}-\hat{\tee}\|_2=O_{\pr}(n^{-2\delta_{1}}+\sqrt{\frac{p\log n}{n}}n^{-\delta_{1}}+n^{-\delta_{1}-\delta_{2}/2}+n^{-\gamma}).
	\end{eqnarray*}
	This yields that
	$\|\hat{\tee}_{1}-\hat{\tee}\|_2=O_{\pr}(n^{-\delta_{1}-r}+n^{-\gamma})$ with $r=\min(\delta_{1},\delta_{2}/2,(1-\kappa_{1})/2)$. Now in the second round of iteration, we let $d_n=n^{-\delta_{1}}$ and $\tau_{n}=C(n^{-\delta_{1}-r}+n^{-\gamma})$. Then $\|\hat{\tee}_{2}-\hat{\tee}\|_2=O_{\pr}(n^{-\delta_{1}-2r}+n^{-\gamma})$. Repeating this argument, we can show that $\|\hat{|\tee}_{K}-\hat{\tee}\|_2=O_{\pr}(n^{-\delta_{1}-Kr}+n^{-\gamma})$ which proves the theorem since  $\gamma$ can be arbitrarily large.
\end{proof}

\subsection{Proof of Thoerem \ref{thm:d-fone-non}}\label{suppsec:d4}

\begin{customthm}{\ref{thm:d-fone-non}}[distributed FONE for non-smooth loss function $f$] Suppose that (C1$^{*}$), (C2), (C3$^{*}$) and \eqref{c4} hold, $N=O(n^{A})$ and $T=O(n^{A})$ for some $A>0$.  Suppose that $\|\hat{\tee}-\tee^{*}\|_{2}+\|\hat\tee_0-\tee^*\|_2=O_{\pr}(n^{-\delta_1})$ for some $\delta_1>0$. Let $\eta_n=n^{-\delta_{2}}$ for some $\delta_{2}>0$,  $\log n=o(\eta_{n}T)$, and $p\log n=o(m)$. For any $0<\gamma<1$, there exists $K_{0}>0$ such that for any $K\geq K_{0}$, we have
	\begin{eqnarray}
	\|\teedis-\hat{\tee}\|_2=O_{\pr}\Big{(}\frac{q_{N}}{N}+\sqrt{\eta_{n}}\frac{p\log n}{m}+\big{(}\frac{p\log n}{n}\big{)}^{\gamma}\Big{)}.
	\end{eqnarray}
\end{customthm}

\begin{proof}
Note that  $\|\hat\tee_0-\tee^{*}\|_2+\|\hat\tee-\tee^{*}\|_2=O_{\pr}(n^{-\delta_{1}})$ and thus $d_n=O(n^{-\delta_{1}})$.  For any $0<\delta<1$, by Holder's inequality, we have
\begin{eqnarray*}
	\tau_{n}\frac{p\log n}{n}\leq \tau^{2+2\delta}_{n}+\Big{(}\frac{p\log n}{n}\Big{)}^{\frac{2+2\delta}{1+2\delta}}.
\end{eqnarray*}
This indicates that
\begin{eqnarray*}
	\|\tilde{\de}_{T}\|_{2}=O_{\pr}\Big{(} \tau^{1+\delta}_{n}+\big(\frac{p\log n}{n}\big)^{\frac{1+\delta}{1+2\delta}}+\sqrt{\eta_{n}}\frac{p\log n}{m}+\tau_{n}n^{-r_{1}}\Big{)}
\end{eqnarray*}
with $r_{1}=\min(\delta_{1},\delta_{2}/2)$. Now we estimate $\tau_{n}$. Let $\tau_{nj}$ be the value of $\tau_{n}$ in the $j$-th round.
We have
\begin{eqnarray*}
	\a&=&\frac{1}{N}\sum_{i=1}^{N}[g(\hat\tee_{j-1},\xii_i)-g(\hat{\tee},\xii_i)]+O_{\pr}\Big{(}\frac{q_{N}}{N}\Big{)}\cr
	&=&G(\hat\tee_{j-1})-G(\hat{\tee})+O_{\pr}(1)\Big{(}\sqrt{\frac{p\log N}{N}}\|\hat\tee_{j-1}-\hat{\tee}\|_2^{1/2}+\frac{q_{N}+p\log N}{N}\Big{)}\cr
	&=&\S(\tee^{*})(\hat\tee_{j-1}-\hat{\tee})+O_{\pr}(1)\Big{(}\sqrt{\frac{p\log N}{N}}\|\hat\tee_{j-1}-\hat{\tee}\|_2^{1/2}+\|\hat\tee_{j-1}-\hat{\tee}\|_2^{2}\cr
	& &+\|\hat\tee_{j-1}-\hat{\tee}\|_2\|\hat\tee-\tee^{*}\|_2 +\frac{q_{N}+p\log N}{N}\Big{)}\cr
	&=:&\S(\tee^{*})(\hat\tee_{j-1}-\hat{\tee})+A_{nj}.
\end{eqnarray*}
So on the event
$$E_{j-1}:=\Big{\{}\|\hat{\tee}_{j-1}-\hat{\tee}\|_{2}\leq Cn^{-b_{j-1}}+C\frac{q_{N}}{N}+C\Big{(}\frac{p\log n}{n}\Big{)}^{\frac{1+\delta}{1+2\delta}} +\sqrt{\eta_{n}}\frac{p\log n}{m}\Big{\}}$$ for some $b_{j-1}>0$, we have
\begin{eqnarray*}
	\tau_{nj}&\leq& Cn^{-b_{j-1}}+C\frac{q_{N}}{N}+C\Big{(}\frac{p\log N}{N}\Big{)}^{\frac{1+\delta}{1+2\delta}}+C\sqrt{\eta_{n}}\frac{p\log n}{m}
\end{eqnarray*}
and
\begin{eqnarray*}
	\|A_{nj}\|_{2}\leq Cn^{-b_{j-1}(1+\delta)}+Cn^{-b_{j-1}-r_{1}}+C\frac{q_{N}}{N}+C\Big{(}\frac{p\log N}{N}\Big{)}^{\frac{1+\delta}{1+2\delta}}+\sqrt{\eta_{n}}\frac{p\log n}{m}
\end{eqnarray*}
by noting that
\begin{eqnarray*}
	\sqrt{\frac{p\log N}{N}}\|\hat\tee_{j-1}-\hat{\tee}\|_2^{1/2}\leq \|\hat \tee_{j-1}-\hat{\tee}\|_2^{1+\delta}+\Big{(}\frac{p\log N}{N}\Big{)}^{\frac{1+\delta}{1+2\delta}}.
\end{eqnarray*}
Hence on the event $E_{j-1}$, we have
\begin{eqnarray*}
	\|\hat{\tee}_{j}-\hat{\tee}\|_2&\leq& \|\tilde{\de}_{T}\|_{2}+\|A_{nj}\|_{2}\cr
	&\leq&Cn^{-b_{j-1}(1+\delta)}+Cn^{-b_{j-1}-r_{1}/2}+C\frac{q_{N}}{N}+C\Big{(}\frac{p\log n}{n}\Big{)}^{\frac{1+\delta}{1+2\delta}}+C\sqrt{\eta_{n}}\frac{p\log n}{m}.
\end{eqnarray*}
Note that we can let $b_{0}=\delta_{1}$. Then it is easy to see that we can let $b_{j}\geq \delta_{1}$ for all $j$. 
Hence $b_{j}$ satisfies $b_{j}\geq b_{j-1}+\min(\delta\delta_{1}, r_{1}/2)$.  This proves that
\begin{eqnarray*}
	\|\teedis-\hat{\tee}\|_2=O_{\pr}\Big{(}n^{-\delta_{1}-K\min(\delta\delta_{1}, r_{1}/2)}+\frac{q_{N}}{N}+\Big{(}\frac{p\log n}{n}\Big{)}^{\frac{1+\delta}{1+2\delta}}+C\sqrt{\eta_{n}}\frac{p\log n}{m}\Big{)}.
\end{eqnarray*}
The proof is complete.
\end{proof}


\section{Proofs for Results of Inference in Section \ref{sec:unitw}}\label{sec:supp_inf}

\begin{customthm}{\ref{thm:sinvw}}[Estimating $\S^{-1}\w$ for a smooth loss function $f$]  Under the conditions of Proposition \ref{prop:fone}, let $\tau_{n}=\sqrt{(p\log n)/n}$. Assuming that $\|\hat\tee_0-\tee^*\|_2=O_{\pr}(d_n)$ and $\log n=o(\eta_{n}T)$, we have
	\begin{eqnarray}
	\|\widehat{\S^{-1}\w}-\S^{-1}\w\|_2=O_{\pr}\Big{(}\sqrt{\frac{p\log n}{n}}+\sqrt{\eta_n}+d_n\Big{)}.
	\end{eqnarray}
\end{customthm}

\begin{proof}	By Proposition \ref{prop:fone}, we have $\hat{{\S}^{-1}\w}=(\hat\tee_0-\z_T)/\tau_n$ and 
	\[
	\|\hat\tee_0-\z_T-{\S}^{-1}\tau_n\w\|_2=O_{\pr}\Big{(}\tau_{n}d_n+\tau^{2}_{n}+\sqrt{\frac{p\log n}{n}}\tau_{n}+\sqrt{\eta_{n}}\tau_{n}+n^{-\gamma}\Big{)}.
	\]
	Therefore, when $\tau_n=\sqrt{(p\log n)/n}$,  we have
	\[
	\big\|\hat{{\S}^{-1}\w}-{\S}^{-1}\w\big\|_2=O_{\pr}\Big{(}\sqrt{\frac{p\log n}{n}}+\sqrt{\eta_n}+d_n\Big{)}.
	\]	
\end{proof}

\begin{customthm}{\ref{thm:sinvw-non}}[Estimating $\S^{-1}\w$ for non-smooth loss function $f$] Under the conditions of Proposition \ref{prop:fone_non}, let $\tau_n=\big((p\log n)/n\big)^{1/3}$. Assuming that $\|\hat\tee_0-\tee^*\|_2=O_{\pr}(d_n)$ and $\log n=o(\eta_{n}T)$, we have 
	\begin{equation}
	\|\widehat{\S^{-1}\w}-\S^{-1}\w\|_2=O_{\pr}\Big(\big(\frac{p\log n}{n}\big)^{1/3}+\sqrt{\eta_n}\big(\frac{n^{1/3}(p\log n)^{2/3}}{m}+1\big)+d_n\Big{)}.
	\end{equation}
\end{customthm}

%
\begin{proof} By Proposition \ref{prop:fone_non}, we have $\hat{{\S}^{-1}\w}=(\hat\tee_0-\z_T)/\tau_n$ and 
	\[
	\|\hat\tee_0-\z_T-{\S}^{-1}\tau_n\w\|_2=O_{\pr}\Big{(}\tau_{n}d_n+\tau^{2}_{n}+\sqrt{\frac{p\log n}{n}}\sqrt{\tau_{n}}+\frac{p\log n}{m}\sqrt{\eta_{n}}+\sqrt{\eta_{n}}\tau_{n}+\frac{p\log n}{n}\Big{)}.
	\]
	Therefore, when $\tau_n=\big({(p\log n)/n}\big)^{1/3}$,  we have
	\begin{eqnarray*}
		\big\|\hat{{\S}^{-1}\w}-{\S}^{-1}\w\big\|_2&=&O_{\pr}\Big{(}\sqrt{\frac{p\log n}{n\tau_{n}}}+\tau_{n}+\big(\frac{p\log n}{\tau_{n}m}+1\big)\sqrt{\eta_n}+d_n\Big{)}\cr
		&=&O_{\pr}\Big(\big(\frac{p\log n}{n}\big)^{1/3}+\sqrt{\eta_n}\big(\frac{n^{1/3}(p\log n)^{2/3}}{m}+1\big)+d_n\Big{)}.
	\end{eqnarray*}
\end{proof}

\section{Technical Lemmas}
\label{sec:tech_lemma}
In this section, we will provide  proofs of the technical lemmas used above. 
\begin{lemma}\label{lem:bn}  Let $\zet_{1},...,\zet_{n}$ be independent $p$-dimensional
	random vectors with $\ep \zet_{i}=\boldsymbol{0}$ and \\ $\sup_{\|\v\|_2=1}\ep (\v'\zet_{i})^{2}\exp(t_{0}|\v'\zet_{i}|)<\infty$ for some $t_0>0$.
	Let  $B_{n}$ be a sequence of positive numbers such that
	\begin{eqnarray*}
		\sup_{\|\v\|_2=1}\sum_{i=1}^{n}\ep (\v'\zet_{i})^{2}\exp(t_{0}|\v'\zet_{i}|)\leq B^{2}_{n}.
	\end{eqnarray*}
	Then for $x>0$ and $2\sqrt{p+x^{2}}\leq B_{n}$, we have
	\begin{eqnarray*}
		\pr\Big{(}\Big{\|}\sum_{i=1}^{n}\zet_{i}\Big{\|}_2\geq C_{t_{0}}B_{n}\sqrt{p+x^{2}}\Big{)}\leq e^{-x^{2}},
	\end{eqnarray*}
	where $C_{t_{0}}$ is a positive constant depending only on $t_{0}$.
\end{lemma}

\noindent{\bf Proof.}
Let $S^{p-1}_{1/2}$ be a $1/2$ net of the unit sphere $S^{p-1}$ in the Euclidean distance in $\mathbb{R}^{p}$. By the proof of Lemma 3 in \cite{cai2010optimal}, we have $d_{p}:=$Card$(S^{p-1}_{1/2})\leq 5^{p}$. So there exist $d_{p}$ points  $\v_{1},...,\v_{d_{p}}$ in $S^{p-1}$  such that for any $\v$ in $S^{p-1}$, we have $\|\v-\v_{j}\|_2\leq 1/2$ for some $j$. Therefore, for any vector $\u\in \R^{p}$, $\|\u\|_2\leq \sup_{j\leq d_{p}}|\v_{j}'\u|+\|\u\|_2/2$.
That is,  $\|\u\|_2\leq 2\sup_{j\leq d_{p}}|\v_{j}'\u|$.  Therefore,
\begin{eqnarray*}
	\pr\Big{(}\Big{\|}\sum_{i=1}^{n}\zet_{i}\Big{\|}_2\geq C_{t_{0}}B_{n}\sqrt{p+x^{2}}\Big{)}&\leq& \pr\Big{(}\sup_{j\leq d_{p}}\Big|\sum_{i=1}^{n}\v'_{j}\zet_{i}\Big|\geq 2^{-1}C_{t_{0}}B_{n}\sqrt{p+x^{2}}\Big{)} \cr
	&\leq& 5^{p}\max_{j}\pr\Big{(}\Big{|}\sum_{i=1}^{n}\v'_{j}\zet_{i}\Big{|}\geq 2^{-1}C_{t_{0}}B_{n}\sqrt{p+x^{2}}\Big{)}\cr
	&\leq& 
	e^{-x^2},
\end{eqnarray*}
where we let $C_{t_{0}}=4(t_{0}+t_{0}^{-1})$. The last inequality follows from Lemma 1 in \cite{cailiu2011}, by noting that $2\sqrt{p+x^{2}}\leq B_{n}$.\qed

Let $h(\u,\xii)$ be a $q$-dimensional  random vector with zero mean.   For some constant $c_4>0$,
define 
\begin{equation}\label{eq:Theta0}
\Theta_{0}=\{\u\in \R^{q}: \|\u-\u_{0}\|_2\leq c_4\},
\end{equation}
where $\u_{0}$ is a point in $\R^{q}$.
Assume the following conditions hold.

{\bf (B1).} $\ep \sup_{\u\in\Theta_{0}}\|h(\u,\xii)\|_2\leq q^{c}$  for some $c>0$. \vspace{3mm}

{\bf (B2).} For $\u\in\Theta_{0}$, assume $\sup_{\|\v\|_2=1}\ep(\v'h(\u,\xii))^{2}\leq b(\u)$ and $b(\u)$ satisfies
$|b(\u_{1})-b(\u_{2})|\leq q^{c}\|\u_{1}-\u_{2}\|_2^{\gamma}$ for some $c,\gamma>0$, uniformly in $\u_{1},\u_{2}\in \Theta_{0}$.
\vspace{3mm}

{\bf (B3).} Assume that for some $t_0>0$ and $0\leq \alpha\leq 1$, $$\sup_{\|\v\|_2=1}\ep (\v'h(\u,\xii))^{2}\exp\Big{(}t_{0}\Big{|}\frac{\v'h(\u,\xii)}{b^{\alpha/2}(\u)}\Big{|}\Big{)}\leq Cb(\u)$$ for some constant  $C>0$, uniformly in $\u\in\Theta_{0}$.\vspace{3mm}

{\bf (B4).}  $\ep \sup_{\u_{1},\u_{2}\in\Theta_{0},\|\u_{1}-\u_{2}\|_2\leq n^{-M}}\|h(\u_{1},\xii)-h(\u_{2},\xii)\|_2\leq q^{c_2}n^{-c_3M}$  for any $M\geq M_{0}$ with some $M_{0}>0$ and some $c_2,c_3>0$.\vspace{3mm}

{\bf (B4$^{*}$)} We have
\begin{eqnarray*}
	\sup_{\u_{1}\in\Theta_{0}}\ep \sup_{\u_{2}\in\Theta_{0}:\|\u_{1}-\u_{2}\|_2\leq n^{-M}}\Big{\|}\frac{h(\u_{1},\xii)-h(\u_{2},\xii)}{b^{\alpha/2}(\u_{2})}\Big{\|}^{4}_2\leq q^{c_2}n^{-c_3M}
\end{eqnarray*}
for some $c_2,c_3>0$,
and
\begin{eqnarray*}
	\sup_{\u_{1}\in\Theta_{0}}\sup_{\|\v\|_2=1}\ep \sup_{\u_{2}\in\Theta_{0}:\|\u_{1}-\u_{2}\|_2\leq n^{-M}}\exp\Big{(}t_{0}\Big{|}\frac{\v'[h(\u_{1},\xii)-h(\u_{2},\xii)]}{b^{\alpha/2}(\u_{2})}\Big{|}\Big{)}\leq C
\end{eqnarray*}
for any $M\geq M_{0}$ with some $M_{0}>0$  and some  $t_0,C>0$.

\begin{lemma}\label{le1} Let $1\leq m\leq n$ and $q\leq n$. Assume (B1)-(B3)  and (B4) (or (B4$^{*}$)) hold. For any $\gamma_{1},\gamma_{2}>0$, there exists a constant $c>0$ such that
	\begin{eqnarray*}
		\pr\Big{(}\sup_{\tee\in \Theta_{0}}\frac{\Big{\|}\frac{1}{m}\sum_{i\in B_{t}}h(\tee,\xii_i)\Big{\|}_2}{\sqrt{b(\tee)+b^{\alpha}(\tee)(q\log n)/m+n^{-\gamma_{2}}}} \geq c\sqrt{\frac{q\log n}{m}}\Big{)}=O(n^{-\gamma_{1} }).
	\end{eqnarray*}
\end{lemma}

\noindent{\bf Proof.}  Since $B_{t}$ and $\{\xii_{i}\}$ are independent, without loss of generality, we can assume that $B_{t}$ is a fixed set.
Let $\{\tee_{1}...,\tee_{s_{q}}\}$ be $s_{q}$ points such that for any $\tee\in\Theta_{0}$, we have $\|\tee-\tee_{j}\|_2\leq n^{-M}$ for sufficiently large $M$ and some $j$.  It is easy to prove that  $s_{q}\leq Cq^{q/2}n^{qM}\leq Cn^{2qM}$ for some $C>0$. For notation briefness,  let $\tilde{b}(\tee)=b(\tee)+b^{\alpha}(\tee)(q\log n)/m+n^{-\gamma_{2}}$. We have
\begin{eqnarray*}
	\frac{\sum_{i\in B_{t}}h(\tee,\xii_i)}{\sqrt{\tilde{b}(\tee)}}-\frac{\sum_{i\in B_{t}}h(\tee_{j},\xii_i)}{\sqrt{\tilde{b}(\tee_{j})}}
	&=&\sum_{i\in B_{t}}h(\tee,\xii_i)\times\frac{\sqrt{\tilde{b}(\tee_{j})}-\sqrt{\tilde{b}(\tee)}}{\sqrt{\tilde{b}(\tee)\tilde{b}(\tee_{j})}}\cr
	& &+\frac{1}{\sqrt{\tilde{b}(\tee_{j})}}\times\Big{(}\sum_{i\in B_{t}}h(\tee,\xii_i)-\sum_{i\in B_{t}}h(\tee_{j},\xii_i)\Big{)}\cr
	&=:&\Gamma_{1}+\Gamma_{2}.
\end{eqnarray*}
By (B1), we can obtain hat
\begin{eqnarray*}
	\ep \sup_{\tee\in\Theta_{0}}\Big\|\sum_{i\in B_{t}}h(\tee,\xii_i)\Big\|_2=O(n^{c})
\end{eqnarray*}
for some $c>0$.  By (B2), we can  show that
$|\tilde{b}(\tee)-\tilde{b}(\tee_{j})|\leq Cn^{c-\alpha'\gamma M}$ for $\|\tee-\tee_{j}\|_2\leq n^{-M}$, uniformly in $j$, where $\alpha'=1$ if $\alpha=0$ and $\alpha'=\alpha$ if $\alpha>0$.
Therefore
\begin{eqnarray*}
	\max_{j}\sup_{\|\tee-\tee_{j}\|_2\leq n^{-M}}\frac{\Big|\sqrt{\tilde{b}(\tee_{j})}-\sqrt{\tilde{b}(\tee)}\Big|}{\sqrt{\tilde{b}(\tee)\tilde{b}(\tee_{j})}}\leq Cn^{c+2\gamma_{2}-\gamma\alpha' M}.
\end{eqnarray*}
This implies that $\ep \max_{j}\sup_{\|\tee-\tee_{j}\|_2\leq n^{-M}}\|\Gamma_{1}\|_2=O(n^{2c+2\gamma_{2}-\gamma\alpha' M})$.

We first consider the case that (B4) holds. Then
$\ep \max_{j}\sup_{\|\tee-\tee_{j}\|_2\leq n^{-M}}\|\Gamma_{2}\|_2=O(n^{\gamma_{2}/2+1+c_2-c_3M})$. Hence, by Markov's inequality, for any $\gamma_{1}>0$, by letting $M$ be sufficiently large, we have
\begin{eqnarray}\label{abd}
\pr\Big{(}\max_{j}\sup_{\|\tee-\tee_{j}\|_2\leq n^{-M}}\Big{\|}\frac{\frac{1}{m}\sum_{i\in B_{t}}(h(\tee,\xii_i)-h(\tee_{j},\xii_i))}{\sqrt{\tilde{b}(\tee_{j})}}\Big{\|}_2\geq c\sqrt{\frac{q\log n}{m}}\Big{)}\cr
=O(n^{-\gamma_{1}}).
\end{eqnarray}

We next prove (\ref{abd}) under (B4$^{*}$). By the proof of Lemma \ref{lem:bn}, we have
\begin{eqnarray*}
	\Big{\|}\sum_{i\in B_{t}}(h(\tee,\xii_i)-h(\tee_{j},\xii_i))\Big{\|}_2&\leq& 2\max_{1\leq l\leq d_{q}}\Big{|}\v'_{l}\sum_{i\in B_{t}}(h(\tee,\xii_i)-h(\tee_{j},\xii_i))\Big{|}\cr
	&\leq&2\max_{1\leq l\leq d_{q}}\Big{|}\sum_{i\in B_{t}}\sup_{\|\tee-\tee_{j}\|_2\leq n^{-M}}|\v'_{l}H(\tee,\tee_{j},\xii_i)|\Big{|},
\end{eqnarray*}
where $H(\tee,\tee_{j},\xii)=h(\tee,\xii)-h(\tee_{j},\xii).$ It is easy to see from (B4$^{*}$) that, for sufficiently large $M$,
\begin{eqnarray*}
	\max_{j}\max_{1\leq l\leq s_{q}}\frac{\Big{|}\sum_{i\in B_{t}}\ep\sup_{\|\tee-\tee_{j}\|_2\leq n^{-M}}|\v'_{l}H(\tee,\tee_{j},\xii_i)|\Big{|}}{\sqrt{\tilde{b}(\tee_{j})}}=o(\sqrt{\frac{q\log n}{m}}).
\end{eqnarray*}
Set $\mathcal{H}_{l,j}(\xii_i)=\sup_{\|\tee-\tee_{j}\|_2\leq n^{-M}}|\v'_{l}H(\tee,\tee_{j},\xii_i)|/b^{\alpha/2}(\tee_{j})$. By (B4$^{*}$) and Holder's inequality, we have
$$\max_{j}\sum_{i\in B_{t}}\ep (\mathcal{H}_{l,j}(\xii_i))^{2}\exp(t_{0} \mathcal{H}_{l,j}(\xii_i)/2)\leq mq^{c_{2}/2}n^{-c_{3}M/2}.$$
We now take $B^{2}_{n}=c_{5}m\tilde{b}(\tee_{j})/b^{\alpha}(\tee_{j})$ and $x^{2}=c_{5}q\log n$ in Lemma 1 in \cite{cailiu2011}, noting that
$mq^{c_{2}/2}n^{-c_{3}M/2}\leq B^{2}_{n}$ and $x^{2}\leq B^{2}_{n}$, we have for any $\gamma, M>0$, there exist $c,c_{5}>0$ such that uniformly in $j$,
\begin{eqnarray*}
	&&\pr\Big{(}\Big{|}\frac{\sum_{i\in B_{t}}[\sup_{\|\tee-\tee_{j}\|_2\leq n^{-M}}|\v'_{l}H(\tee,\tee_{j},\xii_i)|-\ep \sup_{\|\tee-\tee_{j}\|_2\leq n^{-M}}|\v'_{l}H(\tee,\tee_{j},\xii_i)|]}{m\sqrt{\tilde{b}(\tee_{j})}}\geq c\sqrt{\frac{q\log n}{m}}\Big{)}\cr
	&&=O(n^{-\gamma q}).
\end{eqnarray*}
This proves (\ref{abd}) under (B4$^{*}$) by noting that $s_{q}=O(n^{2qM})$ and $d_{q}\leq 5^{q}$.

Now it suffices to show that
\begin{eqnarray}\label{ads}
\pr\Big{(}\max_{j}\frac{\Big{\|}\frac{1}{m}\sum_{i\in B_{t}}h(\tee_{j},\xii_i)\Big{\|}_2}{\sqrt{\tilde{b}(\tee_{j})}} \geq c\sqrt{\frac{q\log n}{m}}\Big{)}=O(n^{-\gamma_{1} }).
\end{eqnarray}
Let $\zet_{i}=h(\tee_{j},\xii_i)/b^{\alpha/2}(\tee_{j})$, $i\in B_{t}$. By (B3), it is easy to see that
\begin{eqnarray*}
	\sup_{\|\v\|_2=1}\sum_{i\in B_{t}}\ep (\v'\zet_{i})^{2}\exp(t_{0}|\v'\zet_{i}|)\leq C_2m[b(\tee_{j})]^{1-\alpha}
\end{eqnarray*}
for some $C_2>0$.
Take $x=\sqrt{(\gamma_{1}+2M)q\log n}$ and 
\begin{eqnarray*}
	B^{2}_{n}&=&4(C_2+\gamma_{1}+2M+1)\Big{(}m[b(\tee_{j})]^{1-\alpha}+q\log n+m(b(\tee_{j}))^{-\alpha}n^{-\gamma_{2}}\Big{)}\cr
	&=&4(C_2+\gamma_{1}+2M+1)m\tilde{b}(\tee_{j})/b^{\alpha}(\tee_{j}).
\end{eqnarray*}
Note that $2\sqrt{q+x^{2}}\leq B_{n}$. By Lemma \ref{lem:bn}, we obtain (\ref{ads}) by letting  $c$ be sufficiently large.\qed

Let $\bar{g}(\tee,\xii)=g(\tee,\xii)-\ep g(\tee,\xii)$.  For some $c_4>0$,
define
\begin{eqnarray*}
	\mathcal{C}_{t}&=&\Big{\{}\sup_{\|\tee-\tee^*\|_2\leq c_4}\Big{\|}\frac{1}{m}\sum_{i\in B_{t}}\bar{g}(\tee,\xii_i)\Big{\|}_2\leq c\sqrt{\frac{p\log n}{m}}\Big{\}},\cr
	\mathcal{C}&=&\Big{\{}\sup_{\|\tee-\tee^*\|_2\leq c_4}\Big{\|}\frac{1}{n}\sum_{i=1}^{n}\bar{g}(\tee,\xii_i)\Big{\|}_2\leq c\sqrt{\frac{p\log n}{n}}\Big{\}},
\end{eqnarray*}
where $c$ is sufficiently large. 

\begin{lemma}\label{lem:ct}
	Under (C3) or (C3$^*$) and $p\log n=o(m)$, for any $\gamma>0$, there exists a constant  $c_4>0$ such that   $$\pr(\mathcal{C}_{t}\cap\mathcal{C})\geq 1-O(n^{-\gamma}).$$ 
	The same result holds with $B_{t}$ being replaced by $H_{t}$.
\end{lemma}

\noindent{\bf Proof.} 
In Lemma \ref{le1}, take $\u=\tee$, $\u_{0}=\tee^{*}$, $q=p$, $\alpha=0$ and $h( \tee,\xii)=g( \tee,\xii)-\ep g(\tee,\xii)$. Then (C3) (or (C3$^*$))  implies that (B1)-(B4) (or (B4$^*$), respectively) hold with $\alpha=0$, and $b(\tee)=C$ for some large $C$. So we have $$\pr(\mathcal{C}_{t}\cap\mathcal{C})\geq 1-O(n^{-\gamma})$$ for any large $\gamma$.\qed

\begin{lemma}\label{lem:an} Suppose that $p\rightarrow \infty$, $r_i=c_0/\max(p,i^{\alpha})$ for $c_0>0$ and $0<\alpha\leq 1$. Let $c>0$, $\tau>0$ and  $d\geq1$.
	
	\noindent (1) For a positive sequence $\{a_{i}\}$ that satisfies $a_{i}\leq (1-cr_{i})a_{i-1}+r^{d}_{i}b_{n}$, $1\leq i\leq n$, we have
	$a_{i}\leq C(r^{d-1}_{i}b_{n}+i^{-\gamma})$ for any $\gamma>0$ and all $i\geq p^{1/\alpha+\tau}$  by letting $c_0$ be sufficiently large.
	
	\noindent (2) For a positive sequence  $\{a_{i}\}$ that satisfies $a_{i}\geq (1-cr_{i})a_{i-1}+r^{d}_{i}b_{n}$, $1\leq i\leq n$, we have
	$a_{i}\geq Cr^{d-1}_{i}b_{n}$ for all $i\geq p^{1/\alpha+\tau}$  by letting $c_0$ be sufficiently large.
\end{lemma}
\noindent{\bf Proof.}
We first prove the first claim.  For  $i\geq p^{1/\alpha+\tau}$, we have
\begin{align}\label{eq:ai}
a_i&\leq (1-cr_{i})a_{i-1}+r^{d}_{i}b_{n}\cr
&=  a_0\prod\limits_{j=1}^i(1-cr_j)+b_n\sum\limits_{k=1}^{i}r_{k}^d\prod_{j=k}^{i-1}(1-cr_{j+1})\cr
&\leq a_0 \exp\big(-c\sum\limits_{j=1}^ir_j\big)+b_n\sum\limits_{k=1}^{i}r_{k}^d\exp\big(-c\sum_{j=k}^{i-1}r_{j+1}\big)\cr
&\leq  a_0\exp\Big(-\tilde{c}\big(p_\alpha/p+\frac12\int_{p_\alpha}^{i}\frac{1}{x^\alpha}\mathrm{d} x\big)\Big)+b_n\sum\limits_{k=p_\alpha+1}^{i}r_{k}^d\exp\Big(-\frac{\tilde{c}}{2}\int_{k}^{i}\frac{1}{x^\alpha}\mathrm{d} x\Big)\cr
&+b_n\sum\limits_{k=1}^{p_\alpha}r_{k}^d\exp\Big(-\tilde{c}\big(\frac{p_\alpha-k}{p}+\frac{1}{2}\int_{p_\alpha}^{i}\frac{1}{x^\alpha}\mathrm{d} x\big)\Big)\cr
&=  a_0\exp\Big(-\tilde{c}\big(p_\alpha/p+\frac12\int_{p_\alpha}^{i}\frac{1}{x^\alpha}\mathrm{d} x\big)\Big)+c_0^db_n\sum\limits_{k=p_\alpha+1}^{i}k^{-\alpha d}\exp\Big(-\frac{\tilde{c}}{2}\int_{k}^{i}\frac{1}{x^\alpha}\mathrm{d} x\Big)\cr
&+c_0^db_n\sum\limits_{k=1}^{p_\alpha}p^{-d}\exp\Big(-\tilde{c}\big(\frac{p_\alpha-k}{p}+\frac{1}{2}\int_{p_\alpha}^{i}\frac{1}{x^\alpha}\mathrm{d} x\big)\Big),
\end{align}
where $p_\alpha=\lfloor p^{1/\alpha}\rfloor$, $\tilde{c}=c_0c$, and $\alpha$, $c_0$ are defined in the step-size $r_i$.

When $\alpha=1$, we have
\begin{eqnarray*}
	\eqref{eq:ai}&= & \frac{a_0p^{\tilde{c}/2}\mathrm{e}^{-\tilde{c}}}{i^{\tilde{c}/2}}+c_0^db_n\sum\limits_{k=p+1}^{i}\frac{k^{\tilde{c}/2-d}}{i^{\tilde{c}/2}}+c_0^db_n\sum\limits_{k=1}^{p}\frac{p^{\tilde{c}/2-d}\exp(\tilde{c}k/p-\tilde{c})}{i^{\tilde{c}/2}}\cr
	&\leq & \frac{a_0p^{\tilde{c}/2}\mathrm{e}^{-\tilde{c}}}{i^{\tilde{c}/2}}+c_0^db_ni^{1-d}+\frac{c_0^db_np^{\tilde{c}/2-d+1}}{i^{\tilde{c}/2}}\cr
	&\leq & C(r_i^{d-1}b_n+i^{-\gamma}),
\end{eqnarray*}
when $c_0$ is large enough such that $\tilde{c}=c_0c\geq 2\max(d,\gamma)(1+1/\tau)$.

When $\alpha<1$, for any $\kappa>0$ and $1\leq u< i$, we have
\begin{align*}
\int_u^{i}x^{-\alpha d}&\exp\big(\frac{\kappa x^{1-\alpha}}{1-\alpha}\big)\mathrm{d}x\cr
&=\frac{1}{\kappa}x^{-\alpha d+\alpha}\exp\big(\frac{\kappa x^{1-\alpha}}{1-\alpha}\big)\Big|_u^i-\int_u^i\frac{\alpha-\alpha d}{\kappa}x^{-\alpha d+\alpha-1}\exp\big(\frac{\kappa x^{1-\alpha}}{1-\alpha}\big)\mathrm{d}x\cr
&\leq \frac{1}{\kappa}x^{-\alpha d+\alpha}\exp\big(\frac{\kappa x^{1-\alpha}}{1-\alpha}\big)\Big|_u^i+u^{\alpha-1}\int_u^i\frac{\alpha(d-1)}{\kappa}x^{ -\alpha d}\exp\big(\frac{\kappa x^{1-\alpha}}{1-\alpha}\big)\mathrm{d}x.
\end{align*}
Therefore, we have
\begin{equation}\label{eq:lemalphad}
\int_u^{i}x^{-\alpha d}\exp\big(\frac{\kappa x^{1-\alpha}}{1-\alpha}\big)\mathrm{d}x\leq \frac{1}{\kappa-\alpha d+\alpha}x^{-\alpha d+\alpha}\exp\big(\frac{\kappa x^{1-\alpha}}{1-\alpha}\big)\Big|_u^i
\end{equation}
for $\kappa>\alpha (d-1)$. By \eqref{eq:lemalphad}, we have for $i\geq p^{1/\alpha+\tau}$,
\begin{eqnarray*}
	\eqref{eq:ai}&= & a_0\exp\Big(-\tilde{c}\big(\frac{p_\alpha}{p}+\frac{i^{1-\alpha}-p_\alpha^{1-\alpha}}{2-2\alpha}\big)\Big)+c_0^db_n\sum\limits_{k=p_\alpha+1}^{i}k^{-\alpha d}\exp\Big(-\frac{\tilde{c}\big(i^{1-\alpha}-k^{1-\alpha}\big)}{2-2\alpha}\Big)\cr
	&&+c_0^db_n\sum\limits_{k=1}^{p_\alpha}p^{-d}\exp\Big(-\tilde{c}\big(\frac{p_\alpha-k}{p}+\frac{i^{1-\alpha}-p_\alpha^{1-\alpha}}{2-2\alpha}\big)\Big)\cr
	&\leq & a_0\exp\Big(-\frac{\tilde{c}(i^{1-\alpha}-p_\alpha^{1-\alpha})}{2-2\alpha}\Big)+c_0^db_ni^{-\alpha d}\cr
	&&+c_0^db_n\exp\Big(-\frac{\tilde{c}i^{1-\alpha}}{2-2\alpha}\Big)\int_{p_\alpha+1}^{i}x^{-\alpha d}\exp\Big(\frac{\tilde{c}x^{1-\alpha}}{2-2\alpha}\Big)\mathrm{d}x\cr
	&&+c_0^db_np_{\alpha}p^{-d}\exp\Big(-\frac{\tilde{c}(i^{1-\alpha}-p_\alpha^{1-\alpha})}{2-2\alpha}\big)\cr
	&\leq & a_0\exp\big(-\frac{\tilde{c}(i^{1-\alpha}-p_\alpha^{1-\alpha})}{2-2\alpha}\Big)+c_0^db_ni^{-\alpha d}\cr
	& &+{c_0^db_n}\Big(\frac{i^{-\alpha (d-1)}}{\tilde{c}/2-\alpha d+\alpha}+p_{\alpha}p^{-d}\exp\big(-\frac{\tilde{c}(i^{1-\alpha}-p_\alpha^{1-\alpha})}{2-2\alpha}\big)\Big)\cr
	&\leq& C(r_i^{d-1}b_n+i^{-\gamma})
\end{eqnarray*}
for large enough $c_0$ such that $\tilde{c}>2\alpha(d-1)$.

To prove the second claim, we first recall that $p\rightarrow \infty$ and $\sup_{i\geq 1}r_{i}=o(1)$. Hence $1-cr_{j}\geq \exp(-2cr_{j})$ for all $j$.
Then
\begin{eqnarray}\label{eq:aii}
a_i&\geq&a_0 \exp\big(-2c\sum\limits_{j=1}^ir_j\big)+b_n\sum\limits_{k=1}^{i}r_{k}^d\exp\big(-2c\sum_{j=k}^{i-1}r_{j+1}\big)\cr
&\geq&b_n\sum\limits_{k=1}^{i}r_{k}^d\exp\Big(-2\tilde{c}\int_{k}^{i}\frac{1}{x^\alpha}\mathrm{d} x\Big).
\end{eqnarray}

When $\alpha=1$,  we have
\begin{eqnarray*}
	\eqref{eq:aii}\geq c_0^db_ni^{-2\tilde{c}}\sum\limits_{k=p+1}^ik^{2\tilde{c}- d}\geq\frac{c_0^db_n(i^{-d+1}-i^{-2\tilde{c}}p^{2\tilde{c}-d+1})}{2\tilde{c}-d+1}\geq c_1r_i^{d-1}b_n,
\end{eqnarray*}
for $2\tilde{c}>d-1$ and $i\geq p^{1+\tau}$.

When $\alpha<1$,  we have for  $i\geq p^{1/\alpha+\tau}$,
\begin{eqnarray*}
	\eqref{eq:aii}&\geq& c_0^db_n\sum\limits_{k=p_\alpha+1}^{i}k^{-\alpha d}\exp\Big(-\frac{2\tilde{c}\big(i^{1-\alpha}-k^{1-\alpha}\big)}{1-\alpha}\Big)\cr
	&\geq &c_0^db_n\exp\Big(-\frac{2\tilde{c}i^{1-\alpha}}{1-\alpha}\Big)\int_{p_\alpha}^{i}x^{-\alpha d}\exp\Big(\frac{2\tilde{c}x^{1-\alpha}}{1-\alpha}\Big)\mathrm{d}x\cr
	&\geq &\frac{c_0^db_n}{2\tilde{c}}x^{-\alpha d+\alpha}\exp\big(\frac{2\tilde{c} (x^{1-\alpha}-i^{1-\alpha})}{1-\alpha}\big)\Big|_{p_\alpha}^i\cr
	&=&\frac{1}{2c}r_i^{d-1}b_n-\frac{c^{d}_{0}}{2\tilde{c}} p_{\alpha}^{-\alpha d+\alpha}b_n\exp\big(\frac{\tilde{c} (p_\alpha^{1-\alpha}-i^{1-\alpha})}{1-\alpha}\big)\cr
	&\geq& Cr_i^{d-1}b_n.
\end{eqnarray*}
The proof is complete.
\qed

\section{Additional Simulations}
\label{sec:add_simu}

In this section, we provide additional simulation studies. We investigate the case of correlated design, the effect of the quality of the initial estimator. The data generating process has been described in Section \ref{sec:exp} in the main text.

\begin{table}[t!]
	\centering
	\caption{$L_2$-errors when covariates $\X$ are generated from different underlying distributions. Here the total sample size $N=10^5$ and dimension $p=100$, and the number of machines $L=20$.  Denote by $\hat\tee_{\mathrm{DC}}$ the DC-SGD estimator and  $\teedis$ the Dis-FONE.}
	\vspace{.25cm}
	\begin{tabular}{cc|cccc|cc}
		\hline
		Model & Covariates &	\multicolumn{4}{c|}{$L_2$-distance to the truth $\tee^*$} & 	\multicolumn{2}{l}{$L_2$-distance }\\
		& &	\multicolumn{4}{c|}{} & 	\multicolumn{2}{c}{to ERM $\hat\tee$}\\
		&& $\hat\tee_0$ & $\hat\tee_{\mathrm{DC}}$ & $\teedis$  & $\hat\tee$ & $\hat\tee_{\mathrm{DC}}$  & $\teedis$ \\
		\hline
		\multicolumn{2}{l|}{Logistic}                  &       &       &       &       &       &       \\
		& Identity          & 1.310 & 0.467 & 0.104 & 0.092 & 0.453 & 0.037 \\
		& Toeplitz ($0.3$)  & 1.427 & 0.535 & 0.114 & 0.100 & 0.525 & 0.045 \\
		& Toeplitz ($0.5$)  & 1.634 & 0.694 & 0.138 & 0.117 & 0.685 & 0.055 \\
		& Toeplitz ($0.7$)  & 1.855 & 0.990 & 0.159 & 0.143 & 0.987 & 0.057 \\
		& Equi Corr ($0.3$) & 1.398 & 0.548 & 0.119 & 0.103 & 0.536 & 0.039 \\
		& Equi Corr ($0.5$) & 2.015 & 0.807 & 0.158 & 0.137 & 0.792 & 0.050 \\
		& Equi Corr ($0.7$) & 2.087 & 1.279 & 0.181 & 0.163 & 1.273 & 0.061 \\
		\hline
		\multicolumn{2}{l|}{Quantile}                  &       &       &       &       &       &       \\
		& Identity          & 0.455 & 0.089 & 0.048 & 0.043 & 0.084 & 0.025 \\
		& Toeplitz ($0.3$)  & 0.500 & 0.140 & 0.055 & 0.046 & 0.138 & 0.031 \\
		& Toeplitz ($0.5$)  & 0.589 & 0.226 & 0.066 & 0.055 & 0.225 & 0.043 \\
		& Toeplitz ($0.7$)  & 0.775 & 0.422 & 0.100 & 0.072 & 0.424 & 0.078 \\
		& Equi Corr ($0.3$) & 0.542 & 0.155 & 0.055 & 0.051 & 0.153 & 0.026 \\
		& Equi Corr ($0.5$) & 0.637 & 0.329 & 0.064 & 0.060 & 0.328 & 0.027 \\
		& Equi Corr ($0.7$) & 0.814 & 0.607 & 0.084 & 0.078 & 0.610 & 0.039 \\
		\hline
	\end{tabular}
	\label{table:design}
\end{table}

\subsection{Effect of the underlying distribution of covariates $\X$}
\label{subsec:design}

\begin{table}[t!]
	\centering
	\caption{$L_2$-errors when covariates $\X$ are generated from different underlying distributions. Here the total sample size $N=10^5$ and dimension $p=100$, and the number of machines $L=20$.  Denote by $\hat\tee$ the ERM estimator, $\teedis$ the Dis-FONE estimator with $K$ rounds, $\hat\tee_{K,\text{rand}}$ the Dis-FONE(random) estimator with $K$ rounds, and  $\hat\tee_{K,\text{avg}}$ the Dis-FONE(avg) with $K$ rounds.}
	\vspace{.25cm}
	\begin{tabular}{cc|ccccc|ccc}
		\hline
		Model & Covariates &	\multicolumn{5}{c|}{$L_2$-distance to the truth $\tee^*$} & 	\multicolumn{3}{l}{$L_2$-distance to ERM $\hat\tee$}\\
		&& $\hat\tee_0$ & $\hat\tee$ & $\teedis$  & $\hat\tee_{K,\text{rand}}$ & $\hat\tee_{K,\text{avg}}$ & $\teedis$  & $\hat\tee_{K,\text{rand}}$ & $\hat\tee_{K,\text{avg}}$   \\
		\hline
		\multicolumn{2}{l|}{Logistic}                  &       &       &       &       &       &       \\
		& $t_{2.5}$ & 1.184 & 0.082 & 0.091 & 0.090 & 0.082 & 0.016 & 0.013 & 0.003 \\
		& $t_5$     & 1.355 & 0.083 & 0.083 & 0.083 & 0.083 & 0.002 & 0.002 & 0.003 \\
		& $t_{10}$    & 1.357 & 0.093 & 0.094 & 0.094 & 0.094 & 0.002 & 0.002 & 0.002 \\
		\hline
		\multicolumn{2}{l|}{Quantile}                  &       &       &       &       &       &       \\
		& $t_{2.5}$ & 0.304 & 0.029 & 0.029 & 0.029 & 0.027 & 0.025 & 0.024 & 0.018 \\
		& $t_5$     & 0.362 & 0.036 & 0.038 & 0.037 & 0.037 & 0.019 & 0.018 & 0.018 \\
		& $t_{10}$    & 0.366 & 0.041 & 0.043 & 0.041 & 0.045 & 0.016 & 0.016 & 0.016\\
		\hline
	\end{tabular}
	\label{table:hetero}
\end{table}

Suppose that $(X_{i,1},X_{i,2},\dots,X_{i,p-1})$ follows a multivariate normal distribution $\mathcal{N}(\mathbf{0},\S^0)$ for $i=1,2,\dots, N$. In the previous simulation studies, we adopt the covariance matrix $\S^0=\mathbf{I}_{p-1}$. In this section, we consider two different structures of $\S^0$:
\begin{itemize}
	\setlength\itemsep{2pt}
	\item Toeplitz:\qquad $\S^0_{i,j}=\varsigma^{|i-j|}$,
	\item Equi Corr: \quad $\S^0_{i,j}=\varsigma$ for all $i\neq j$,\quad $\Sigma_{i,i}=1$ for all $i$.
\end{itemize}
For both structures, we consider the correlation parameter $\varsigma$ varying from $\{0.3, 0.5, 0.7\}$. In Table \ref{table:design}, we report the $L_2$-estimation errors of the proposed estimators. In all cases of the covariance matrix, Dis-FONE results are very close to those of the ERM in \eqref{eq:erm}. Meanwhile, the $L_2$-errors of DC-SGD and SGD increase significantly when the correlation of the design matrix increases. 
We further consider the cases when the underlying distribution of the covariates $\X$ are not identical among different machines. 
In particular, on half of the local machines, the covariate $(X_{i,1},X_{i,2},\dots,X_{i,p-1})$ follows a standard normal distribution $\mathcal{N}(\mathbf{0},\mathbf{I}_{p-1})$. On the other half, the covariate follows a Student's $t_\nu$-distribution standardized by its standard deviation such that it has the covariance matrix  $\mathbf{I}_{p-1}$. As the variance of a $t$-distribution only exists when $\nu>2$, we choose and report results from $\nu=2.5, 5, 10$.

We consider three candidate estimators: Dis-FONE and two of its variants. We denote by Dis-FONE(random) the algorithm that uses a random local machine to implement FONE in different rounds, and we denote by Dis-FONE(avg) the algorithm that lets each local machine run FONE based on the aggregated gradient simultaneously and then take the average of all the local estimators for each round. We further denote by $\hat\tee_{K,\text{rand}}$ and $\hat\tee_{K,\text{avg}}$ their estimators after $K$ rounds, respectively. 

Table \ref{table:hetero} reports the $L_2$ estimation errors for the three algorithms. From Table \ref{table:hetero}, we can see that the three estimators achieve almost the same performance when $\nu$ is rather large (e.g., $\nu=5$ or $10$). When $\nu=2.5$ (i.e., the half of the data is heavy tailed), Dis-FONE(avg) achieves slightly better performance than the other two methods.

\subsection{Effect of the initial estimator $\hat\tee_0$}
\label{subsec:init}

\begin{table}[t!]
	\centering
	\caption{$L_2$-errors when varying the size $n_0$ of the fresh sample used in constructing the initial estimator $\widehat{\tee}_0$. Here the total sample size $N=10^5$ and dimension $p=100$, and the number of machines $L=20$.  Denote by $\hat\tee_{\mathrm{DC}}$ the DC-SGD estimator and  $\teedis$ the Dis-FONE.}
	\vspace{.25cm}
	\begin{tabular}{cc|cccc|cc|ccc}
		\hline
		Model &$n_0$  &	\multicolumn{4}{c|}{$L_2$-distance to the $\tee^*$} & 	\multicolumn{2}{l|}{$L_2$-distance to $\hat\tee$}&\multicolumn{3}{c}{Norm of Average Gradient}\\
		& &	\multicolumn{4}{c|}{$\|\tee-\tee^*\|_2$} & 	\multicolumn{2}{c|}{$\|\tee-\hat\tee\|_2$}&\multicolumn{3}{c}{$\tau_n=\Big\|\frac1N\sum_{i=1}^Ng(\tee,\xii)\Big\|_2$}\\
		&& $\hat\tee_0$ & $\hat\tee_{\mathrm{DC}}$ & $\teedis$  & $\hat\tee$ & $\hat\tee_{\mathrm{DC}}$ & $\teedis$ &$\hat\tee_0$ & $\hat\tee_{\mathrm{DC}}$ & $\teedis$  \\
		\hline
		Logistic& &&&&&&&&&\\
		&$5p$  & 3.095 & 1.211 & 0.102 & 0.093 & 1.203 & 0.040 &0.1771 &0.0296 & 0.0001\\
		&$10p$ & 1.251 & 0.447 & 0.103 & 0.093 & 0.445 & 0.038 &0.1173 &0.0212 & 0.0000\\
		&$20p$ & 0.791 & 0.266 & 0.102 & 0.093 & 0.265 & 0.035 &0.0806 &0.0165 & 0.0000\\
		\hline
		Quantile& &&&&&&&&&\\
		&$5p$  & 0.681 & 0.109 & 0.050 & 0.044 & 0.105 & 0.027&0.2006 &0.0333 & 0.0175 \\
		&$10p$ & 0.450 & 0.079 & 0.047 & 0.043 & 0.073 & 0.020 &0.1456 &0.0291 & 0.0149\\
		&$20p$ & 0.311 & 0.082 & 0.048 & 0.043 & 0.077 & 0.024&0.0972 & 0.0208& 0.0139\\
		\hline
	\end{tabular}
	\label{table:init}
\end{table}

Recall that our methods require a consistent initial estimator $\widehat{\tee}_0$ to guarantee the convergence. We  investigate the effect on the accuracy of the initial estimator in our methods. In particular, we fix the total sample size $N=10^5$, the dimension $p=100$, the number of machines $L=20$ and varies $n_0$ from $5p$, $10p$ and $20p$, where $n_0$ denotes the size of the fresh sample used to construct the initial estimator $\hat\tee_0$. From Table \ref{table:init}, the error of the initial estimator $\hat\tee_0$ decreases as $n_0$ increases. As a consequence, DC-SGD has a better performance. On the other hand, the $L_2$-errors of Dis-FONE have already been  quite small even when the initial estimator is less accurate. We further add a column to present the $\ell_2$ norm of the average gradient $\tau_n=\Big\|\frac1N\sum_{i=1}^Ng(\tee,\xii)\Big\|_2$, where $\tee$ can be specified as $\tee_0$, $\teeDC$, or $\teedis$.

\begin{table}[t!]
	\centering
	\caption{$L_2$-errors when varying the constant in the step-size specification. Here the total sample size $N=10^5$ and dimension $p=100$, and the number of machines $L=20$.  Denote by $\hat\tee_{\mathrm{DC}}$ the DC-SGD estimator and  $\teedis$ the Dis-FONE.}
	\vspace{.25cm}
	\begin{tabular}{cc|cccc|cc}
		\hline
		Model & Step-size &	\multicolumn{4}{c|}{$L_2$-distance to the truth $\tee^*$} & 	\multicolumn{2}{l}{$L_2$-distance }\\
		& Constant&	\multicolumn{4}{c|}{} & 	\multicolumn{2}{c}{to ERM $\hat\tee$}\\
		&& $\hat\tee_0$ & $\hat\tee_{\mathrm{DC}}$ & $\teedis$  & $\hat\tee$ & $\hat\tee_{\mathrm{DC}}$  & $\teedis$ \\
		\hline
		\multicolumn{2}{l|}{Logistic}                  &       &       &       &       &       &       \\
		& $\tilde{c}_0$  & 1.251 & 0.447 & 0.103 & 0.093 & 0.453 & 0.037 \\
		& $2\tilde{c}_0$  & 1.264 & 0.532 & 0.126 & 0.092 & 0.551 & 0.042 \\
		& $5\tilde{c}_0$  & 1.257 & 0.623& 0.148 & 0.092 & 0.614 & 0.050 \\
		& $\tilde{c}_0/2$& 1.295 & 0.453 & 0.098 & 0.092 & 0.432 & 0.036 \\
		& $\tilde{c}_0/5$ & 1.274 & 0.447 & 0.099 & 0.092 & 0.427 & 0.036 \\
		\hline
		\multicolumn{2}{l|}{Quantile}                  &       &       &       &       &       &       \\
		& $\tilde{c}_0$   & 0.450 & 0.079 & 0.047 & 0.043 & 0.073 & 0.020 \\
		& $2\tilde{c}_0$ & 0.464 & 0.141 & 0.052 & 0.043 & 0.098 & 0.028 \\
		&$5\tilde{c}_0$   & 0.452 & 0.178 & 0.059 & 0.043 & 0.116 & 0.033 \\
		& $\tilde{c}_0/2$  & 0.461 & 0.079 & 0.047 & 0.043 & 0.078 & 0.021 \\
		& $\tilde{c}_0/5$  & 0.439 & 0.077 & 0.046 & 0.043 & 0.075 & 0.024 \\
		\hline
	\end{tabular}
	\label{table:step}
\end{table}

{
\subsection{Effect of the step-size $\eta_n$}
\label{subsec:step}
In this section, we study the effect of constant in the step-size specification  $r_i=c_0/\max(i^\alpha,p)$ for DC-SGD and  $\eta=c_0'm/n$ for Dis-FONE. In the previous experiments, we choose the best $c_0$ that achieves the smallest objective function in \eqref{eq:erm}  with $\tee=\teeSGD^{(1)}$ using data points from the first machine  (see Algorithm \ref{algo:dc-sgd}), i.e.,
$
\tilde c_0=\argmin_{c \in \mathcal{C}} \frac{1}{n} \sum_{i=1}^n f(\teeSGD^{(1)}, \xii_i^{(1)}),
$
where $\{\xii_i^{(1)},i=1,2,\dots,n\}$ denotes the samples on the first machine. Analogously, we choose the best tuning constant $\tilde c_0$ that achieves the smallest objective in  \eqref{eq:erm} with $\tee=\hat\tee_{\mathrm{dis},1}$ and samples from the first machine. Here,  $\hat\tee_{\mathrm{dis},1}$ is the output of Dis-FONE after the first round of the algorithm. That is, 
$
\tilde c_0=\argmin_{c \in \mathcal{C}} \frac{1}{n} \sum_{i=1}^n f(\hat\tee_1, \xii_i^{(1)}).
$

In the step-size specification for DC-SGD, i.e., $r_i=c_0/\max(i^\alpha,p)$, we specify the constant $c_0$ from $5\tilde{c}_0$, $2\tilde{c}_0$, $\tilde{c}_0$, $\tilde{c}_0/2$ to $\tilde{c}_0/5$. For Dis-FONE, i.e., $\eta_n=c_0'm/n$, we vary the constant $c_0'$ from $5\tilde{c}_0$ to $\tilde{c}_0/5$, respectively, and adjust the number of iterations $T$ in the FONE step accordingly to keep $\eta_nT$ unchanged. We report the $L_2$-error for the both estimators in Table \ref{table:step}. As we can see, our method is pretty robust with respect to different choices of stepsizes.

\subsection{Comparison to existing state-of-art methods}
\label{subsec:comp}

\begin{table}[t!]
	\centering
	\caption{Upper rows: comparison to $\hat\tee_{\mathrm{CSL}}$ \citep{jordan2016communication} under the logistic regression settings. Lower rows: comparison to $\hat\tee_{\mathrm{DC-QR}}$ \citep{volgushev2017distributed} under the quantile regression settings. Report the $L_2$-distance to the truth $\tee^*$ and computation time. Here the total sample size $N=5\times 10^5$ and the number of machines $L=20$. Denote by $\hat\tee_{\mathrm{DC-SGD}}$ the DC-SGD estimator and  $\teedis$ the Dis-FONE.}
	\vspace{.25cm}
	\begin{tabular}{cc|ccccc|cccc}
		\hline
		& $p$ & \multicolumn{5}{c|}{$L_2$-distance to the truth $\tee^*$} & \multicolumn{4}{c}{Computation time (seconds)} \\
		& & $\hat\tee_0$ & $\hat\tee_{\mathrm{DC-SGD}}$  & $\teedis$  & $\hat\tee$  & $\hat\tee_{\mathrm{CSL}}$& $\hat\tee_{\mathrm{DC-SGD}}$  & $\teedis$  & $\hat\tee$  & $\hat\tee_{\mathrm{CSL}}$\\
		\hline
		Logistic& &&&&&\\
		& 100 & 1.342 & 0.313  & 0.046  & 0.042 & 0.043 & 0.050 & 0.104 & 3.428  & 0.143\\
		& 200 & 1.833 & 0.874  & 0.073  & 0.068 & 0.068 & 0.174 & 0.581 & 6.445 & 1.144\\
		& 500 & 4.835 & 3.885   & 0.141 & 0.130 & 0.133 & 0.671 & 3.401 & 15.267 & 13.471\\
		\hline
		& $p$ & \multicolumn{5}{c|}{$L_2$-distance to the truth $\tee^*$} & \multicolumn{4}{c}{Computation time (seconds)} \\
		&  & $\hat\tee_0$ & $\hat\tee_{\mathrm{DC-SGD}}$ & $\teedis$  & $\hat\tee$  & $\hat\tee_{\mathrm{DC-QR}}$& $\hat\tee_{\mathrm{DC-SGD}}$  & $\teedis$  & $\hat\tee$  & $\hat\tee_{\mathrm{DC-QR}}$\\
		\hline
		Quantile& &&&&&\\
		& 100 & 0.451     & 0.043     & 0.029 & 0.025    & 0.039  & 0.069 & 1.514 & 29.632 &1.061\\
		& 200 & 0.719     & 0.067     & 0.041 & 0.037    & 0.065 &  0.293 & 4.156 & 113.235 & 4.614\\
		& 500 & 1.294     & 0.105     & 0.074 & 0.057   & 0.098 & 6.014 & 17.024 & 848.149 & 32.497\\
		\hline
	\end{tabular}
	\label{table:csl}
\end{table}

 In this section, we compare the Dis-FONE algorithm with two baseline methods \cite{jordan2016communication, volgushev2017distributed}. 
For smooth loss functions, our method can be viewed as a stochastic gradient implementation of \cite{jordan2016communication}, where we solve the Newton-step in the CSL algorithm in \cite{jordan2016communication} by mini-batch SGD. On the other hand, CSL requires the loss function to be second order differentiable, which is not applicable to non-smooth loss functions; while our method provides a general framework for both smooth and non-smooth loss functions. 

For quantile regression (QR), \cite{volgushev2017distributed} proposed a one-shot divide-and-conquer (DC) algorithm, which solves QR exactly on each local machine, and then takes the average. We compare our algorithm with CSL for logistic regression, and with DC-QR \citep{volgushev2017distributed} for quantile regression. 

We report the computation time of the candidate methods in a distributed environment and compare them with the time of the non-distributed ``oracle'' ERM estimator. The experiment is performed on a Linux cluster containing 100 computing nodes inter-connected by high speed networks. Linux operating system runs on each of the nodes individually. On each computer node, we use an Intel Xeon E5-2690v2 3.0GHz CPU with 8GB memory. The computation time are reported  based on the average of $100$ independent runs of the experiments.

Table \ref{table:csl} reports the $L_2$ errors and computation time of the candidate methods with total sample size $N=5\times 10^5$, the number of machines $L=20$,  and dimensionality $p$ varying from $p=100$ to $p=500$. The other parameters are set in the same way as previous experiments. Among all the settings, DC-SGD has the fastest method as it only evaluates each data point once. Nevertheless, DC-SGD $\teeDC$ fails to converge to the truth $\tee^*$ when $p$ is large. For logistic regression, both Dis-FONE and CSL methods achieve almost optimal performance as compared to the ERM, while Dis-FONE accelerates CSL. For quantile regression, Dis-FONE outperforms DC-QR in terms of both computation time and statistical accuracy since the DC method \citep{volgushev2017distributed} suffers from the restriction on the sub-sample size analogously as in the case of DC-SGD. 
}
\bibliographystyle{chicago}
\bibliography{ref}

\end{document}